\documentclass[11pt,a4paper]{article}
\usepackage{emnlp2020}

\usepackage[utf8]{inputenc} 
\usepackage[T1]{fontenc}    
\usepackage{hyperref}       
\usepackage{url}            
\usepackage{booktabs}       
\usepackage{amsfonts}       
\usepackage{nicefrac}       
\usepackage{graphicx}
\usepackage{subcaption}
\usepackage{amsmath}
\usepackage{amssymb}
\usepackage{times}

\usepackage{latexsym}
\usepackage{multirow,multicol}
\usepackage{array}
\usepackage{soul}
\usepackage{pifont}
\usepackage[colorinlistoftodos]{todonotes}
\usepackage[capitalize]{cleveref}

\usepackage{mathptmx}
\usepackage{anyfontsize}
\usepackage{t1enc}
\usepackage{natbib}

\newcommand{\PreserveBackslash}[1]{\let\temp=\\#1\let\\=\temp}
\newcolumntype{C}[1]{>{\PreserveBackslash\centering}p{#1}}
\newcolumntype{R}[1]{>{\PreserveBackslash\raggedleft}p{#1}}
\newcolumntype{L}[1]{>{\PreserveBackslash\raggedright}p{#1}}
\newcommand{\xmark}{\ding{55}}%

\newcommand*\rot{\rotatebox{90}}

\usepackage{array,etoolbox}
\preto\tabular{\setcounter{magicrownumbers}{0}}
\newcounter{magicrownumbers}

\title{Utterance-level Dialogue Understanding: \\ An Empirical Study to Understand the Behavior of the Baselines}
\title{Utterance-level Dialogue Understanding: An Empirical Study}

\author{Deepanway Ghosal$^\dagger$, Navonil Majumder$^\dagger$,
  Rada Mihalcea$^\triangle$,
  Soujanya Poria$^\dagger$\\\\
  $^\dagger$ Singapore University of Technology and Design, Singapore\\
  $^\triangle$ University of Michigan, USA\\
  \texttt{deepanway\_ghosal@mymail.sutd.edu.sg}\\
  \texttt{\{navonil\_majumder, sporia\}@sutd.edu.sg}\\ \texttt{mihalcea@umich.edu}
  }

\aclfinalcopy
\begin{document}

\maketitle

\begin{abstract}

The recent abundance of conversational data on the Web and elsewhere calls for effective NLP systems for dialog understanding. Complete utterance-level understanding often requires context understanding, defined by nearby utterances. In recent years, a number of approaches have been proposed for various utterance-level dialogue understanding tasks. Most of these approaches account for the context for effective understanding. In this paper, we explore and quantify the role of context for different aspects of a dialogue, namely \textit{emotion, intent, and dialogue act} identification, using  state-of-the-art dialog understanding methods as baselines. Specifically, we employ various perturbations to distort the context of a given utterance and study its impact on the different  tasks and baselines. This provides us with insights into the fundamental contextual controlling factors of different  aspects of a dialogue. Such insights can inspire more effective dialogue understanding models, and provide support for future text generation approaches. The implementation pertaining to this work is available at \url{https://github.com/declare-lab/dialogue-understanding}.

\end{abstract}

\section{Introduction}
\begin{figure}[t]
	\centering
	\includegraphics[width=\linewidth]{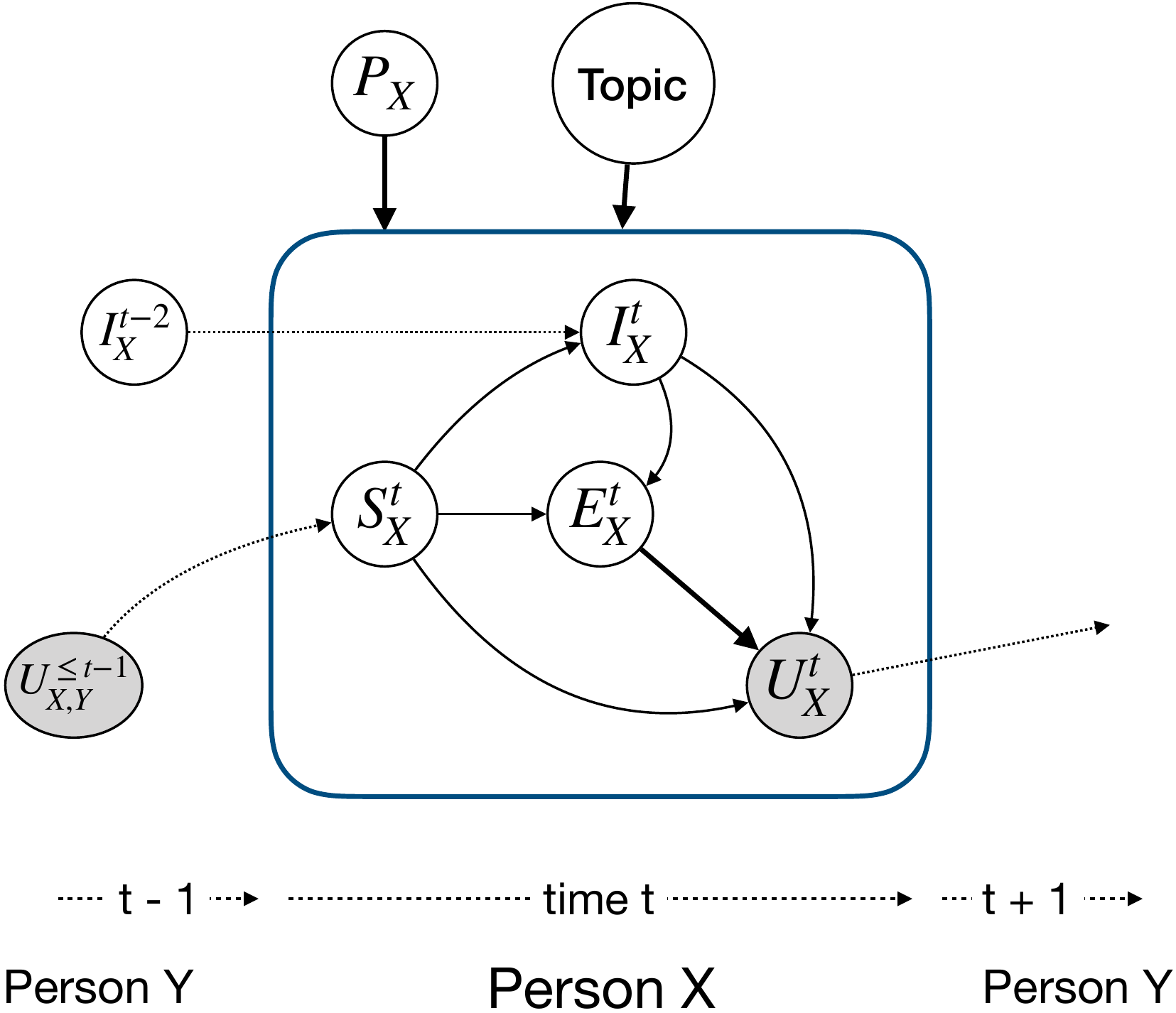}
	\caption{Dyadic conversation--between person X and Y--are governed by interactions between several latent factors such as intents, emotions.}
	\label{fig:controlling_vars}
\end{figure}

Human-like conversational systems are a long-standing goal of Artificial Intelligence (AI). However, the development of such systems is not a trivial task, as we often participate in dialogues by relying on several factors such as emotions, sentiment, prior assumptions, intent, or personality traits. In \cref{fig:controlling_vars}, we illustrate a dialogue-generation mechanism that leverages these key variables. In this illustration, $P$ represents the personality of the speaker; $S$ represents the speaker-state; $I$ denotes the intent of the speaker; $E$ refers to the speaker's emotional-aware state, and $U$ refers to the observed utterance. Speaker personality and the topic always influence these variables. At turn $t$, the speaker conceives several pragmatic concepts, such as argumentation logic, viewpoint, and inter-personal relationships---which we collectively represent using the speaker-state $S$~\citep{hovy1987generating}. Next, the intent $I$ of the speaker is formulated based on the current speaker-state and previous intent of the same speaker (at $t-2$). These two factors influence the emotion of the speaker. Finally, the intent, the speaker state, and the speaker's emotion jointly manifest as the spoken utterance.
\begin{figure*}[t]
    \centering
    \begin{subfigure}{0.49\textwidth}
     \includegraphics[width=\linewidth]{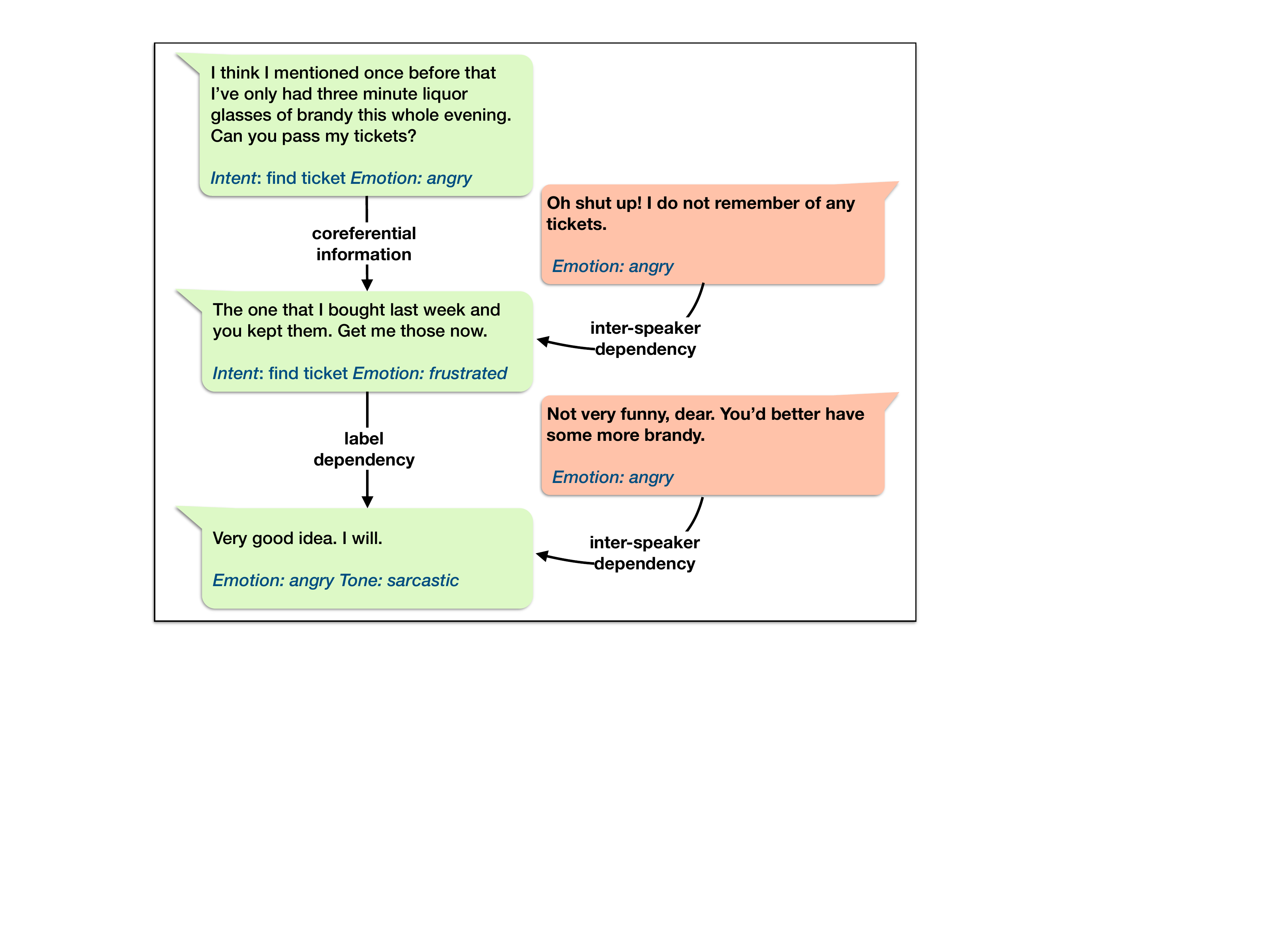}
      \label{fig:context-role}
     \end{subfigure}
     \begin{subfigure}{0.49\textwidth}
     \includegraphics[width=\linewidth]{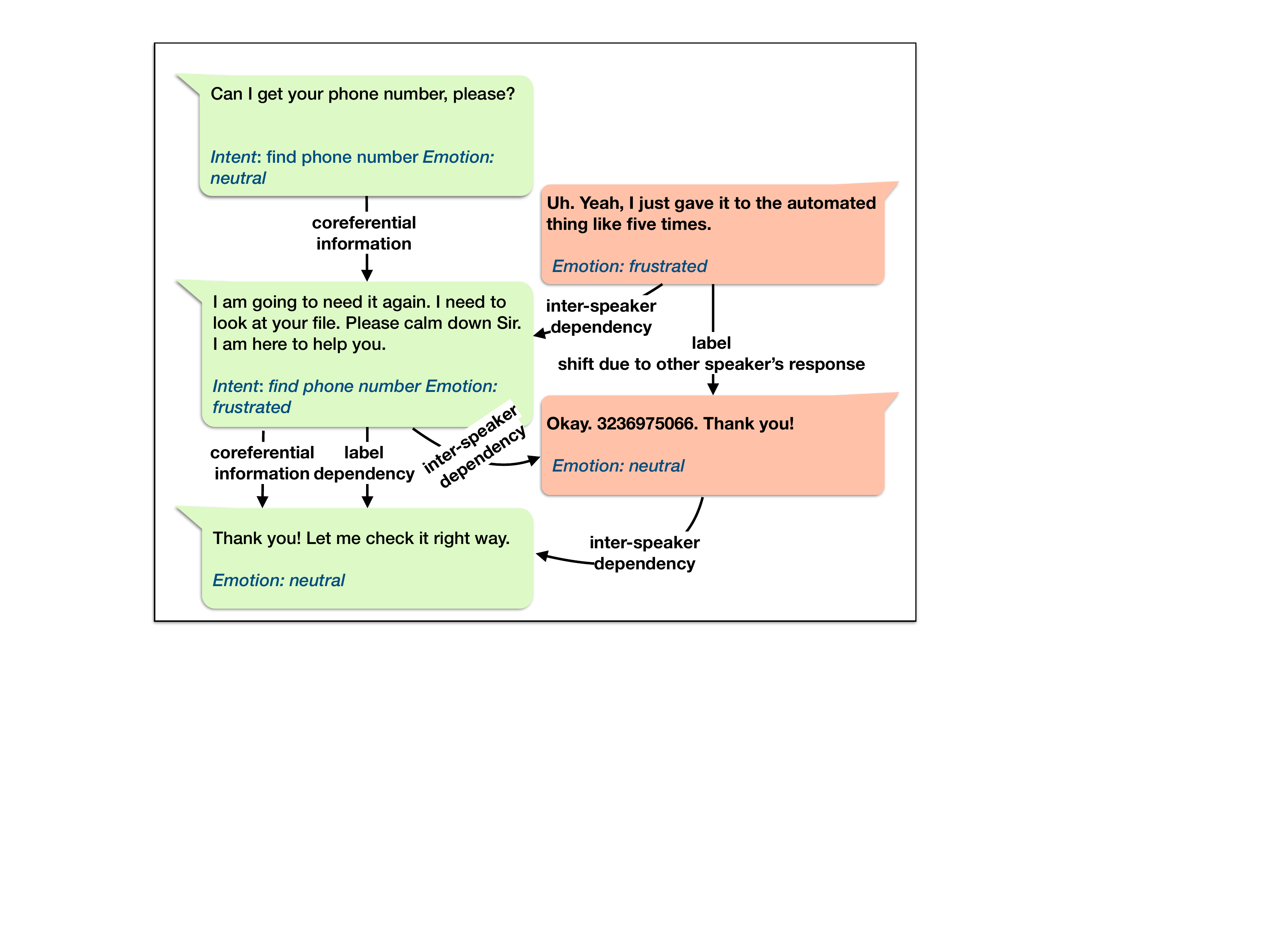}
      \label{fig:context-role2}
     \end{subfigure}
     \caption{Role of Context in Utterance Level Dialogue Understanding.}
     \label{fig:context-role-all}
\end{figure*}
It is thus not surprising that the landscape of dialogue understanding research embraces several challenging tasks, such as emotion recognition in conversations (ERC), dialogue intent classification, user state representation, and others. These tasks are often performed at utterance level and can be conjoined together under the umbrella of utterance-level dialogue understanding. Due to the fast-growing research interest in dialogue understanding, several novel approaches have lately been proposed~\citep{qindcr2020dcrnet,rashkin2018empathetic, xing2020adapted,lian2019domain,wang2020contextualized,saha2020towards} to address these tasks by adopting speaker-specific and contextual modeling. However, to the best of our knowledge, no unified baselines have been established for varied utterance-level dialogue understanding tasks that allow comparison and analysis of these tasks under the same framework. In this work, the purpose of using a unified baseline for all the utterance-level dialogue understanding tasks is to compare the characteristics of this baseline across different tasks and datasets. As a result, we can also learn interesting attributes of the datasets and task which we discuss in detail in \cref{sec:analysis}. Recently, \citet{sankar2019neural} attempted to measure the efficacy of multi-turn contextual information in dialogue generation by probing the models tasked to generate dialogues given multi-turn contexts. According to them, the baseline dialogue models used in their work are not capable to efficiently utilize the long multi-turn sequences for dialogue generation as they are rarely sensitive to most perturbations which diverges from the findings of this work.

\paragraph{Conversational Context Modeling.}
Context is at the core of NLP research. According to several recent studies~\citep{peters2018deep,devlin2018bert}, contextual sentence and word embeddings can improve the performance of the state-of-the-art NLP systems by a significant margin.

The notion of context can vary from problem to problem. For example, while calculating word representations, the surrounding words carry contextual information. Likewise, to classify a sentence in a document, other neighbouring sentences are considered as its context. In \citet{poria-EtAl:2017:Long}, surrounding utterances are treated as context and they experimentally show that contextual evidence indeed aids in classification.

Similarly in the tasks such as conversational emotion or intent detection, to determine the emotion of an utterance at time $t$, the preceding utterances at time $<t$ can be considered as its context. However, computing this context representation often exhibits major difficulties due to emotional dynamics.

\begin{figure}[t]
    \centering
    \includegraphics[width=\linewidth]{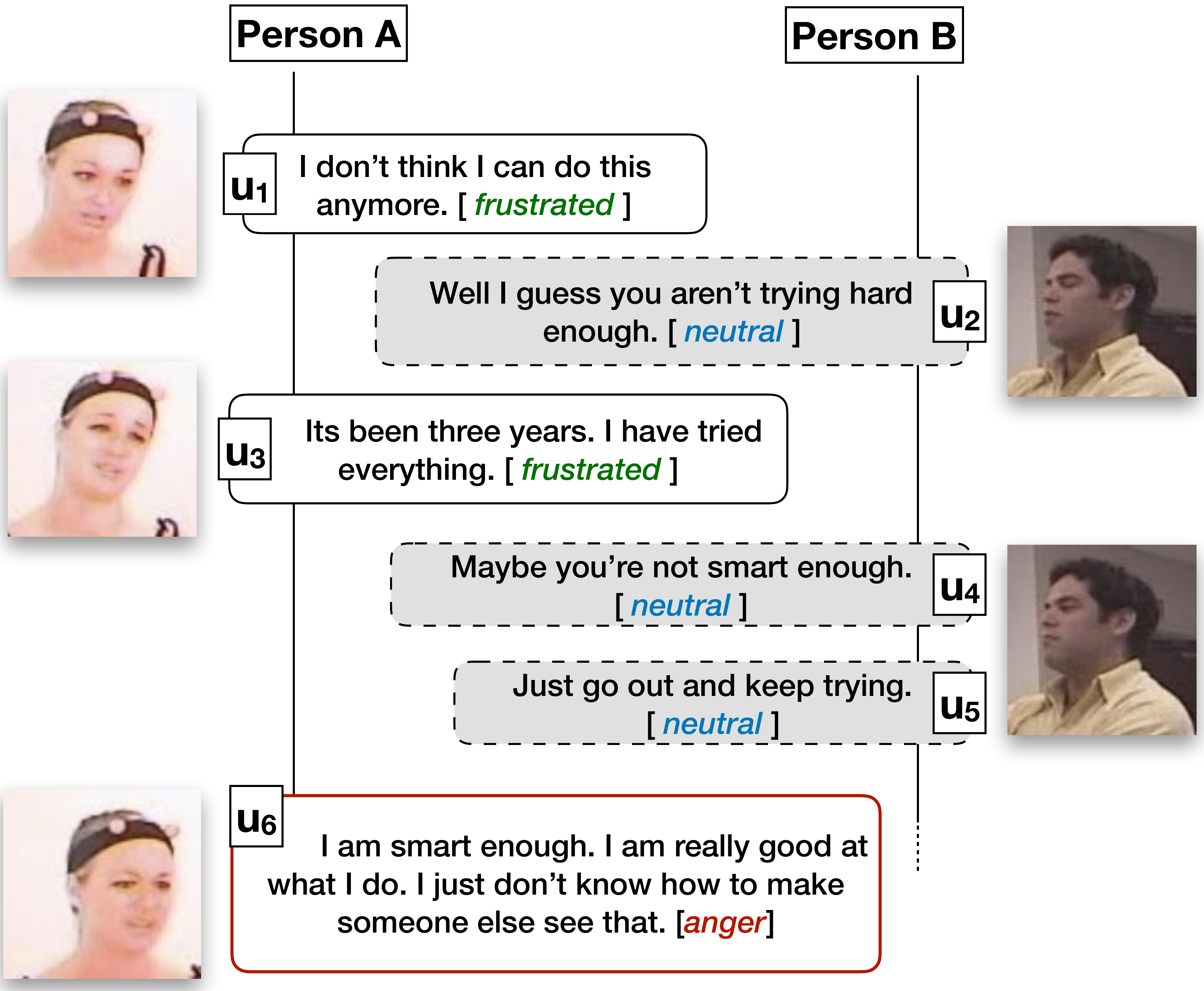}
    \caption{An abridged dialogue from the IEMOCAP dataset.}
    \label{fig:example}
\end{figure}

The dynamics of conversations consists of two important aspects: \textit{self} and \textit{inter-personal dependencies}~\citep{morris2000emotions}. Self-dependency, also known as \textit{intra inertia}, deals with the aspect of influence that speakers have on themselves during conversations~\citep{kuppens2010emotional}.
On the other hand, inter-personal dependencies relate to the influences that the counterparts induce into a speaker. Conversely, during the course of a dialogue, speakers also tend to mirror their counterparts to build rapport~\citep{navarretta2016mirroring}. This phenomenon is illustrated in \cref{fig:example}. Here, $P_a$ is frustrated over her long term unemployment and seeks encouragement ($u_1, u_3$). $P_b$, however, is pre-occupied and replies sarcastically ($u_4$). This enrages $P_a$ to appropriate an angry response ($u_6$). In this dialogue, \textit{emotional inertia} is evident in $P_b$ who does not deviate from his nonchalant behavior. $P_a$, however, gets emotionally influenced by $P_b$. Modeling self and inter-personal relationship and dependencies may also depend on the topic of the conversation as well as various other factors like argument structure, interlocutors’ personality, intents, viewpoints on the conversation, attitude towards each other etc.. Hence, analyzing all these factors are key for a true self and inter-personal dependency modeling that can lead to enriched context understanding.

The contextual information can come from both local and distant conversational history. While the importance of local context is more obvious, as stated in  recent works, distant context often plays a less important role in understanding the utterances. Distant contextual information is useful mostly in the scenarios when a speaker refers to earlier utterances spoken by any of the speakers in the conversational history.
\begin{figure*}[ht!]
    \centering
    \includegraphics[width=\linewidth]{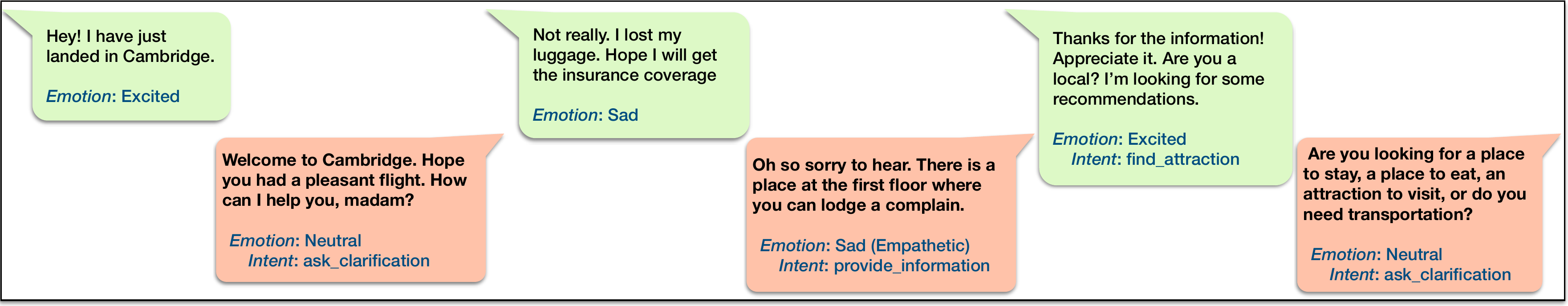}
    \caption{Utterance level tagging of a dialogue.}
    \label{fig:exampleshift0}
\end{figure*}
The usefulness of context is more prevalent in classifying short utterances, like {\it ``yeah''}, {\it ``okay''}, {\it ``no''}, that
can express different emotions depending on the context and discourse of the dialogue.

\textbf{Although the role of context is reasonably clear in dialogue generation, it may not be equally transparent in the case of utterance level dialogue understanding. There can be occasions where contextual information may not provide any useful information. In such cases, the target utterance can be sufficient for necessary inferences such as intent, act, and emotion prediction.}
However, contextual utterances in a conversation should always help in understanding an utterance at a given time as they provide key background information. Modeling representation of these contextual utterances are not trivial as there can be long-chain complex coreferential or other kinds of inferences involved in the process and sometimes various other confounding factors such as sarcasm, irony, etc. can make the task extremely challenging (see \cref{fig:context-role-all}). An ideal context modeling approach should have the ability to understand such factors efficiently, fuse them, and perform inference accordingly. \textbf{Contextual information may not be necessary to tag all the utterances in a dialogue, but we must acknowledge that full understanding of an utterance is incomplete without its conversational context. The scope of this understanding spans beyond just tagging utterances with only these predefined labels -- intents, dialogue acts, emotions -- to large-scale reasoning that includes difficult problems like identifying causes of eliciting a particular response, slot filling, etc.} Efficient and well-formed contextual representation modeling can also depend on the type of conversations i.e., modeling persuasive dialogue can be different from a debate. Finding contextualized conversational utterance representations is an active area of research. Leveraging such contextual clues is a difficult task. Memory networks, RNNs, and attention mechanisms have been used in previous works~\cite{qindcr2020dcrnet} to grasp information from the context. However, these networks do not address most of the abovementioned aspects e.g., lack of coreference resolution, understanding long-chained inferences between the speakers. In this work, we, of course, do not attempt to propose a network that inherits all these factors rather we try to probe existing contextual and non-contextual models in utterance level dialogue understanding and understand theirs underneath working-principle.

In this work, we adapt, modify, and employ two strong contextual utterance-level dialogue understanding baselines---bcLSTM~\citep{poria-EtAl:2017:Long} and DialogueRNN~\citep{dialoguernn}---that we evaluate on four large dialogue classification datasets across five different tasks. As shown in \cref{fig:controlling_vars}, conversational context, inter-speaker dependencies, and speaker states can play important roles in addressing these utterance-level dialogue understanding tasks.
bcLSTM and DialogueRNN are two such frameworks that leverage these factors and thus considered as the baselines in our experiment.
Moreover, we present several unique probing strategies and experimental designs that evaluate the role of context in utterance-level dialogue understanding. To summarize, the purpose of this work is to decipher the role of the context in utterance-level dialogue understanding by the means of different probing strategies. These strategies can be easily adapted to other tasks for similar purposes.

The contribution of this work is five-fold:
\begin{itemize}
    \item We setup contextual utterance-level dialogue understanding baselines for five different utterance-level dialogue understanding tasks with traditional word embeddings (GloVe) and recent transformer-based contextualized word embeddings (RoBERTa);
    \item We modified the existing strong baselines LSTM and DialogueRNN by introducing residual connections that improve these baselines by a significant margin;
    \item We showcase a detailed dataset analysis of different tasks and present interesting frequent label transition patterns and dependencies;
    \item We perform an evaluation of two different mini-batch construction paradigms: utterance- and dialogue-level mini-batches;
    \item We propose varied probing strategies to decipher the role of context in utterance-level dialogue understanding.
\end{itemize}


\section{Task definition}
Given the transcript of a conversation along with speaker information
of each constituent utterance, the utterance-level dialogue understanding
task aims to identify the label of each utterance from a set of  pre-defined labels that can be either a set of emotions, dialogue acts, intents etc. \cref{fig:exampleshift0} illustrates one such
conversation between two people, where each utterance is labeled by the
underlying emotion and intent. Formally, given the input sequence of $N$ number of
utterances $[(u_1, p_1), (u_2,p_2),\dots, (u_N,p_N)]$, where each utterance $u_i=[u_{i,1},u_{i,2},\dots,u_{i,T}]$ consists of $T$ words $u_{i,j}$ and spoken by
party $p_i$, the task is to predict the label $e_i$ of
each utterance $u_i$. In this process, the classifier can also make use of the conversational context. There are also cases where not all the utterances in a dialogue have corresponding labels. In this paper, we limit utterance-level dialogue understanding to only tagging utterances with emotions, dialogue-acts, and intents. However, we do not claim these are the only tasks that give us a full understanding of the utterances in dialogues. As discussed in the introduction, the scope of this research topic includes various other harder problems that we do not address in this paper such as slot filling, identifying sarcasm, finding causes of responses, etc.

\section{Models}
We train all our classification models in an end-to-end setup. We first extract utterance level features with either i) a CNN module with pretrained GloVe embeddings or ii) a pretrained RoBERTa model. The resulting extracted features are non-contextual in nature as they are obtained from utterances without the surrounding context. We then classify the utterances with one of the following three models: i) Logistic Regression, or ii) bcLSTM, or iii) DialogueRNN. Among these models, the Logistic Regression model is non-contextual in nature, whereas the other two are contextual. We expand on the feature extractor and the classifier in more detail next.
\begin{figure*}[hbtp]
    \centering
     \includegraphics[width=0.9\linewidth]{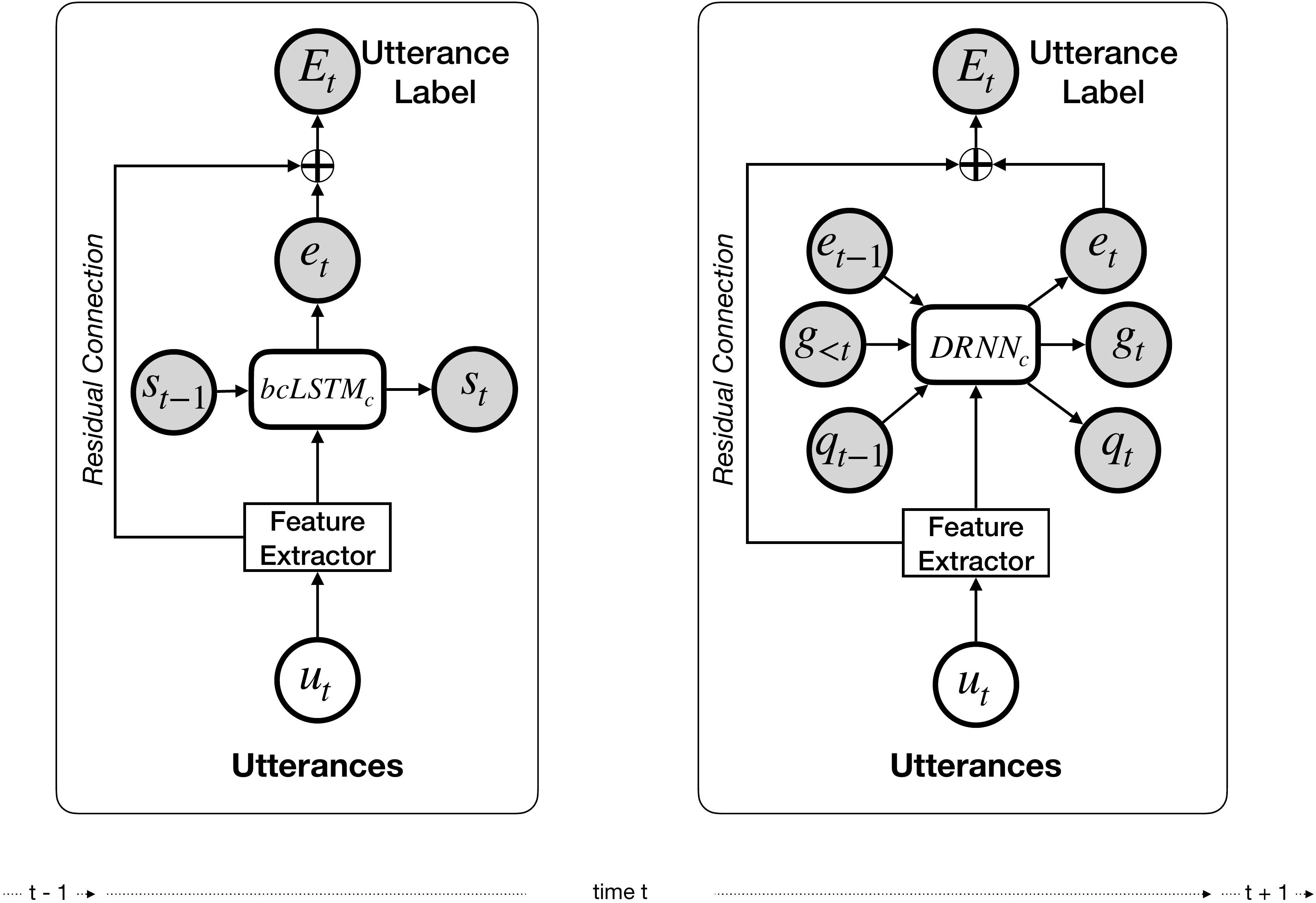}
     \label{fig:residual}
     \caption{Modified bcLSTM and DialogueRNN with residual connection.}
\end{figure*}
\subsection{Utterance Feature Extractor}
Utterance level features are extracted using one of the following two methods:

\paragraph{GloVe CNN.}
A convolutional neural network \citep{kim2014convolutional} is used to extract features from the utterances of the conversation. We use a single convolutional layer followed by max-pooling and a fully-connected layer to obtain the representation of the utterance. The inputs to this network are the utterances. Each word in the utterances is initialized with 300-dimensional pretrained GloVe embeddings \citep{pennington2014glove}. We pass these word to convolutional filters of sizes 1, 2, and 3, each having 100 feature maps. The output of these filters are then max-pooled across all the words of an utterance. These are then concatenated and fed to a $100$ dimensional fully-connected layer followed by ReLU activation~\citep{nair2010rectified}. The output after the activation form the final representation of the utterance.

\paragraph{RoBERTa.}
We employ the RoBERTa-Base model \cite{liu2019roberta} to extract utterance level feature vectors. RoBERTa-Base follows the original BERT-Base \citep{devlin2018bert} architecture having 12 layers, 12 self-attention heads in each block, and a hidden dimension of 768 resulting in a total of 125M parameters. Let an utterance $x$ consists of a sequence of BPE tokenized tokens $x_1, x_2, \dots, x_N$. A special token $[CLS]$ is appended at the beginning of the utterance to create the input sequence for the model: $[CLS], x_1, x_2, \dots, x_N$. This sequence is passed through the model, and the activation from the last layer corresponding to the $[CLS]$ token is used as the utterance feature.

\begin{figure*}[t]
    \centering
     \includegraphics[width=\linewidth]{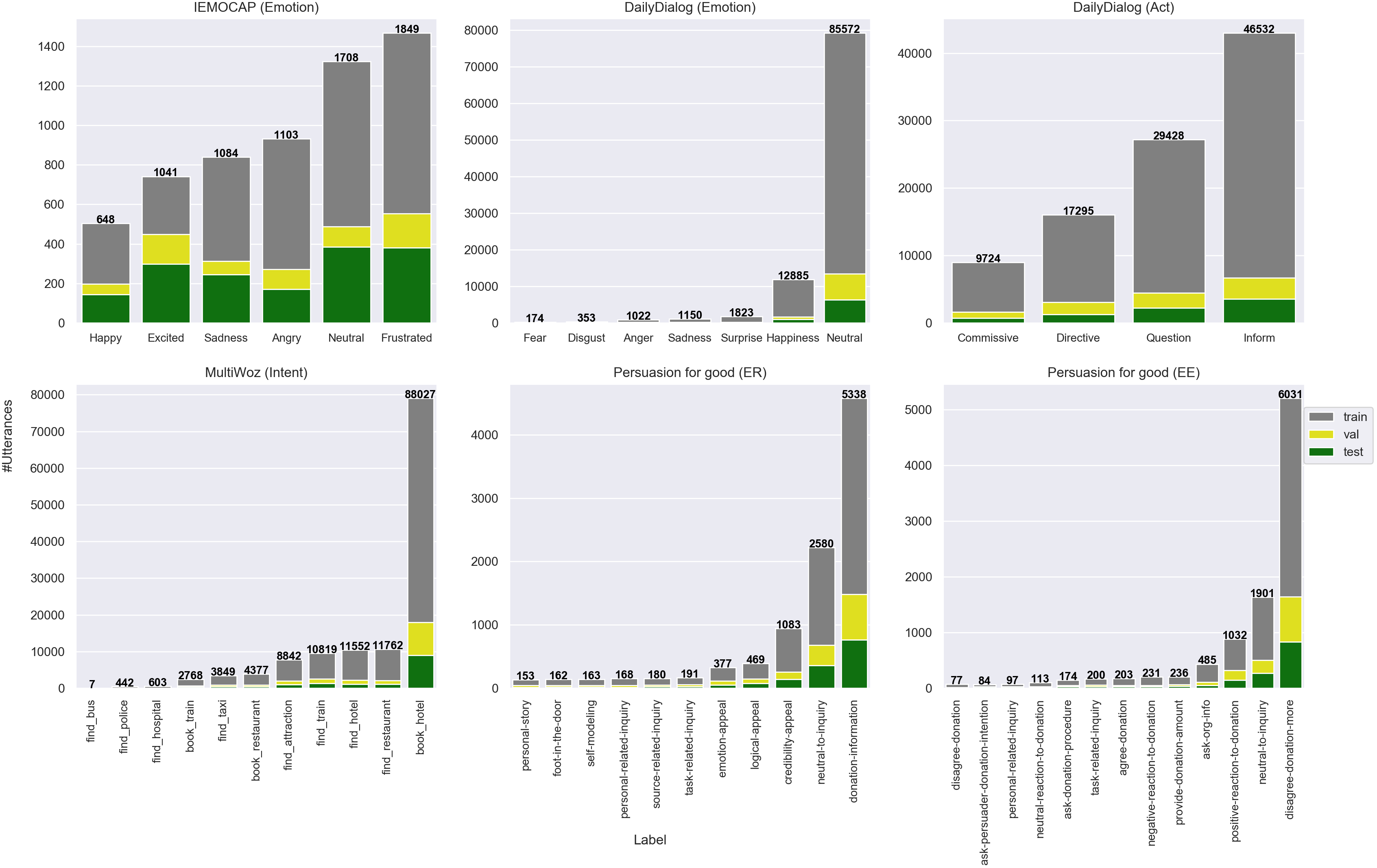}
     \caption{Label distribution of the datasets.}
     \label{fig:dataset-distro}
\end{figure*}

\subsection{Utterance Classifier.}
The representations obtained from the \textit{Utterance Feature Extractor} are then classified using one of the following three methods:

\paragraph{Without Context Classifier.} In this model, classification of an utterance is performed using a fully connected multi-layer perceptron layer. This classification setup is non-contextual in nature as there is no flow of information from the contextual utterances. This strategy translates to simple GloVe CNN based or RoBERTa feature based fine-tuning in isolation w.r.t other utterances in the conversation as we don't take those into account. For simplicity, we call this model GloVe CNN or RoBERTa LogReg (Logistic Regression).
\begin{figure*}[ht!]
  \centering
  \begin{subfigure}{\textwidth}
    \centering
     \includegraphics[width=0.7\linewidth]{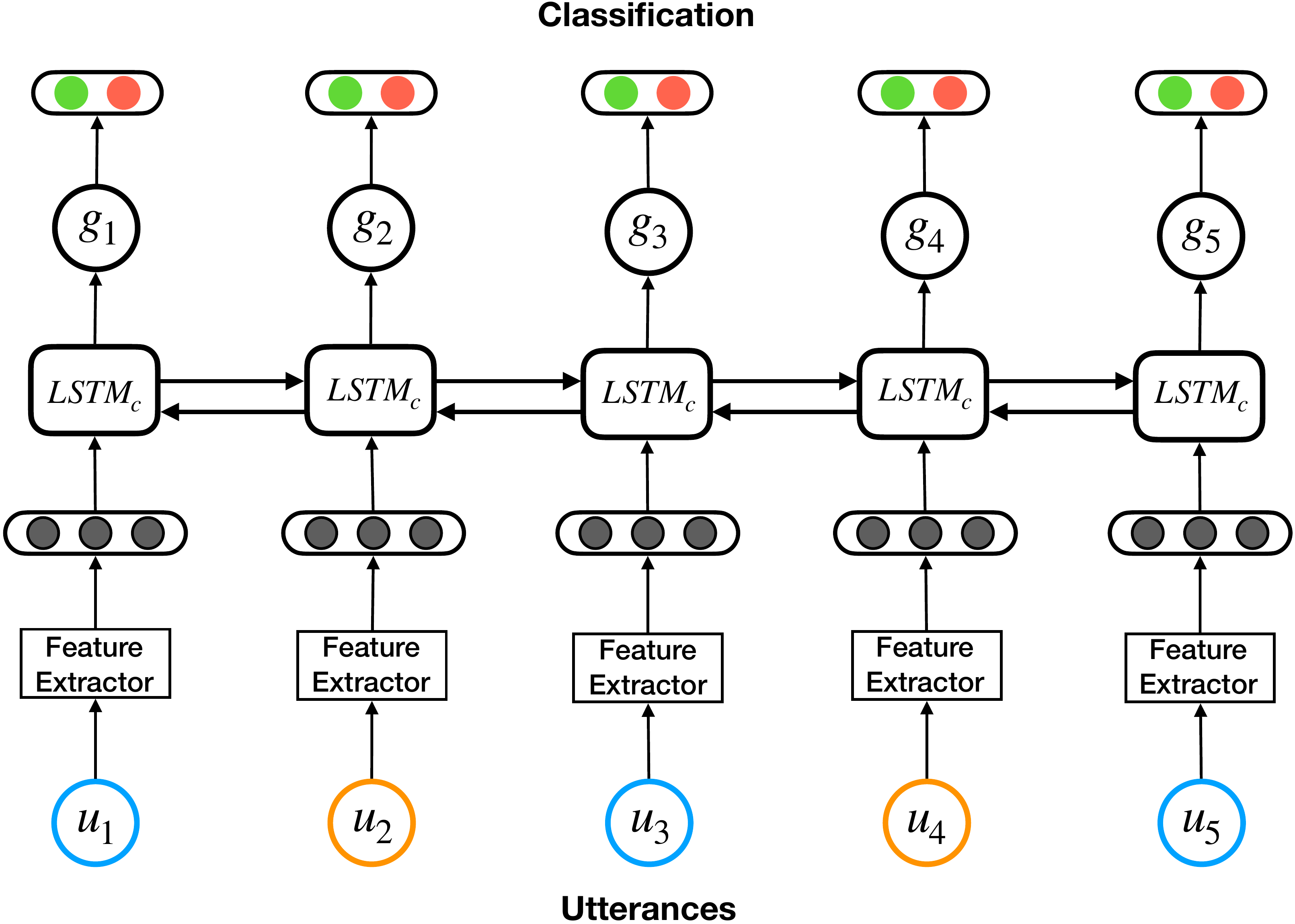}
     \caption{}\label{fig:bclstm}
\end{subfigure}
  \begin{subfigure}{\textwidth}
  \centering
  \includegraphics[width=0.75\linewidth]{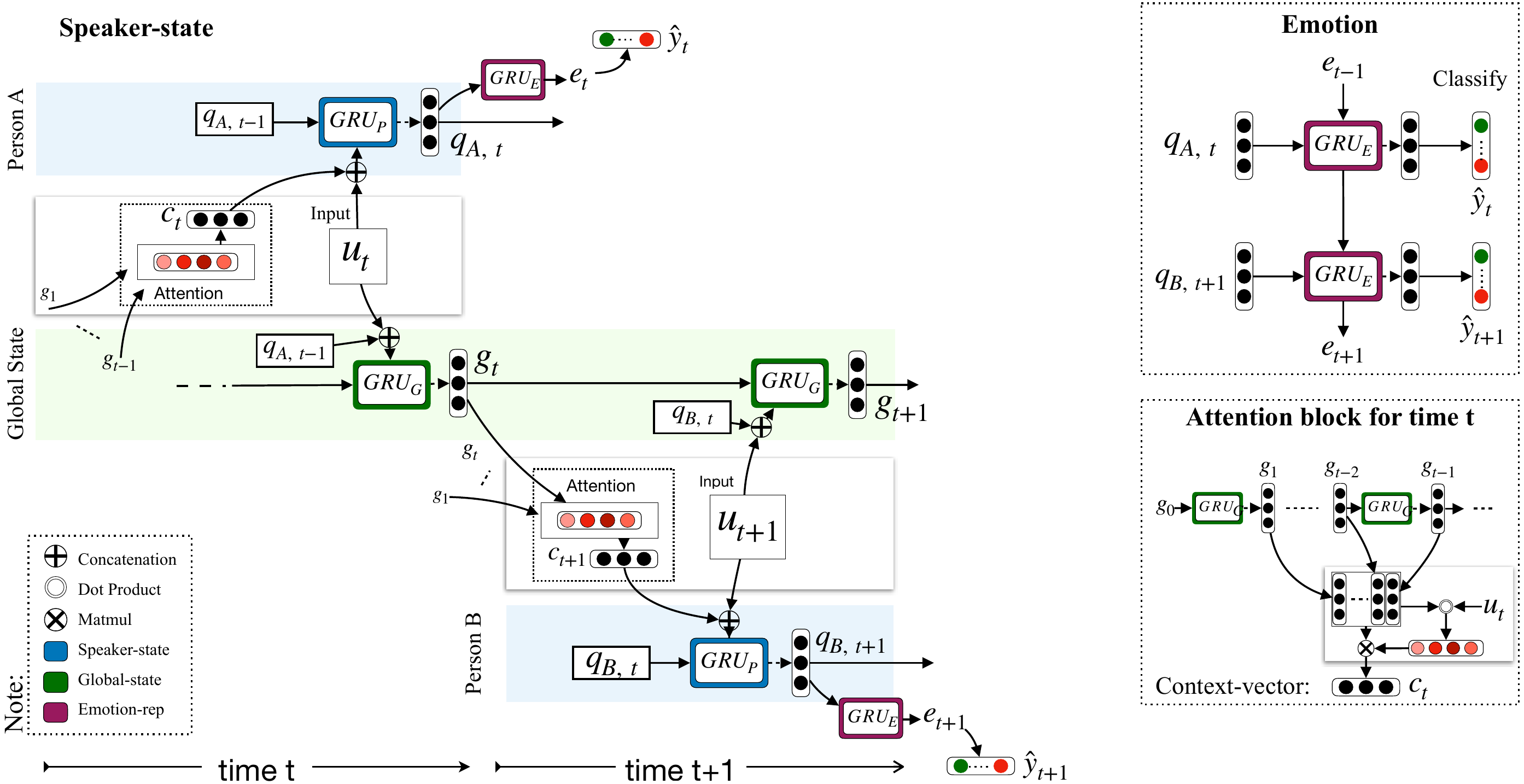}
  \includegraphics[width=0.24\linewidth]{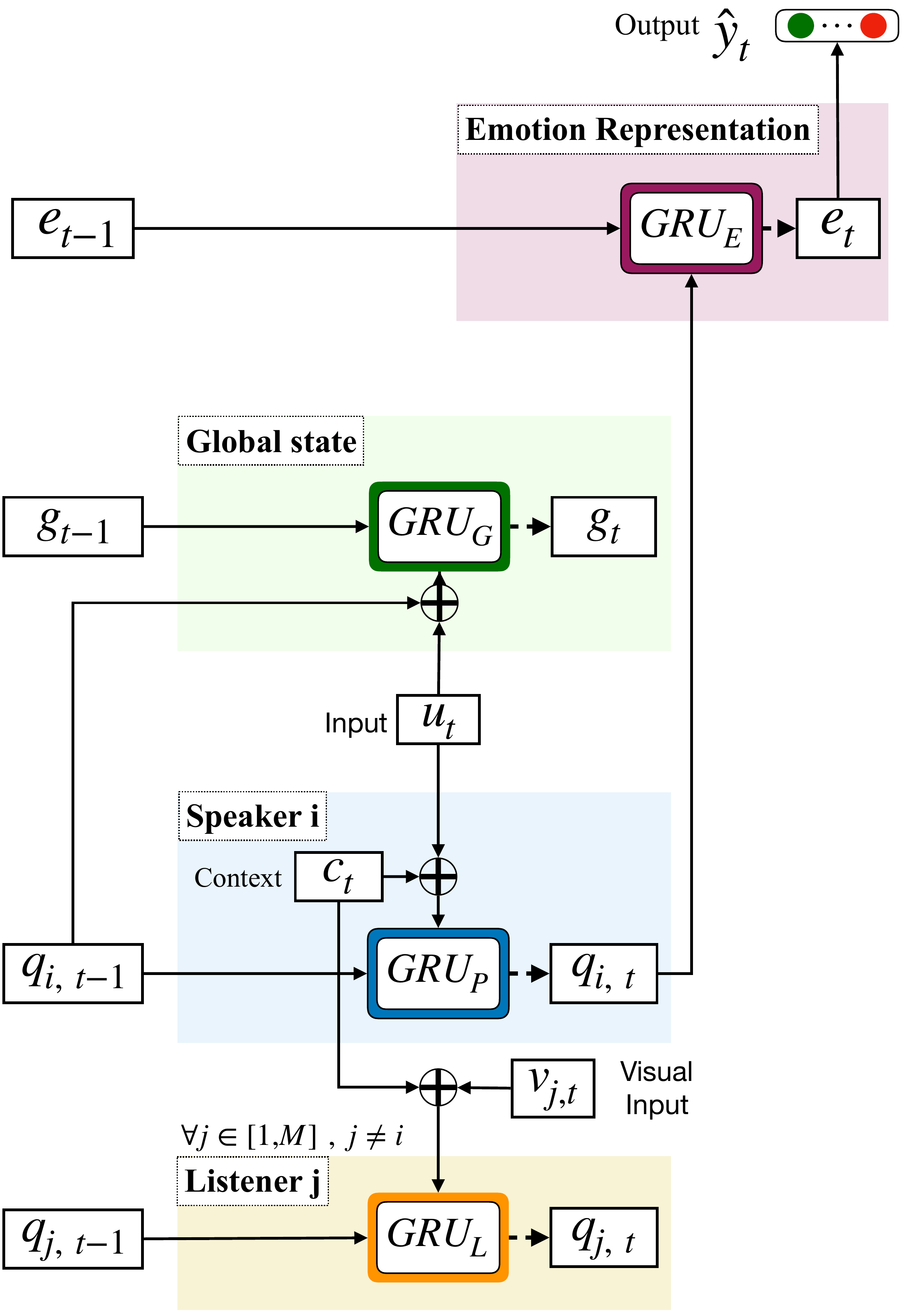}
  \caption{}\label{fig:drnn}
  \end{subfigure}
  \caption{(a). Original bcLSTM model architecture with GloVe CNN/RoBERTa utterance feature extractor. (b). Left: Original DialogueRNN architecture. Right: Update schemes for global, speaker, listener, and emotion states for $t^{th}$ utterance in a
	dialogue. Here, Person $i$ is the speaker and Persons $j \in [1, M] \ \text{and} \ j \neq i$ are the listeners.}
	\label{fig:main}
\end{figure*}
\paragraph{bcLSTM.}
The Bidirectional Contextual LSTM model (bcLSTM) \citep{poria-EtAl:2017:Long} creates context-aware utterance representations by capturing the contextual content from the surrounding utterances using a Bi-directional LSTM \citep{hochreiter1997long} network. The feature representations extracted by the \textit{Utterance Feature Extractor} serves as the input to the LSTM network. Finally, the context-aware utterance representations from the output of the LSTM are used for the label classification. The contextual-LSTM model is speaker independent as it doesn't model any speaker level dependency. An illustration of the bcLSTM model is shown in \cref{fig:bclstm}.

\paragraph{cLSTM.}
Similar to bcLSTM but without the bidirectionality in the LSTM, this model is intended to ignore the presence of future utterances while classifying an utterance $U_t$.

\paragraph{DialogueRNN.} \citep{dialoguernn} is a recurrent network based model for emotion recognition in conversations. It uses two GRUs to track individual speaker states and global context during the conversation. Further, another GRU is employed to track emotion state through the conversation. In this work, we consider the emotion state to be a general state which can be used for utterance level classification (i.e., not limited to only emotion classification). Similar to the bcLSTM model, the features extracted by the \textit{Utterance Feature Extractor} is the input to the DialogueRNN network. DialogueRNN aims to model inter-speaker relations and it can be applied on multiparty datasets. An illustration of the DialogueRNN model is shown in \cref{fig:drnn} which shows the speaker state modeling using its recurrent structure to count on both intra- and inter-speaker dependencies.

\paragraph{cLSTM, bcLSTM and DialogueRNN with Residual Connections.}
Deep neural networks can often have difficulties in information prorogation. Multi-layered RNN-like in particulars often succumb to vanishing gradient problems while modeling long range sequences. Residual connections or skip connections \citep{he2016deep} are an intuitive way to tackle this problem by improving information propagation and gradient flow. Inspired by the early works in residual LSTM~\cite{wu2006emotion,kim2017residual}, in our recurrent contextual models - bcLSTM and DialogueRNN we adopt a simple strategy to introduce a residual connection. For each utterance, a residual connection is formed between the output of the feature extractor and the output of the bcLSTM/DialogueRNN module. These two vectors are added and the final classification is performed from the resultant vector.

\section{Experimental Setup}
\label{sec:experiment-setup}
\begin{table*}[ht!]
\centering
        \begin{tabular}{l|c|c|c|c|c|c}
            \toprule
            \multirow{2}{*}{Dataset}&\multicolumn{3}{c|}{$\#$ dialogues}&\multicolumn{3}{c}{$\#$ utterances}\\
            &train&val&test&train&val&test\\
            \hline
            IEMOCAP & 108 & 12 & 31 & 5163 & 647 & 1623\\
            DailyDialog & 11,118 & 1,000 & 1,000 & 87,179 & 8,069 & 7,740\\
            MultiWOZ & 8438 & 1000 & 1000 & 113556 & 14748 & 14744 \\
            Persuasion For Good & 220 & 40 & 40 & 7902 & 1451 & 1511 \\
            \bottomrule
            \multicolumn{7}{c}{}
        \end{tabular}
        \begin{tabular}{l|c|c|c}
            \toprule
            Dataset & Classification Task &{$\#$ classes} & Metric \\
            \hline
            IEMOCAP & Emotion & 6 & Weighted Avg F1 \\
            \multirow{2}{*}{DailyDialog} & Emotion & 6* & Weighted Avg, Macro, Micro F1\\
            & Act & 4 & Weighted Avg, Macro F1\\
            MultiWOZ & Intent & 10 & Weighted Avg, F1\\
            \multirow{2}{*}{Persuasion For Good} & Persuader & 11 &  Weighted Avg, Macro F1\\
            & Persuadee & 13 & Weighted Avg, Macro F1\\
            \bottomrule
        \end{tabular}
    \caption{Statistics of splits and evaluation metrics used in different datasets. \textit{Neutral}* classes constitutes to 83\% of the DailyDialog dataset. These are excluded when calculating the metrics in DailyDialog.}
    \label{table:data}
\end{table*}
\subsection{Datasets}
All the dialogue classification datasets that we consider in this work consists of two-party conversations. We benchmark the models on the following datasets:

\textbf{IEMOCAP} \citep{iemocap} is a dataset of two person conversations among ten different unique speakers. The train and validation set dialogues come from the first eight speakers, whereas the test set dialogues are from the last two speakers. Each utterance is annotated with one of the following six emotions: \textit{happy, sad, neutral, angry, excited,} and \textit{frustrated}.

\textbf{DailyDialog} \citep{li2017dailydialog} is a manually labelled multi-utterance dialogue dataset. The dataset covers various topics about our daily life and follows the natural human communication approach. All utterances are labeled with both emotion categories and dialogue acts (intention). The emotion can belong to one of the following seven labels: \textit{anger, disgust, fear, joy, neutral, sadness}, and \textit{surprise}. The \textit{neutral} label is the most frequent emotion category in this dataset, with around 83\% utterances belonging to the class. The emotion class distribution is thus highly imbalanced in this dataset. In comparison, the dialogue act label distribution is relatively more balanced. The act labels can belong to the following four categories: \textit{inform}, \textit{question}, \textit{directive}, and \textit{commissive}.
\begin{figure*}[t]
    \centering
    \begin{subfigure}{\textwidth}
     \includegraphics[width=\linewidth]{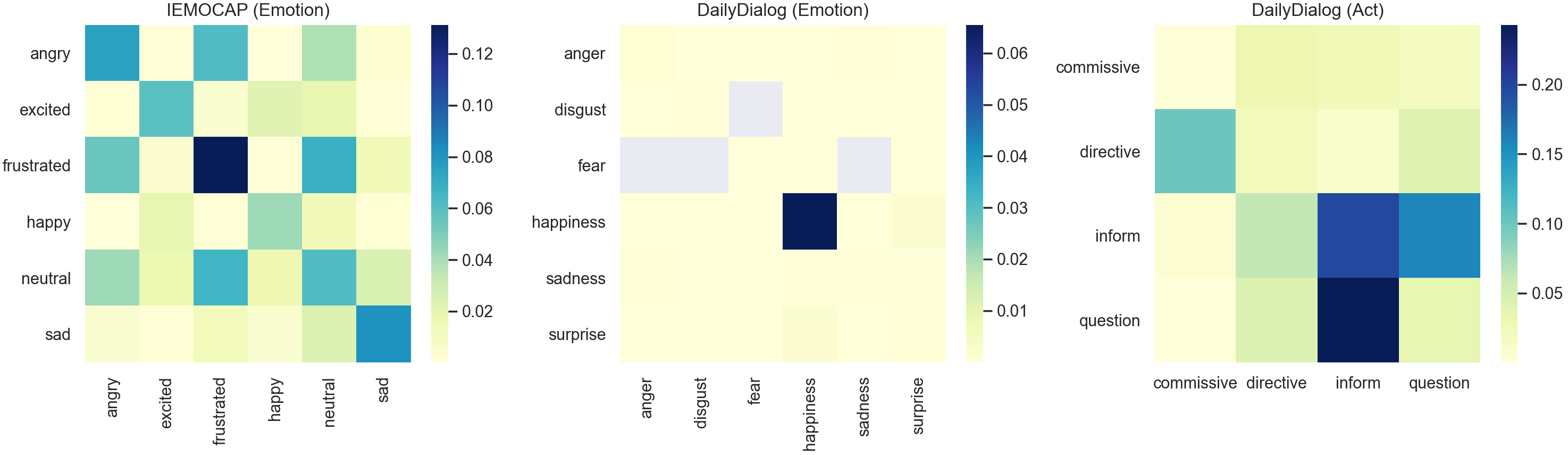}
      \label{fig:iemocap-dd-global}
     \end{subfigure}
     \begin{subfigure}{\textwidth}
     \includegraphics[width=\linewidth]{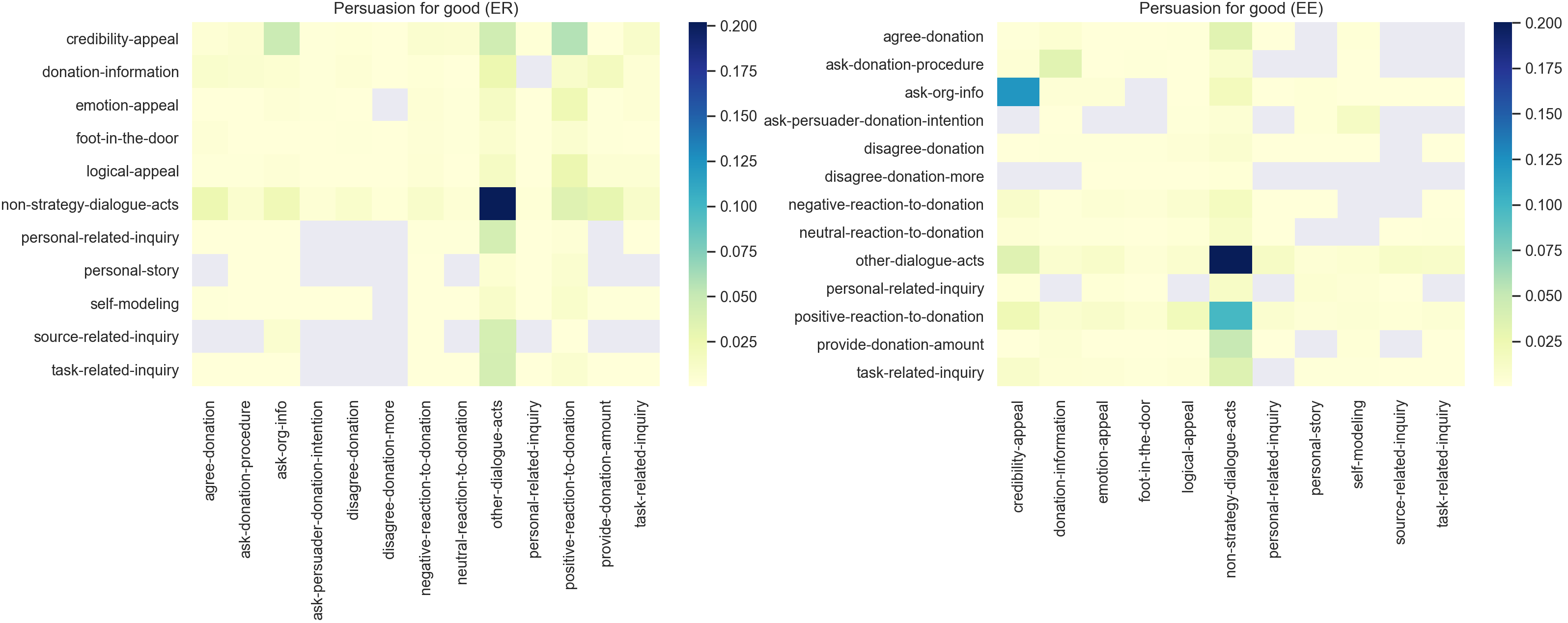}
    \label{fig:er-ee-multiwoz-global}
    \end{subfigure}
         \caption{The heatmap of inter-speaker label transition statistics in the datasets. The color bar represents normalized number of inter-speaker transitions such that elements of each matrix add up to 1. Inter-speaker transitions are not defined in MultiWOZ as system side utterances are not labeled. Note: For the DailyDialog dataset, we ignore the \emph{neutral} emotion in this figure.}
     \label{fig:heatmap1}
\end{figure*}

\begin{figure*}[hbtp!]
    \centering
    \begin{subfigure}{\textwidth}
     \includegraphics[width=\linewidth]{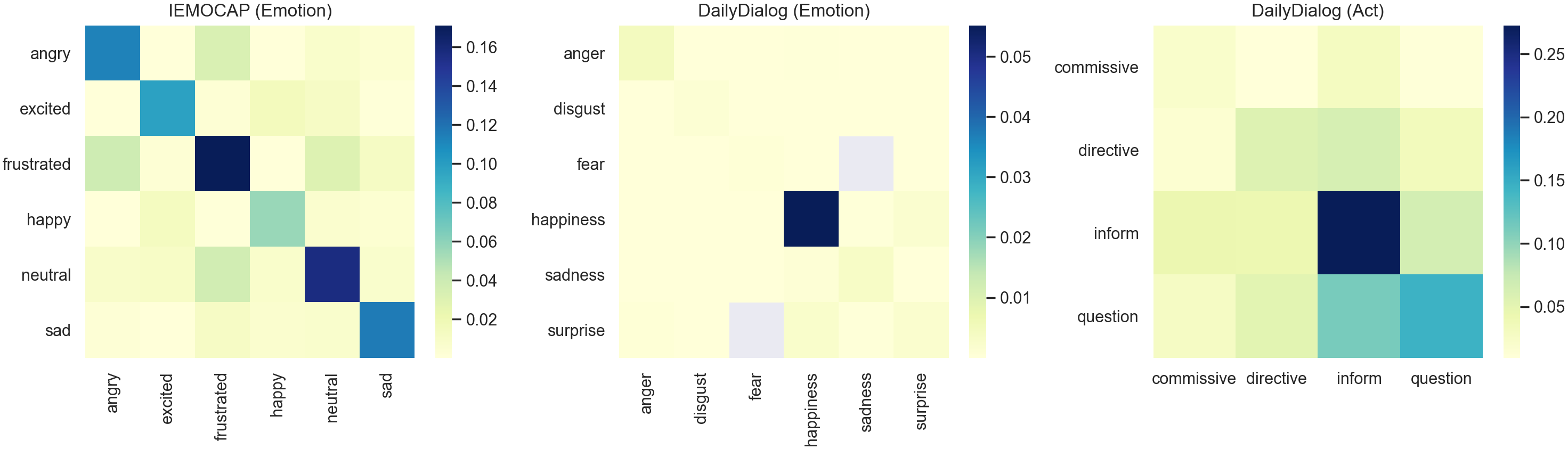}
      \label{fig:iemocap-dd-speaker}
     \end{subfigure}
     \begin{subfigure}{\textwidth}
     \includegraphics[width=\linewidth]{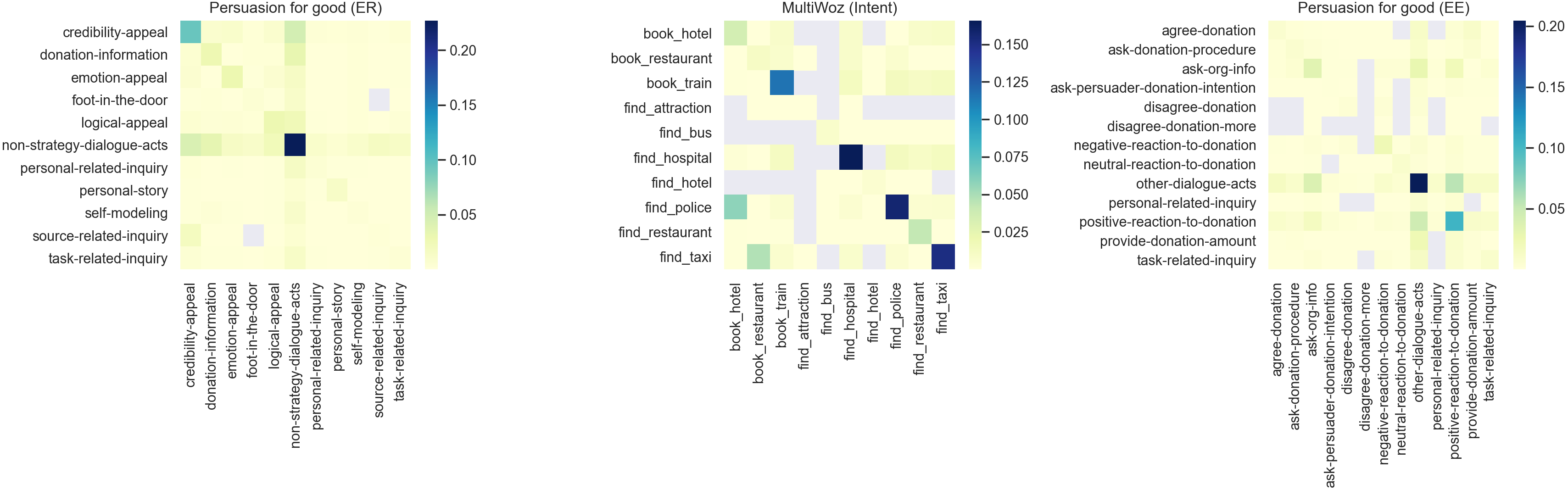}
    \label{fig:er-ee-multiwoz-speaker}
    \end{subfigure}
     \caption{The heatmap of intra-speaker label transition statistics in the datasets. The color bar represents normalized number of intra-speaker transitions such that elements of each matrix add up to 1. Note: For the DailyDialog dataset, we ignore the \emph{neutral} emotion in this figure.}
     \label{fig:heatmap2}
\end{figure*}

\begin{figure*}[ht!]
    \centering
    \begin{subfigure}{0.49\textwidth}
     \includegraphics[width=\linewidth]{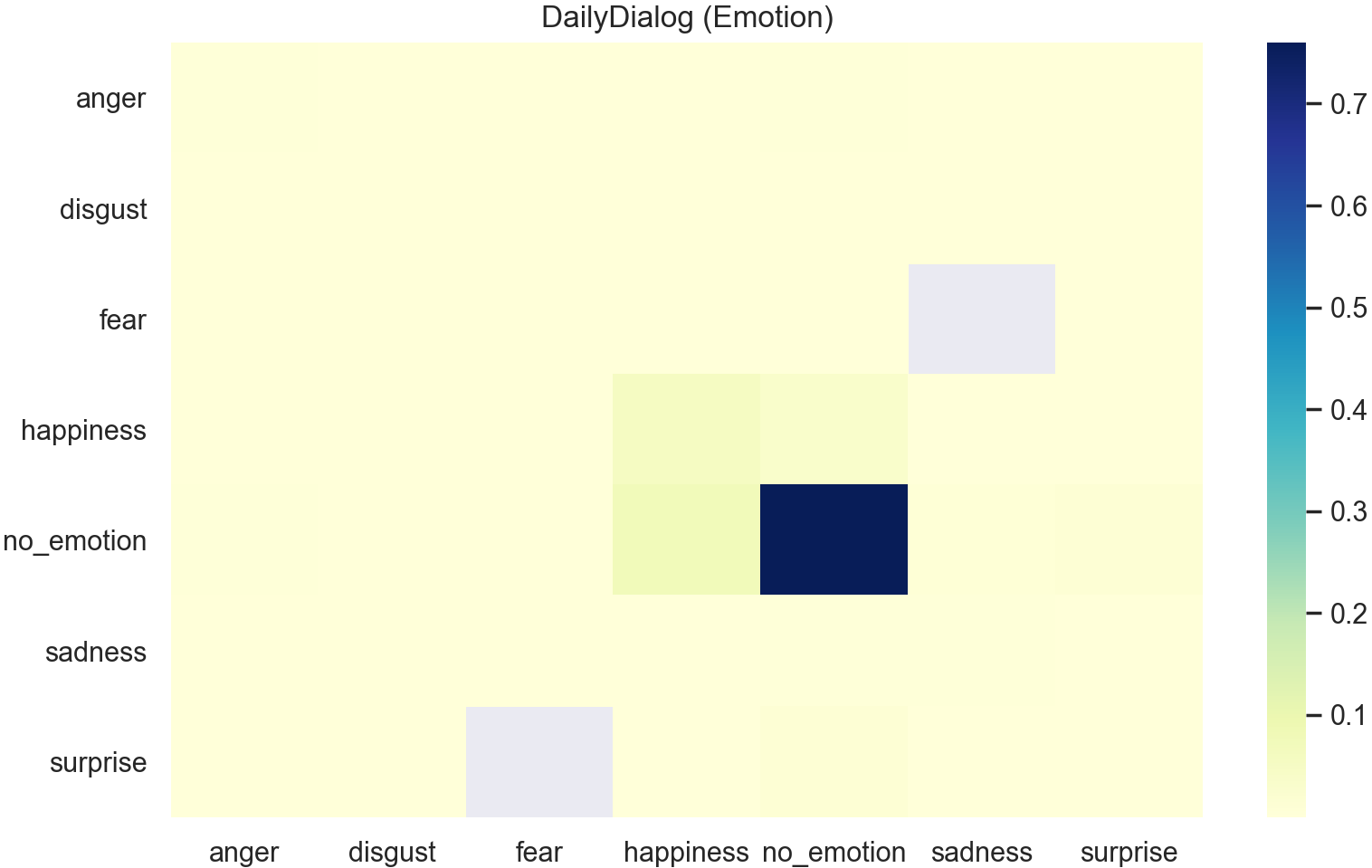}
      \label{fig:dd-intra-w/-neutral}
     \end{subfigure}
     \begin{subfigure}{0.49\textwidth}
     \includegraphics[width=\linewidth]{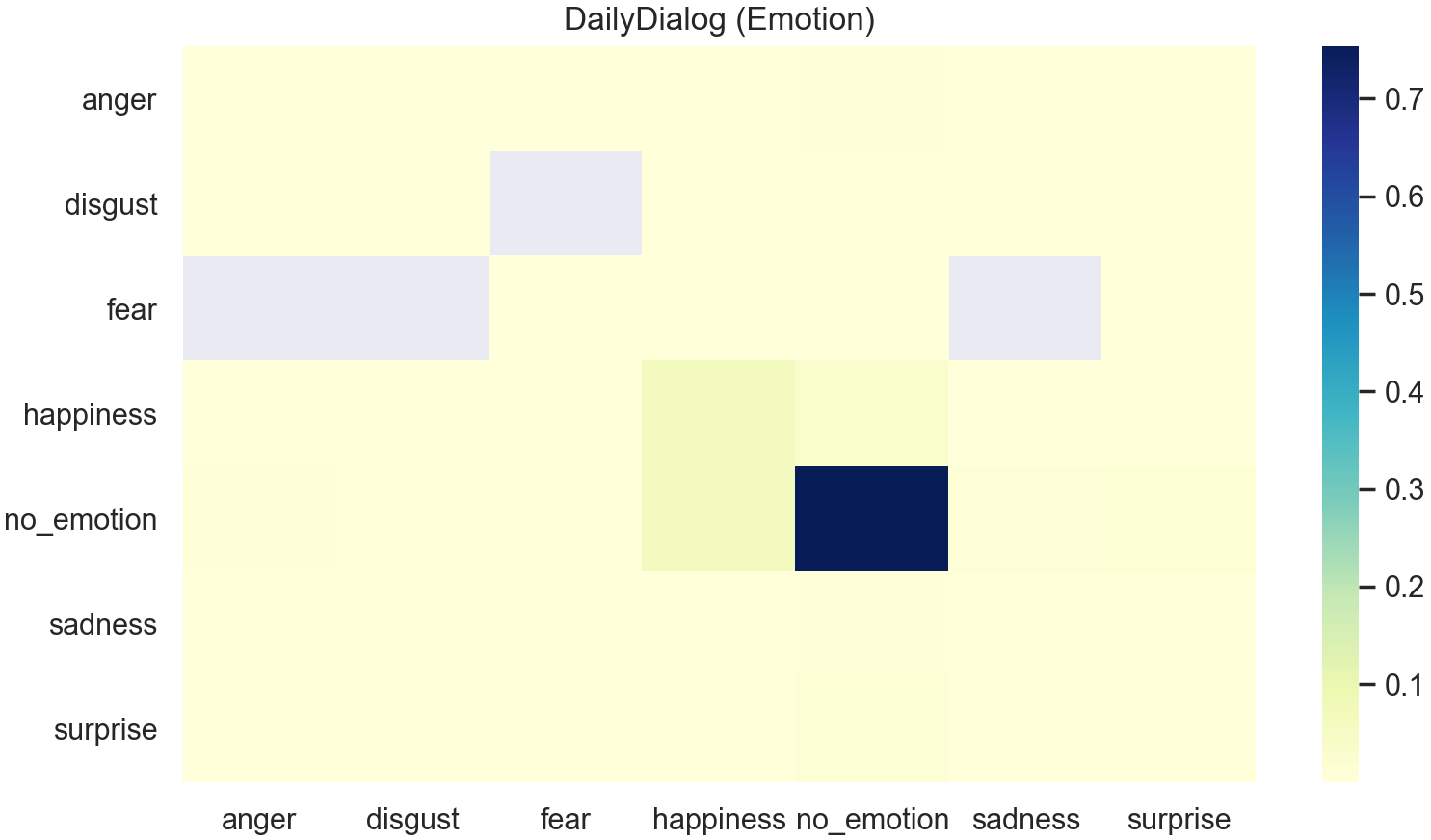}
    \label{fig:dd-inter-w/-neutral}
    \end{subfigure}
     \caption{The heatmap of intra-speaker (left) and inter-speaker (right) label transition statistics in the DailyDialog dataset including \emph{neutral} emotion. The color bar represents normalized number of inter-speaker and intra-speaker transitions such that elements of each matrix add up to 1.}
     \label{fig:heatmap3}
\end{figure*}

\textbf{MultiWOZ} \citep{budzianowski2018multiwoz} or Multi-Domain Wizard-of-Oz dataset is a fully-labeled collection of human-human written conversations spanning over multiple domains and topics. The dataset has been created for task-oriented dialogue modelling and has 10,000 dialogues, which is at-least an order bigger than previously available task-oriented corpora. The dialogues are labelled with belief states and actions. It contains conversations between an user and a system from the following seven domains: restaurant, hotel, attraction, taxi, train, hospital and police. In this work we focus on classifying the intent of the utterances from the user which belong to one of the following categories: \textit{book restaurant}, \textit{book train}, \textit{find restaurant}, \textit{find train}, \textit{find attraction}, \textit{find bus}, \textit{find hospital}, \textit{find hotel}, \textit{find police}, \textit{find taxi}, and \textit{None}. The \textit{None} utterances are not included in evaluation. Note that, utterances from the system side are not labelled and thus are not classified in our framework.

\textbf{Persuasion For Good} \citep{wang2019persuasion} dataset is a persuasive dialogue dataset where one participant aims to persuade the other participant to donate his/her earning using different persuasion strategies. The two participants are denoted as \textit{Persuader} aka \textbf{ER} and \textit{Persuadee} aka \textbf{EE} respectively. In this work, we formulate our problem to classify the utterances of Persuader and Persuadee separately using the full context of the conversation. The Persuader strategies are to be classified into the following eleven categories: \textit{donation-information}, \textit{logical-appeal}, \textit{personal-story}, \textit{emotion-appeal}, \textit{personal-related-inquiry}, \textit{foot-in-the-door}, \textit{source-related-inquiry}, \textit{task-related-inquiry}, \textit{credibility-appeal}, \textit{self-modeling}, and \textit{non-strategy-dialogue-acts}. In contrast, the Persuadee strategy can belong to one of the following thirteen categories: \textit{disagree-donation-more}, \textit{ask-org-info}, \textit{agree-donation}, \textit{provide-donation-amount}, \textit{personal-related-inquiry}, \textit{disagree-donation}, \textit{task-related-inquiry}, \textit{negative-reaction-to-donation}, \textit{ask-donation-procedure}, \textit{positive-reaction-to-donation}, \textit{neutral-reaction-to-donation}, \textit{ask-persuader-donation-intention}, and \textit{other-dialogue-acts}.

The dataset consists of 1017 dialogues, in which 300 dialogues are annotated with persuasion strategies. In this work, we use the annotated dialogues and partition them into train (220 dialogues), validation (40 dialogues), and test (40 dialogues) split to conduct our experiments.

\textbf{Statistics} Some statistics about the number of dialogues and utterances in the four datasets are presented in \cref{table:data}.

\paragraph{Label Transitions.}
To check whether there lies any patterns in the label sequences of the datasets, in \cref{fig:heatmap1} and \ref{fig:heatmap2}, we plot frequency of the label pairs $(x,y)$ where $x$ and $y$ are the labels of $U_{s_{t-1}, t-1}$ and $U_{s_t, t}$ respectively. Figure \cref{fig:heatmap1} explains inter-speaker label transition and \cref{fig:heatmap2} illustrates the intra-speaker label transition. Both these plots reveal the same emotion labels appearing in the consecutive utterances with high frequency in the IEMOCAP dataset. This induces label dependencies and consistencies and can be called as the \textbf{label copying feature} of the dataset. From our empirical analysis in \cref{sec:lca}, we confirm this property of the IEMOCAP dataset. Although not as strong as IEMOCAP, the intra-speaker label copying feature is also prevalent in the MultiWOZ and DailyDialog (Act) dataset (refer to \cref{fig:heatmap1}). Moreover, we observe interesting patterns in DailyDialog (Act). A directive utterance is commonly followed by a commissive utterance. This indicates that utterances with acts such as request and instruct (directive label) are followed by accepting/rejecting the request or order (commissive label). We also notice that an utterance with the act of questioning is commonly followed by the utterances with the act of answering (which is quite natural). \cref{fig:heatmap1} also corroborates the high frequent joint appearance of similar emotions in both speaker's utterances e.g., negative emotions --- \emph{anger, frustration, sad} expressed by one speaker is replied with a similar negative emotion by the other speaker. Interestingly, the DailyDialog dataset for emotion classification does not elicit any such patterns. We can attribute this to the scripted utterances present in the IEMOCAP that has specifically been designed to invoke more emotional content to the utterances. On the other hand, the DailyDialog dataset comprises naturalistic utterances that are more dynamic in nature as they depend on interlocutors' personality. In both IEMOCAP and DailyDialog datasets, the repetitions of the same emotions can be found in consecutive utterances of a speaker. The repetition of the same or similar emotions for a speaker is frequent and often forms long chains in IEMOCAP. However, such repetitions are much less prevalent in DailyDialog. Readers are referred to \cref{fig:heatmap2} for a clearer view. \textbf{This two different types of datasets used in this work is purposefully crafted in order to study dataset-specific nuances to attempt the same task.} In DailyDialog, approximately 80\% of utterances are labeled as \emph{no-emotion} (see \cref{fig:heatmap3}) which poses a difficult challenge to perform emotion classification. These two datasets also differ from each other in the average dialogue length. While the average number of utterances per dialogue in the IEMOCAP dataset is more than 50, the average number of utterances per dialogue in the DailyDialog dataset is just 8 which is much shorter.

Among other semantically plausible label transitions, we can see in \cref{fig:heatmap2}, the intent \emph{book restaurant} to be frequently followed by the intent \emph{find taxi} in the MultiWOZ dataset. \textbf{We believe this is potentially one of the reasons why contextual models perform so well on these three datasets and tasks} compared to the rest which we discuss in the subsequent sections. Further, label dependency and consistency can aid filtering likely labels given the prior labels. Notably, such patterns are not visible in the other datasets. Hence, one can use Conditional Random Field (CRF) to find any hidden label patterns and dependencies.

\subsection{Evaluation Metrics}
In our experiments, we use evaluation metrics as specified in \cref{table:data}. Weighted average (W-Avg) F1 score is the main metric in IEMOCAP emotion and MultiWOZ intent classification. However, we report this metric in all the other tasks as well.
For the other tasks - DailyDialog emotion, DailyDialog act, Persuader and Persuadee strategy classification, the label distribution is highly imbalanced. Hence we also report Macro F1 scores. In DailyDialog emotion classification, neutral labels are excluded (masked) while calculating the metrics. However, this utterances are still passed in the input of the different recurrent models. In DailyDialog emotion classification, we also report Micro-F1 scores.

\subsection{Training Setup}
We use the 300 dimensional pretrained 840B GloVe vectors in our CNN based feature extractor. The GloVe pretrained embeddings are kept fixed throughout the training process. For RoBERTa based feature extractor, we use the pretrained RoBERTa-Base model which is fine-tuned during training. In our bcLSTM and DialogueRNN models, we use LSTM and GRU hidden sizes of $D_h$ (100/200/1024 in \cref{table:params}) with forward and backward state concatenation. The GloVe based models are trained with a learning rate of 1e-3 and a batch size of 32 dialogues with the Adam optimizer. The RoBERTa based models are trained with a learning rate of 1e-5 and a batch size of 4 dialogues with the AdamW optimizer. All models are trained for 100 epochs. A chart of used hyperparameters is shown in \cref{table:params}.

\begin{table*}[ht!]
\centering
        \begin{tabular}{l|c|c|c|c|c|c|c|c}
            \toprule
            Models &\multicolumn{3}{c|}{bcLSTM}&\multicolumn{5}{c}{DialogueRNN}\\
            & $D_h$ & $lr$ & $bs$ & $D_g$ & $D_p$ & $D_e$ & $lr$ & $bs$ \\
            \hline
             GloVe All Datasets & 100 & 1e-3 & 32 & 100 & 100 & 100 & 1e-3 & 32 \\
             RoBERTa All Datasets \emph{except} IEMOCAP & 200 & 1e-5 & 4 & 200 & 200 & 200 & 1e-5 & 4 \\
             RoBERTa IEMOCAP & 1024 & 1e-5 & 4 & 1024  & 1024  & 1024 & 1e-5 & 4  \\
            \bottomrule
        \end{tabular}
    \caption{Hyperparameter details of the experiments. Note: lr$ \rightarrow$Learning rate; bs$\rightarrow${Batch size}.}
    \label{table:params}
\end{table*}

\section{Results}
\begin{table*}[ht]
  \centering
 \resizebox{\linewidth}{!}{
   \begin{tabular}{l||c||ccc|cc}
    \toprule
    \multirow{4}{*}{Methods} & \textbf{IEMOCAP} & & \multicolumn{4}{c}{\textbf{DailyDialog}} \\
     & Emotion & \multicolumn{3}{c|}{Emotion} &  \multicolumn{2}{c}{Act}\\
     
     \cline{2-7} & W-Avg F1 & W-Avg F1& Micro F1 & Macro F1 & W-Avg F1 & Macro F1\\
\hline
    GloVe CNN & 52.04 & 49.36 & 50.32 & 36.87 & 80.71 & 72.07 \\
    GloVe cLSTM & 59.10 & 52.56 &53.67 &38.14 & 83.90 & 78.89 \\
    \quad  \footnotesize{w/o Residual} & 55.07  & 52.56 & 53.26 & 38.12 & 84.06 &  78.54 \\
    GloVe bcLSTM & 61.74 & 52.77 & 53.85 & 39.27 & 84.62 & 79.12 \\
    \quad  \footnotesize{w/o Residual} & 58.32  & 54.74 & \textbf{56.32} & 39.24 & 84.10 &  78.98 \\
    GloVe DialogueRNN & \textbf{62.57} & \textbf{55.18} & 55.95 & \textbf{41.80} & \textbf{84.71} & \textbf{79.60}   \\
     \quad  \footnotesize{w/o Residual} & 61.32 & 54.50 & 55.29 & 40.05 & 83.98 &  79.16\\
    \hline
    RoBERTa LogReg & 54.12 & 52.63 & 52.42 & 40.02 & 82.55 & 75.62 \\
    RoBERTa bcLSTM & 62.72 & 56.05 & 56.77 & 43.26 & 85.17 & 82.16  \\
    \quad  \footnotesize{w/o Residual} & 62.86 & 55.92 & 57.32 & 43.03 & \textbf{86.35} & 80.69\\
    RoBERTa DialogueRNN & \textbf{64.12} & \textbf{59.07} & \textbf{59.50} & \textbf{45.19} & 86.31 & \textbf{82.20} \\
    \quad  \footnotesize{w/o Residual} & 63.96 & 57.57 & 57.76 & 44.25 & 86.28 & 82.08\\
    \bottomrule
   \end{tabular}
   }
  \caption{Classification performance in test data for emotion prediction in IEMOCAP, emotion prediction in DailyDialog, and act prediction in DailyDialog. Scores of the Glove-based models are reported after averaging 20 different runs. RoBERTa-based models were run 5 times and we report the average scores. Test F1 scores are calculated at best validation F1 scores.}
  \label{table:result1}
\end{table*}

\begin{table*}[ht]
  \centering
   \begin{tabular}{l||c||cc|cc}
    \toprule
    \multirow{3}{*}{Methods} & \textbf{MultiWOZ} & \multicolumn{4}{c}{\textbf{Persuasion For Good}} \\
     & Intent & \multicolumn{2}{c|}{Persuader} &  \multicolumn{2}{c}{Persuadee}\\
     
     \cline{2-6} & W-Avg F1 & W-Avg F1 & Macro F1 & W-Avg F1 & Macro F1\\
     \hline
    GloVe CNN & 84.30 & 67.15 & 54.33 & 58.00 & 41.03 \\
     GloVe cLSTM & 95.03 & 68.75  & 54.36 & 59.46 & 41.62\\
    \quad  \footnotesize{w/o Residual} & 95.12 &64.62 & 49.08 & 54.87 & 36.36 \\
    GloVe bcLSTM & 96.14 & \textbf{69.26}  & 55.27 & \textbf{61.18} & \textbf{42.19}\\
    \quad  \footnotesize{w/o Residual} & 96.21 &67.20 & 52.75 & 55.02 & 37.72 \\
    GloVe DialogueRNN & \textbf{96.32} & 68.96  & \textbf{56.29}  & 61.11  & 42.18\\
    \quad  \footnotesize{w/o Residual} & 96.08 & 68.77& 54.20 & 58.72 & 39.06 \\
    \hline
    RoBERTa LogReg & 85.70 & 71.98 & 60.36 & 63.45 & \textbf{51.74} \\
    RoBERTa bcLSTM & 95.46 & 71.85 & 61.05 & 64.14 & 50.11 \\
    \quad  \footnotesize{w/o Residual} & \textbf{95.61} & 71.06 & 58.72 & 62.73 & 44.74 \\
    RoBERTa DialogueRNN & \textbf{95.61} & \textbf{72.91} & \textbf{62.03} & \textbf{64.33} & 49.22\\
    \quad  \footnotesize{w/o Residual} & 95.29 & 72.45 & 60.49 & 64.21 & 49.71\\
    \bottomrule
   \end{tabular}
  \caption{Classification performance in test data for intent prediction in MultiWOZ, persuader and persuadee strategy prediction in Persuasion for Good. Scores of the Glove-based models are reported after averaging 20 different runs. RoBERTa-based models were run 5 times and we report the average scores. Test F1 scores are calculated at best validation F1 scores.}
  \label{table:result2}
\end{table*}
We report results for IEMOCAP, DailyDialog dataset in \cref{table:result1} and MultiWOZ, Persuasion for Good dataset in \cref{table:result2}. We ran each experiment multiple times and report the average test scores based on the best validation scores.

We observe that there is a general trend of improvement in performance when moving to the RoBERTa based feature extractor from the GloVe CNN feature extractor except in the intent prediction task in MultiWOZ dataset. As the RoBERTa model has been pre-trained on a large amount of textual data and has considerably more parameters, this improvement is expected. The results could possibly be improved even more if a RoBERTa-Large model is used instead of the RoBERTa-Base model that we use in this work.

We also observe that contextual models --- bcLSTM and DialogueRNN perform much better than the non-contextual Logistic Regression models in most cases. Context information is crucial for emotion, act, and intent classification and models such as bcLSTM or DialogueRNN are some of the most prominent methods to model the contextual dependency between utterances and their labels. In IEMOCAP, DailyDialog and MultiWOZ there is a sharp improvement in performance in contextual models compared to the non-contextual models. However, for the strategy classification task in \emph{Persuasion for Good dataset}, the improvement in contextual models is relatively lesser. Notably, for Persuadee classification, the RoBERTa non-contextual model achieves the best result, outperforming the contextual models. Without the presence of residual connections, the \emph{GloVe cLSTM} and \emph{GloVe bcLSTM} baselines perform poorly than the non-contextual \emph{GloVe CNN} baseline in the \emph{Persuasion for Good dataset}. This beckons the need for better contextual models for this dataset. To analyze the results of the different models we look at the following aspects:

\paragraph{Importance of the Residual Connections in the Models.} It is also to be noted that the introduction of the residual connections generally improves the performance of the contextual models. We obtain better performance and improved stability during training for most of the models with residual connections.
In particular, residual connections are mostly effective in IEMOCAP and Persuasion for Good datasets that comprise long dialogues. Residual connections are used in deep networks to aid information propagation and tackle vanishing gradient problems~\cite{wu2006emotion,kim2017residual} in RNNs by improving gradient flow. As multi-layered RNN-like architectures often find it difficult to model long-range dependencies in a sequence due to vanishing gradient problems~\cite{pascanu2013difficulty}, we conjecture, that could be one of the reasons why we see a great performance boost with residual connections by helping propagate key information form the CNN layers to the output of LSTM layers that might be lost due to the long deep sequence modeling in the LSTM layer. Residual connections also help in combating vanishing gradient issues by improving gradient flow.
Unlike IEMOCAP and Persuasion for Good, in DailyDialog and MultiWOZ datasets, the improvement in performance caused by the residual connections is only little which can be attributed to the relatively shorter dialogues present in these two datasets.

\paragraph{Importance of Bidirectionality in the Models.}
The use of bidirectionality without the presence of residual connection (as shown in \cref{fig:residual}) in bcLSTM improves the score by 1-3\% (refer to \cref{table:result1} and \cref{table:result2}) across all the datasets. Contrary to IEMOCAP,
we found that bidirectionality is less useful for emotion classification in DailyDialog dataset which comprises of very short dialogues. The difference in performance between cLSTM and bcLSTM indicates the importance and need of bidirectionality in the models for capturing contextual information from the future utterances while classifying utterance $U_t$. To have a better idea about this difference, one should compare the cLSTM in \cref{table:result1} and \cref{table:result2} with row no. 33 in \cref{tab:cc}. Although the underneath model architecture in these two settings is different, they are definitely comparable as both attempts to measure the importance of future utterances while classifying utterance at time $t$. From these tables, a conclusive trend is found --- future utterances are very important in the IEMOCAP dataset but not much in the rest of the datasets. We conjecture the long dialogues and the inter-utterance label dependency and consistency in the IEMOCAP dataset is the predominant factor for this observation. In \cref{sec:lca}, we empirically confirm the existence of this label dependence and consistence in the IEMOCAP dataset.

\paragraph{Does the Use of Conversational Context Help?}
The bcLSTM and DialogueRNN architecture have been primarily developed as emotion/sentiment classifiers in utterance level dialogue classification tasks. To explain the performance of these models across a diverse range of dialogue classification tasks, it is important to understand the nature of the task itself. In emotion classification, labels are largely dependent on other contextual utterances. In IEMOCAP, except for some cases of emotion shift (sudden change from positive group of emotion to negative group or vice versa) labels are correlated and often appears in continuation. bcLSTM and DialogueRNN can model this inter-relation between utterances and perform much better than non-contextual models. For intent classification in MultiWOZ, the label dependency between utterances is even stronger. MultiWOZ is a task oriented dialogue dataset between a user and a system. When the user has some query or some information need, all utterances tend to have the same intent label until the query gets resolved. Hence, all contextual models can perform this task relatively easily (evident from the very high F1 scores).
We also observe this trend in DailyDialog emotion and act classification tasks. In both of these tasks, contextual information is helpful and provides a significant improvement in performance. However, in Persuasion for Good dataset, the introduction of context is much less helpful.
The persuader and persuadee strategy labels are relatively more difficult to model even with contextual information.

\paragraph{Can Speaker-specific Model like DialogueRNN Help in Varied Dialogue Classification Tasks?}
DialogueRNN is a speaker specific model which distinguishes different speaker in a conversation by keeping track of individual speaker states. Identifying speaker level context is fundamental for the task of emotion recognition in conversations \citep{ghosal2019dialoguegcn,zhang2019modeling}. It is thus expected that DialogueRNN would perform better than non-contextual or bcLSTM models in this task as evidenced in \cref{table:result1}. Additionally DialogueRNN produces the best results or close to the best results in the other tasks as well. In intent classification or persuasion strategy classification, we only classify utterances coming from only one of the parties (user party in MultiWOZ intent, persuader or persuadee party in persuasion strategy classification). This might explain why the performance of DialogueRNN is occasionally lower than bcLSTM in some of these tasks. But, overall it can be concluded that speaker specific modelling is indeed important and helps in various dialogue classification tasks.

\paragraph{Variance in the Results.} As deep learning models tend to yield varying results across multiple training runs, we trained each model multiple times and report the average score in \cref{table:result1} and \cref{table:result2}. In general, we observed that the RoBERTa-based models show lesser variance compared to the GloVe-based models.

\textbf{Variance in the Glove-based models:} The observed variance is higher for emotion classification in IEMOCAP and DailyDialog as compared to act and intent classification in DailyDialog and MultiWOZ, respectively. Both baseline models -- Glove CNN and bcLSTM show standard deviation of about 1.28\% in the IEMOCAP dataset across different runs. In the Persuasion for Good dataset, for both persuader's and persuadee's act classification tasks, the deviation remains around 1.6\% when we consider the Macro-F1 metric. However, for the Weighted-F1 metric, the performance is relatively stable as upon accumulating multiple runs the standard deviation is about 0.99\% across the baselines. A similar trend is also prevalent in the DailyDialog dataset for emotion classification. In this task, the baselines -- Glove CNN and bcLSTM show standard deviation of about 1.19\% when Weighted-F1 and Micro-f1 are considered. According to Macro-F1 metric, however, these baselines are exposed to relatively higher standard deviation of 2.88\%. This is likely to be a consequence of severe label imbalance in the dataset, that is having 80\% neutral utterances. We have observed that a majority of these neutral samples do not exhibit neutral emotion. Therefore, this poor labeling quality may have precipitated this large variance in the results. On the other hand, the baseline models perform consistently in the intent and act classification tasks in MultiWOZ and DailyDialog datasets respectively showing standard deviation of around 0.55\% across different runs. When comparing among the baselines, we found higher variances in the results obtained with the Glove CNN than the bcLSTM.

\begin{figure*}[t]
    \centering
     \includegraphics[width=\linewidth]{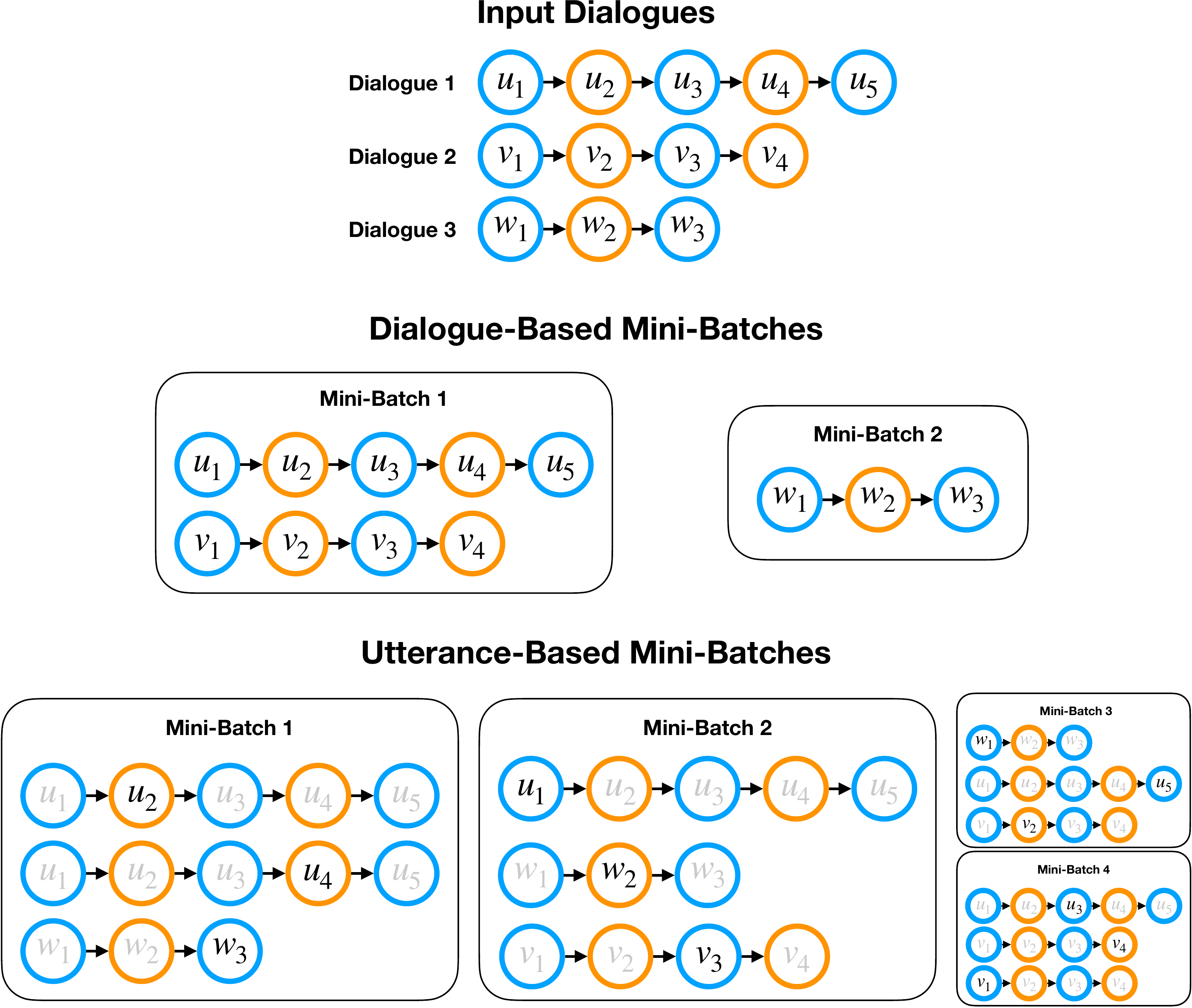}
     \caption{Two different types of mini-batch construction from the input dialogues. Faded utterances in \emph{utterance-based mini-batches} indicate context utterances, which are not present in the calculation of loss.}
     \label{fig:batch-split}
\end{figure*}

\textbf{One possible reason behind the variances in the results of the GloVe-based models could be the end-to-end training setup that renders the model deeper.} The original bcLSTM and DialogueRNN model employed a two-stage training method where the utterance feature extractor is first pretrained and then kept unchanged during the contextual model training. This setting may make those original models more stable. Similarly, we think, in our end-to-end setup, a more sophisticated training regime could result in a lesser variance of the results. For example, the utterance feature extractor could be trained only for the first few epochs and then kept frozen during subsequent epochs of the training. Due to this high variance in the end-to-end Glove-based models, the future works on these datasets and tasks which employ this setting should report the average results of multiple runs for a fair comparison of the models.

\textbf{Variance in the RoBERTa-based models:} The RoBERTa based models show much lesser variance in performance across different runs. In particular, the standard deviations in the results of Roberta-based bcLSTM are 0.57 on the IEMOCAP, 0.08, and 0.48 in the DailyDialog for emotion and act classification tasks, respectively, 0.07 in the MultiWoz dataset, 0.9 and 1.04 in the Persuasion for Good dataset for persuader's and persuadee's act classification tasks respectively. RoBERTa-based DialogueRNN shows a similar trend.  We surmise that this is the case because the feature extractor's weights are initialized from a pretrained checkpoint. Thus, the feature extractor already provides meaningful features from the beginning of training and is not required to be trained from scratch, resulting in greater stability in the performance.

\subsection{Mini-Batch Formation Technique}
To understand the sensitivity of the training to utterance distribution across mini-batches, we consider two scenarios (\cref{fig:batch-split}):
\begin{itemize}
    \item Dialogue distribution across mini-batches,
    \item Utterance distribution across mini-batches.
\end{itemize}

\begin{figure*}[htbp]
    \centering
     \includegraphics[width=\linewidth]{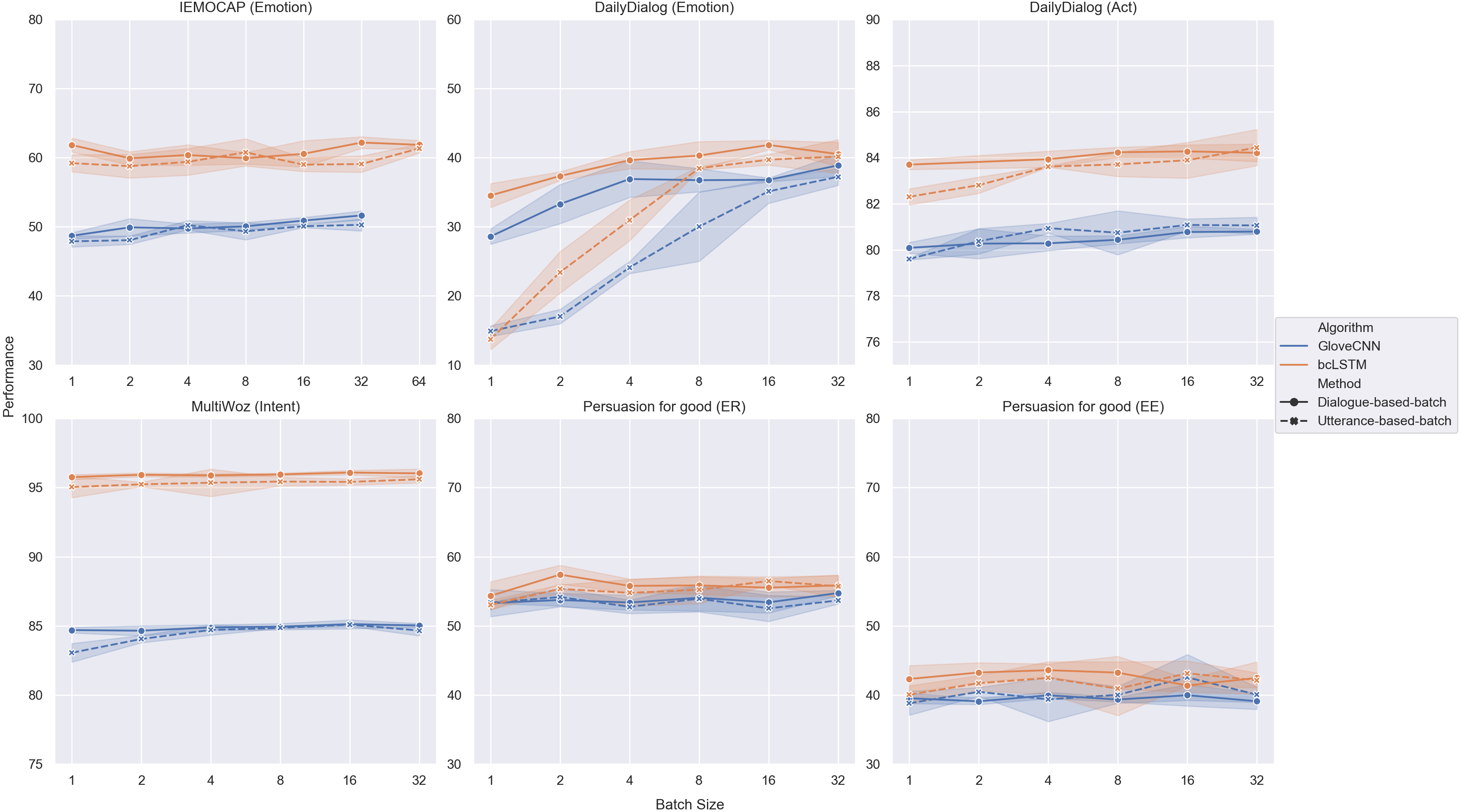}
     \caption{Sensitivity of GloVe bcLSTM and GloVe CNN to Utterance Distribution across Mini-Batches.}
     \label{fig:batch-split-bclstm-glovecnn}
\end{figure*}
The first scenario keeps all the constituent utterances of a dialogue in a single mini-batch that allows classifying them using a single RNN pass with $O(n)$ cost ($n$ is the number utterances in a dialogue)---this is the default mode of operation for bcLSTM and DialogueRNN. In contrast, the second scenario may distribute the target utterances to be classified across different mini-batches. This leads to one RNN pass per utterance, that is considerably computationally costlier than the previous scenario---$O(n^2)$. Further, the representation of a particular context utterance is likely to vary based on the mini-batch due to the update of network parameters by a fraction of their gradients per mini-batch. This curtails joint training of all the constituent utterances in a dialogue, precipitating poorer performance as evidenced by \cref{fig:batch-split-bclstm-glovecnn}.
We found when the batch size is small, baselines with dialogue level mini-batch perform better than their counterparts with utterance-level mini-batch configuration.
However, the difference gradually reduces with the increase in the batch size. In fact, in several datasets (as shown in ~\cref{fig:batch-split-bclstm-glovecnn}, with the batch size equal to or greater than 16, the performance difference between these two different configurations are not significant.

\paragraph{Key Consideration.}
When comparing two models tasked for utterance classification in dialogues, the mini-batch construction process should be paid careful attention. As we have explained above, the batch size is a crucial hyperparameter when juxtaposing a model with utterance-level mini-batch with another model with dialogue-level mini-batch. While we acknowledge the fact that most of the GloVe embedding-based models tend to be small and using a larger batch size in those models for utterance-level mini-batch construction is not a problem, we think this can be a colossal issue in contextualized word embedding-based models such as BERT. Models like BERT have billions of parameters and using large batch sizes to train these models can be one of the bottlenecks. In such scenarios, we recommend using dialogue-level mini-batch which can still perform decently with relatively smaller batch sizes as can be seen in \cref{fig:batch-split-bclstm-glovecnn} and be computationally inexpensive compared to utterance-level mini-batch.

The experiments conducted in \cref{sec:analysis} are based on \textit{utterance-based mini-batch} training. We make different contextual modifications for each target utterance in a dialogue. Hence, \textit{utterance-based} training and evaluation is necessary.

\section{Analysis}
\label{sec:analysis}
We set up different scenarios to probe the GloVe bcLSTM and GloVe CNN baselines because these two are conceptually much more straightforward than DialogueRNN. For example, in addition to context, DialogueRNN also tracks the speaker states based on the utterances. Thus, the perturbations in the input would influence speaker modeling along with the context modeling. This may result in more complex deviations than bcLSTM, which are more difficult to analyze. Simple models are likely to be more interpretable. E.g., owing to DialogueRNN's complexity, we need to perform different levels of ablation studies, such as speaker GRU removal, listener state's addition, and removal, to explain its behavior.

We stick to GloVe word embeddings as RoBERTa-based word embeddings are trained using the masked language model (MLM) objective that is already very powerful in modeling cross sentential context representation as demonstrated by other works~\cite{liu2019roberta,lewis2019bart}. Hence, to conduct a fair comparison between non-contextual and contextual models and further, for an easier apprehension on the role of contextual information in utterance-level dialogue understanding, we resort to the GloVe CNN model. Additionally, GloVe CNN is computationally more efficient. In all the experiments explained from subsection \cref{sec:cc_drop} onward, we use the utterance-level mini-batch setting as it allows contextual baseline model --- bcLSTM to be flexible to context corruption and altering.
In the following subsections, we use GloVe bcLSTM and bcLSTM interchangeably.

For the analyses requiring training, we trained all the models at least 5 times and report their the mean test score. For the remaining analyses, we evaluated the test results using saved models at checkpoints of different runs and report the average scores. The trends in these results, delineated in the following sections, were found consistent at these checkpoints, i.e., although models in different runs yield varied performance on the original test data, they behave similarly to the same input perturbation. As most of the results presented in the following sections are obtained from the experiments conducted in the utterance-level mini-batch setting, there is a disparity between these tables with \cref{table:result1} and \ref{table:result2}. In utterance-level mini-batch setup, we obtain better performance in some of the tasks which is reported in the subsequent sections. The readers are recommended to refer to \cref{table:result1} and \ref{table:result2} for the baseline results which were obtained by averaging the outcomes of more than 20 runs. The rows in the following tables where no perturbation were applied to the inputs can be used as points of reference to analyze the effect of perturbations at the inputs.

\subsection{Classification in Shuffled Context}
\begin{table*}[ht]
  \centering
 \resizebox{\linewidth}{!}{
   \begin{tabular}{cccc||c||ccc|cc}
    \toprule
     \multicolumn{4}{c||}{\textbf{GloVe bcLSTM}} & \textbf{IEMOCAP} & & \multicolumn{4}{c}{\textbf{DailyDialog}} \\
     \multicolumn{4}{c||}{Context Shuffling Strategy} & Emotion & \multicolumn{3}{c|}{Emotion} & \multicolumn{2}{c}{Act}\\\cline{5-10}
     Train & Val & Test & OP & W-Avg F1 & W-Avg F1 & Micro F1 & Macro F1 & W-Avg F1 & Macro F1 \\
     \hline
    \xmark & \xmark & \xmark & \xmark & \textbf{61.74} & \textbf{52.77} & \textbf{53.8}5 & \textbf{39.27} & \textbf{84.62} & \textbf{79.46}\\
    \checkmark & \checkmark & \xmark & \xmark & 59.74 & 52.29 & 51.97 & 36.87 & 81.82 & 74.88 \\
    \xmark & \xmark & \checkmark & \xmark & 57.63 & 48.35 & 50.32 & 34.58 & 76.65 & 66.81 \\
    \checkmark & \checkmark & \checkmark & \xmark & 59.82 & 52.17 & 52.92 &  37.69 & 81.84 & 74.62 \\
    \checkmark & \checkmark & \checkmark & \checkmark & 59.47 & 49.16 & 51.67 & 32.53 & 81.29 & 73.83 \\
    
    \bottomrule
   \end{tabular}
   }
  \caption{Performance of GloVe bcLSTM models in IEMOCAP and DailyDialog for various shuffling strategies. In Train, Val, Test column \checkmark denotes shuffled context and \xmark denotes unchanged context. In OP column \checkmark denotes additional order prediction objective.}
  \label{tab:shuffle1}
\end{table*}

\begin{table*}[ht]
  \centering
 \resizebox{\linewidth}{!}{
   \begin{tabular}{cccc||c||cc|cc}
    \toprule
    \multicolumn{4}{c||}{\textbf{GloVe bcLSTM}} & \textbf{MultiWOZ} & \multicolumn{4}{c}{\textbf{Persuasion for good}} \\
    \multicolumn{4}{c||}{Context Shuffling Strategy} & Intent & \multicolumn{2}{c|}{Persuader} & \multicolumn{2}{c}{Persuadee}\\
    \cline{5-9}
     Train & Val & Test & OP & W-Avg F1 & W-Avg F1 & Macro F1 & W-Avg F1 & Macro F1 \\
     \hline
    \xmark & \xmark & \xmark & \xmark & \textbf{96.14} & \textbf{69.26} & \textbf{55.27} & \textbf{61.18} & \textbf{42.19}\\
    \checkmark & \checkmark & \xmark & \xmark & 91.34 & 68.06 & 54.91 & 59.27 & 41.52 \\
    \xmark & \xmark & \checkmark & \xmark & 67.91 & 65.30 & 50.69 & 55.07 & 37.17 \\
    \checkmark & \checkmark & \checkmark & \xmark & 90.78 & 66.32 & 53.60 & 58.46 & 40.96\\
    \checkmark & \checkmark & \checkmark & \checkmark & 90.67 & 67.60 & 53.50 & 58.69 & 41.62 \\
    
    \bottomrule
   \end{tabular}
   }
  \caption{Performance of GloVe bcLSTM models in MultiWOZ and Persuasion for good for various shuffling strategies. In Train, Val, Test column \checkmark denotes shuffled context and \xmark denotes unchanged context. In OP column \checkmark denotes additional order prediction objective.}
  \label{tab:shuffle2}
\end{table*}
To analyze the importance of context, we shuffle the utterance order of a dialogue and try to classify the correct label from the shuffled utterance sequence. For example, a dialogue having utterance sequence of \{$u_1, u_2, u_3, u_4, u_5$\} is shuffled to \{$u_5, u_1, u_4, u_2, u_3$\}. This shuffling is carried out randomly, resulting in an utterance sequence whose order is different from the original sequence. We design three such shuffling experiments: i) dialogues in train and validation sets are shuffled, dialogues in test set are kept unchanged, ii) dialogues in train and validation sets are kept unchanged, whereas dialogues in test set are shuffled, iii) dialogues in train, validation and test sets are all shuffled.

We analyze these shuffling strategies in the GloVe bcLSTM model. In theory, the recurrent nature of the bcLSTM model allows it to be capable of modelling contextual information from the very beginning of the utterance sequence to the very end. However, when classifying an utterance, the most crucial contextual information comes from the neighbouring utterance. In an altered utterance context, the model would find it difficult to predict the correct labels because the original neighbouring utterances may not be in immediate context after shuffling. This kind of perturbation would make the context modelling less efficient, and performance is likely to drop compared to their non-shuffled context counterparts. This is empirically shown in \cref{tab:shuffle1} and \cref{tab:shuffle2}.

We observe that, whenever there is some shuffling in train, validation, or test set, the performance decreases a few points in both the datasets across all the tasks and all the evaluation metrics. Notably, the performance drop is highest when the dialogues in train and validation sets are kept unchanged and dialogues in test set are shuffled.

\subsection{Classification in Shuffled Context with Order Prediction}
In some of the shuffling strategies, we enforce an additional utterance order prediction (OP) objective to see how it affects the results. We assume that if the network learns to predict how the original order of the utterances has been shuffled, then it may improve the main utterance level dialogue classification task as well. In this setup, the order prediction objective is realized through the same bcLSTM network and an additional fully connected layer with softmax activation on top. In the previous example, when \{$u_1, u_2, u_3, u_4, u_5$\} is shuffled to \{$u_5, u_1, u_4, u_2, u_3$\}, the additional objective is to predict the shuffled sequence order as class labels (in the new fully connected softmax layer). Here, the sequence order labels to be predicted are \{5, 1, 4, 2, 3\} respectively. The network is thus trained with utterance order prediction objective and the main classification objective (emotion, act, intent or persuasion strategy prediction) jointly.

We report results of this additional objective with the shuffling strategy iii) i.e., train, validation, and test set are all shuffled. Results are reported in \cref{tab:shuffle1} and \cref{tab:shuffle2}. In most of the tasks, the additional order prediction objective doesn't help. We only observe some improvement in performance in the strategy classification tasks in Persuasion for Good.

\subsection{Controlled Context Dropping}
\label{sec:cc_drop}

\begin{table*}[t!]
  \centering
  \small
 \resizebox{\linewidth}{!}{
   \begin{tabular}{l|c|c||c|c||cccccc}
    \toprule
    & \multicolumn{2}{c||}{\textbf{Train, Val}} & \multicolumn{2}{c||}{\textbf{Test}} & \textbf{IEMOCAP} & \textbf{Dailydialog}  & \textbf{Dailydialog} &\textbf{MultiWOZ}   & \textbf{Persuasion} &  \textbf{Persuasion}\\
    \multirow{2}{*}{\#}& \multirow{2}{*}{Past} & \multirow{2}{*}{Future} & \multirow{2}{*}{Past} & \multirow{2}{*}{Future} & Emotion & Emotion & Act & Intent & ER& EE\\
    &  & & & & W-Avg F1 & Macro F1 & Macro F1 & W-Avg F1 & Macro F1 & Macro F1\\
    \midrule
1 & -5 & --  & -5 & -- &  60.45 & 37.38 & 73.11 & 93.97 & 54.54 & 40.44 \\
2 & -5 & --  & -ALL & -- &  58.05 & 36.04 & 72.79 & 93.18 & 53.41 & 40.38 \\
3 & -5 & --  & -5 & -5 &  58.29 & 34.91 & 71.41 & 76.92 & 53.53 & 40.0 \\
4 & -5 & --  & -- & -5 &  58.65 & 34.78 & 70.54 & 78.09 & 53.37 & 40.1 \\
5 & -5 & --  & -- & -ALL &  50.94 & 35.05 & 70.83 & 85.72 & 51.18 & 38.11 \\
6 & -5 & --  & +5 & +5 &  55.92 & 37.63 & 72.92 & 94.06 & 53.12 & 39.09 \\
7 & -5 & --  & -- & -- &  60.64 & 37.09 & 72.7 & 94.51 & 54.62 & 41.56 \\
8 & -ALL & -- & -5 & -- &  57.77 & 36.67 & 73.09 & 93.37 & 52.68 & 38.68 \\
9 & -ALL & -- & -ALL & -- &  57.39 & 36.84 & 73.13 & 93.39 & 53.97 & 41.02 \\
10 &-ALL  & --  & -5 & -5 &  55.23 & 34.34 & 71.22 & 73.28 & 51.65 & 38.62 \\
11 &-ALL  & --  & -- & -5 &  55.8 & 33.73 & 71.27 & 75.6 & 51.43 & 38.89 \\
12 &-ALL  & --  & -- & -ALL &  45.43 & 33.72 & 71.5 & 83.11 & 46.83 & 40.17 \\
13 &-ALL  & --  & -- & +5 &  53.9 & 36.66 & 72.96 & 93.56 & 53.9 & 41.06 \\
14 &-ALL  & --  & -- & -- &  58.11 & 36.23 & 72.99 & 93.68 & 52.81 & 40.27 \\
15 & -5 &-5  &  -5 & -- &  59.89 & 34.13 & 72.65 & 88.66 & 53.94 & 39.66 \\
16 & -5 &-5  &  -ALL & -- &  55.34 & 33.13 & 72.3 & 84.76 & 51.49 & 38.58 \\
17 & -5 &-5  &  -5 & -5 &  59.31 & 36.28 & 72.67 & 88.51 & 53.73 & 40.34 \\
18 & -5 &-5  &  -- & -5 &  59.26 & 36.94 & 72.73 & 89.09 & 53.52 & 40.91 \\
19 & -5 &-5  &  -- & -ALL &  53.88 & 36.5 & 72.66 & 84.93 & 54.38 & 38.83 \\
20 & -5 &-5  &  +5 & +5 &  55.91 & 34.68 & 72.48 & 86.04 & 53.27 & 39.54 \\
21 & -5 &-5  &  -- & -- &  60.1 & 34.78 & 72.59 & 90.24 & 53.91 & 40.03 \\
22 & -- & -5 &  -5 & -- &  58.16 & 31.35 & 62.56 & 78.32 & 54.24 & 37.38 \\
23 & -- & -5 &  -ALL & -- &  52.45 & 31.3 & 60.51 & 81.57 & 55.81 & 37.77 \\
24 & -- & -5 &  -5 & -5 &  58.14 & 33.21 & 62.62 & 78.66 & 54.73 & 38.82 \\
25 & -- & -5 &  -- & -5 &  60.79 & 38.54 & 79.11 & 95.13 & 55.57 & 40.34 \\
26 & -- & -5 &  -- & -ALL &  54.41 & 39.34 & 79.08 & 95.1 & 55.59 & 42.79 \\
27 & -- & -5 &  +5 & +5 &  56.23 & 37.5 & 78.92 & 94.81 & 55.73 & 39.45 \\
28 & -- & -5 &  -- & -- &  60.86 & 37.1 & 79.08 & 95.11 & 55.35 & 40.25 \\
29 & -- & -ALL  &  -5 & -- &  53.41 & 33.82 & 62.41 & 77.66 & 53.05 & 39.05 \\
30 & -- & -ALL  &  -ALL & -- &  44.34 & 32.74 & 60.47 & 79.51 & 53.43 & 40.63 \\
31 & -- & -ALL  &  -5 & -5 &  53.67 & 34.43 & 62.35 & 77.16 & 52.91 & 39.05 \\
32 & -- & -ALL  &  -- & -5 &  56.58 & 40.22 & 78.35 & 95.06 & 55.37 & 41.4 \\
33 & -- & -ALL  &  -- & -ALL & 56.66  & 40.02 & 78.42 & 95.09 & 56.0 & 42.41 \\
34 & -- & -ALL  &  +5 & +5 &  53.46 & 39.96 & 78.43 & 94.75 & 55.61 & 41.83 \\
35 & -- & -ALL  &  -- & -- &  56.61 & 40.63 & 78.4 & 95.09 & 56.01 & 41.65 \\
36 & +5 & +5  &  -5 & -- &  57.62 & 34.58 & 63.44 & 88.5 & 53.63 & 41.81 \\
37 & +5 & +5  &  -ALL & -- &  53.41 & 33.95 & 60.95 & 90.35 & 54.42 & 44.14 \\
38 & +5 & +5  &  -5 & -5 &  56.77 & 33.54 & 62.64 & 73.27 & 52.66 & 39.23 \\
39 & +5 & +5  &  -- & -5 &  58.73 & 38.43 & 78.21 & 90.03 & 55.02 & 44.35 \\
40 & +5 & +5  &  -- & -ALL &  55.03 & 38.4 & 77.68 & 93.01 & 51.26 & 41.51 \\
41 & +5 & +5  &  +5 & +5 &  58.62 & 39.94 & 79.56 & 95.86 & 56.06 & 45.46 \\
42 & +5 & +5  &  -- & -- &  60.01 & 39.93 & 79.71 & 95.87 & 55.21 & 45.72 \\
43 & -- & --  &  -5 & -- &  59.17 & 36.27 & 63.24 & 87.55 & 53.77 & 43.19 \\
44 & -- & --  &  -ALL & -- &  53.86 & 35.69 & 60.97 & 89.59 & 54.49 & 44.29 \\
45 & -- & --  &  -5 & -5 &  57.64 & 34.79 & 62.37 & 71.09 & 54.0 & 39.68 \\
46 & -- & --  &  -- & -5 &  59.5 & 39.6 & 77.44 & 89.82 & 55.46 & 43.85 \\
47 & -- & --  &  -- & -ALL &  52.56 & 39.37 & 76.9 & 93.47 & 54.84 & 42.99 \\
48 & -- & --  &  +5 & +5 &  57.31 & 40.74 & 78.87 & 95.81 & 54.79 & 45.25 \\
49 & -- & --  &  -- & -- &  61.9 & 41.16 & 79.46 & 96.22 & 56.28 & 44.83 \\
    \bottomrule
   \end{tabular}
   }
  \caption{Results for controlled context dropping experiments in different settings. ER and EE denote Persuader and Persuadee strategy classification, respectively. In past (future) columns, \textbf{-5} $\implies$ dropping immediate 5 utterances from the past (future), \textbf{-All} $\implies$ dropping all utterances from the past (future), \textbf{+5} $\implies$ keeping only the immediate 5 utterances from the past (future), \textbf{--} $\implies$ keeping all utterances from the past (future). Scores are W-Avg F1 in IEMOCAP Emotion and MultiWOZ Intent; Macro F1 in the rest.}
  \label{tab:cc}
\end{table*}

In \cref{table:result1} and \cref{table:result2} we observe a large improvement in performance in the contextual models (bcLSTM, DialogueRNN) compared to the non-contextual CNN models. Now, we intend to analyze why this improved performance is observed and how recurrent models such as bcLSTM uses contextual dialogue information from past and future utterances effectively.

\begin{figure*}[t]
    \centering
    \includegraphics[width=\linewidth]{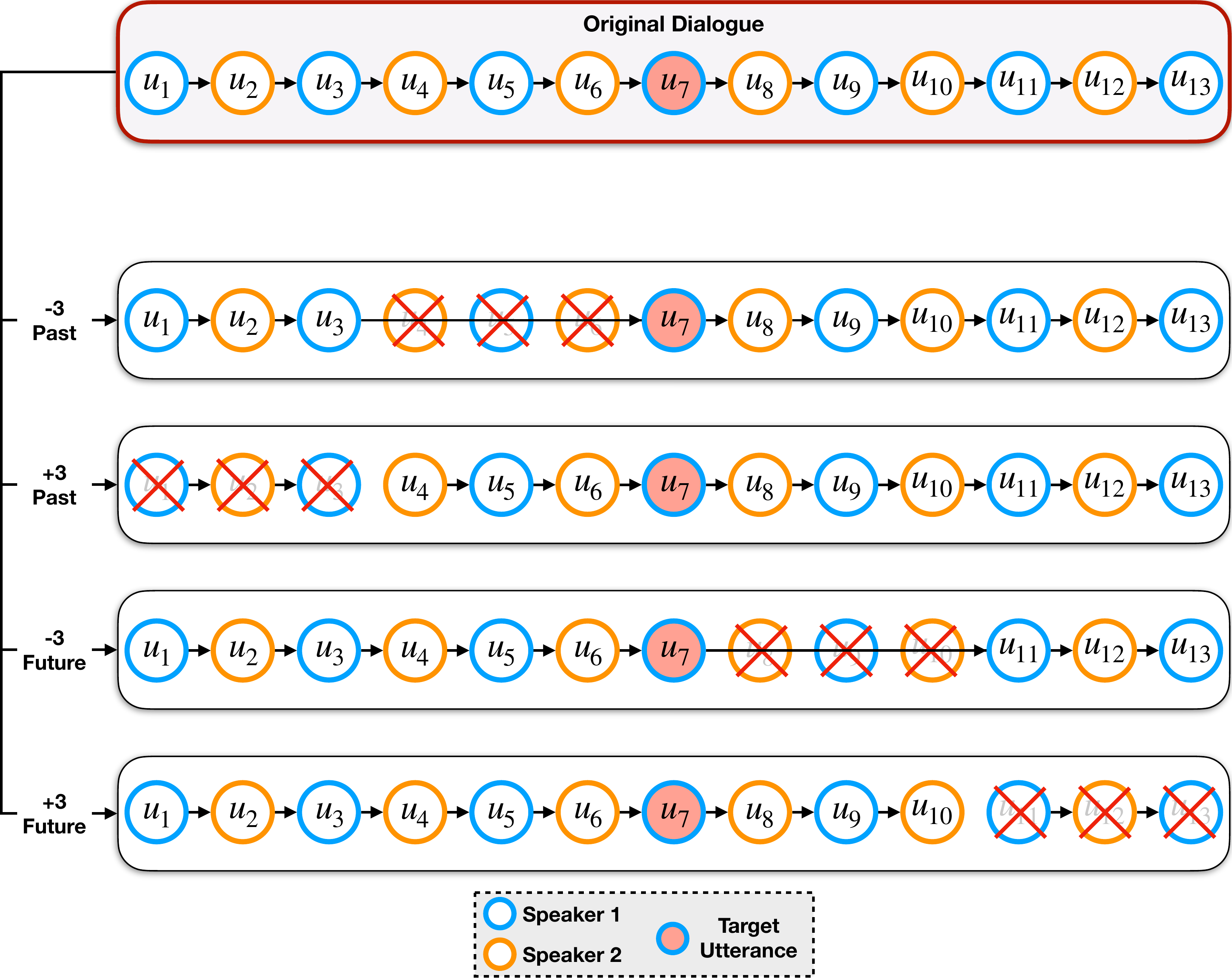}
    \caption{Illustration of controlled context drops of three utterances.}
    \label{fig:context-drop}
\end{figure*}

To understand this effect, we make a comprehensive analysis of the GloVe bcLSTM model and design an experimental study with controlled context dropping. In the default setting of bcLSTM, the full dialogue history and future is available to the model. Now, through a number of different experimental settings, we vary and limit the contextual information that is available to the model and study how the results are affected by it. Please refer to \cref{fig:context-drop} for a visual representation of the \emph{controlled context dropping} method.

\paragraph{Method.} For each target utterance $u_t$ in a dialogue, we control the contextual information available to it in as following:
\begin{itemize}
    \item We control the availability of contextual utterances from the past, or the future, or both.
    \item While we are controlling the availability of only past utterances, we vary it in one of the following ways:
    \begin{itemize}
        \item Dropping the 5 previous utterances. In other words, access only to $u_0, \dots , u_{t-6}$.
        \item Dropping all previous utterances.
        \item Dropping all previous utterances except the previous 5. In other words, access only to $u_{t-5}, \dots , u_{t-1}$.
    \end{itemize}
    \item Similarly, we modify the utterance access while controlling future, or both past and future.
    \item We vary the control over past, future utterances in different combinations during training (train, val set) and evaluation (test set).
\end{itemize}

\paragraph{Observations.}
We report the results for controlled context dropping experiments in \cref{tab:cc}. We observe that long distance context is much more important for emotion classification in IEMOCAP compared to the rest of the tasks. Row 48 in \cref{tab:cc} refers to the experimental setting where full conversational context is used during training, but only 5 contextual utterances from the past and future is used in the test set during evaluation. In this setting, the performance drop in IEMOCAP emotion classification is significantly higher than the other tasks indicating the importance of long distant contextual utterances.

Dropping all the future utterances further worsens the results in all the tasks. The reduction in performance is most significant in IEMOCAP (Row 47). We also observe some interesting results for the setting of training on full context and evaluating by removing all the past context (Row 44) or removing all the future context (Row 47). For DailyDialog emotion and act classification, there is a significant difference in performance in these two kinds of context control settings. Completely removing the past context results in much poorer performance signifying the importance of contextual information flow from past utterances in DailyDialog.

The configuration in row 12 removes all the past utterances and keeps all future utterances during training. However, during evaluation, all future utterances are discarded and predictions are based only on past utterances. Row 30 is also based on a similar setup where training is performed only on past utterances but evaluation is based only on future utterances. This contextual disparity during training and evaluation causes the bcLSTM model to perform very poorly across all the tasks.

The Persuasion for Good dataset contains dialogues with large number of utterances. However, we found in our experiments that a window size of 5 contextual utterances is generally sufficient in producing good results (\cref{tab:cc} Row 41). From \cref{tab:cc}, we also observe that the bcLSTM model performs better than the GloVe CNN model on all the datasets apart from Persuasion for Good under the various context control configurations. For strategy classification in Persuasion for Good, any perturbation beyond window 5 exposes the model to noise and causes the performance to drop below GloVe CNN baseline.

In any conversational classification setup, past contextual information is elemental for recurrent models to understand the flow of the dialogue. Additionally, from the results in \cref{tab:cc}, we can also conclude that future utterances provide key contextual information for the various classification tasks. If the bcLSTM model is not trained on full context, then there is a significant drop in performance even if we evaluate on the full context setup (Row 14, 35).


\subsection{Speaker-specific Context Control}

\begin{table*}[ht]
  \centering
 \resizebox{\linewidth}{!}{
   \begin{tabular}{l||c||ccc|cc}
    \toprule
    \multirow{4}{*}{Methods} & \textbf{IEMOCAP} & & \multicolumn{4}{c}{\textbf{DailyDialog}} \\
     & Emotion & \multicolumn{3}{c|}{Emotion} &  \multicolumn{2}{c}{Act}\\
     
     \cline{2-7} & W-Avg F1 & W-Avg F1& Micro F1 & Macro F1 & W-Avg F1 & Macro F1\\
\hline
    GloVe CNN & 52.04 & 49.36 & 50.32 & 36.87 & 80.71 & 72.07 \\
    GloVe bcLSTM & 61.74 & 52.77 & \textbf{53.85} & 39.27 & \textbf{84.62} & \textbf{79.12} \\
    \quad  \footnotesize{w/o inter} & \textbf{63.73}  & 52.39 & 52.86 & \textbf{39.99} & 81.32 &  74.50 \\
    \quad  \footnotesize{w/o intra} & 56.45  & \textbf{52.81} & 53.54 & 35.93 & 83.80 &  78.69 \\
    \bottomrule
   \end{tabular}
   }
  \caption{Classification performance in test data for emotion prediction in IEMOCAP, emotion prediction in DailyDialog, and act prediction in DailyDialog. Utterances from other speakers and the same speaker are absent respectively in the \emph{w/o inter} and \emph{w/o intra} settings. All scores are average of 20 different runs. Test F1 scores are calculated at best validation F1 scores.}
  \label{table:result-w/o-inter}
\end{table*}

\begin{table*}[ht]
  \centering
   \begin{tabular}{l||c||cc|cc}
    \toprule
    \multirow{4}{*}{Methods} & \textbf{MultiWOZ} & \multicolumn{4}{c}{\textbf{Persuasion For Good}} \\
     & Intent & \multicolumn{2}{c|}{Persuader} &  \multicolumn{2}{c}{Persuadee}\\
     
     \cline{2-6} & W-Avg F1 & W-Avg F1 & Macro F1 & W-Avg F1 & Macro F1\\
     \hline
    GloVe CNN & 84.30 & 67.15 & 54.45 & 58.00 & 41.03 \\
    GloVe bcLSTM & \textbf{96.14} & \textbf{69.26}  & \textbf{55.27} & \textbf{61.18} & \textbf{42.19}\\
    \quad  \footnotesize{w/o inter} & 95.05 &67.81 & 53.24 & 59.44 & 40.63 \\
    \quad  \footnotesize{w/o intra} & 95.75 & 66.06 & 52.23 & 58.65 & 38.93 \\
    \bottomrule
   \end{tabular}
  \caption{Classification performance in test data for intent prediction in MultiWOZ, persuader and persuadee strategy prediction in Persuasion for Good. Utterances from other speakers and the same speaker are absent respectively in the \emph{w/o inter} and \emph{w/o intra} settings. All scores are average of 20 different runs. Test F1 scores are calculated at best validation F1 scores.}
  \label{table:result2-w/o-inter}
\end{table*}

\begin{figure*}[ht]
    \centering
     \includegraphics[width=\linewidth]{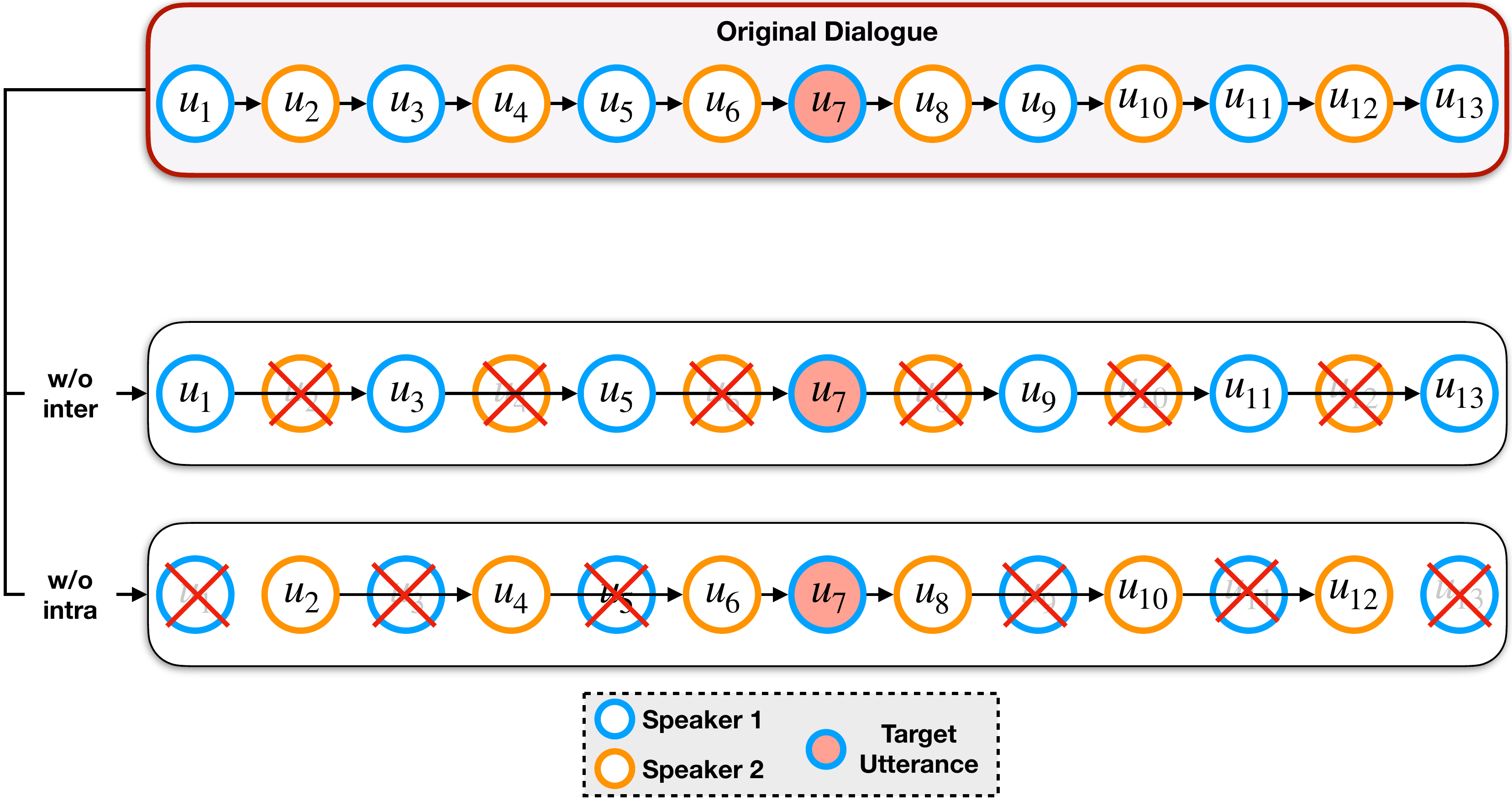}
     \label{fig:speaker-context-cc}
     \caption{Speaker-specific context control schemes.}
\end{figure*}

To further evaluate the intra- and inter-speaker dependence and relation across the different tasks, we adopted two different settings as follows --
\begin{itemize}
    \item \textbf{w/o inter:} when classifying a target utterance from speaker A, we drop the utterances of the speaker B from the context and vice versa.
    \item \textbf{w/o intra:} when classifying a target utterance from speaker A, we only keep utterances of the speaker B and drop all other utterances of speaker A from the context and vice versa.
\end{itemize}
These two different settings are visually illustrated in \cref{fig:speaker-context-cc}.
\paragraph{Utterances of the Non-target Speaker are Important.}
The first setting coerces bcLSTM to only rely on the \emph{target} speaker's (speaker of the target utterance) context in prediction. The results are reported in \cref{table:result-w/o-inter} and \ref{table:result2-w/o-inter}. As expected, performance drops are observed for all the datasets but IEMOCAP and DailyDialog for emotion recognition, reinforcing the fact that the contextual utterances from the non-target speakers are important. Performance drop in DailyDialog dataset for act classification is noticeably the steepest. To understand the behavior on the IEMOCAP dataset, we need to refer to the \cref{fig:heatmap1} and \ref{fig:heatmap2}. In \cref{fig:heatmap2} all the diagonal cells have high intensity, thereby indicating a pattern of the speakers maintaining the same emotion along a dialogue. On the other hand, \cref{fig:heatmap1} illustrates inter-speaker label transitions that maintain emotion consistency---emotion of a speaker is reciprocated by the same or another emotion of the same sentiment by the non-target speaker. Notably, \cref{fig:heatmap2} shows distinctively higher density in its diagonal cells as compared to the diagonal cells in \cref{fig:heatmap1}. This suggests that the speakers in the IEMOCAP dataset repeat the same emotion along consecutive utterances. This tendency of the IEMOCAP dataset, as showed in \cref{fig:heatmap2}, overwhelms the inter-speaker patterns depicted in \cref{fig:heatmap1}. Consequently, this induces a dataset bias. Hence, removing other interlocutor's utterances from the context makes it easier and less confusing for the bcLSTM model to learn relevant contextual representations for the prediction. Contrary to this, although exist, repetitions of same or similar emotions in consecutive utterances of a speaker are less prevalent for emotion recognition in the DailyDialog dataset. Hence, the `w/o inter' setting does not improve the performance as great it does for IEMOCAP.

\paragraph{Utterances of the Target Speaker are \emph{also} Important.}
`w/o intra' scenario reported in \cref{table:result-w/o-inter} and \ref{table:result2-w/o-inter} exhibits the importance of the utterances of the non-target speaker in the classification of the target utterance. In DailyDialog act and MultiWOZ intent classification, even when we remove the contextual utterances from the same speaker, the utterances from the non-target speaker provides key contextual information as evidenced by the performance in the `w/o intra' setting. In those tasks, dropping the utterances of the non-target speaker results in more performance degradation as compared to the case when utterances from the target speaker are removed from the target utterance's context. This observation also supports the dialogue generation works~\cite{zhou2017emotional} that mainly consider previous utterances of the non-target speaker as the context for response generation. For emotion classification in DailyDialog and strategy classification in Persuasion For Good, the results obtained from `w/o intra' setting are
also relatively lesser compared to the
baseline bcLSTM setting. This confirms the higher contextual salience of the target speaker's utterances over the non-target speaker's utterances for these particular tasks. In the case of the IEMOCAP emotion classification, removing the target speaker's utterances from the context causes a substantial performance dip for the reasons stated in the last paragraph.

Interestingly, the ``w/o inter'' setting in the DailyDialog dataset manifests two distinct trends for two different tasks -- act classification and emotion recognition. While non-target speakers' utterances carry a little value for emotion recognition, they are extremely beneficial for act classification. This calls for task-specific context modeling techniques which should be the focus of the future works.

\paragraph{The Key Takeaways of this Experiment.} Although both target and non-target speakers' utterances are useful in several utterance-level dialogue understanding tasks, we observe some divergent trends in some of the tasks used in our experiments. Hence, we surmise that a task-agnostic unified context model may not be optimal in solving all the tasks. In the future, we should strive for task-specific contextual models as each task can have unique futures that make it distinct from the other. One can also think of multi-task architectures where two tasks can corroborate each other in improving the overall utterance-level dialogue understanding performance.

Logically, dropping contextual utterances in a dialogue leads to inconsistency in the context and consequently, it should degrade the performance of a model that relies on the context for inference. Hence, given an unmodified dialogue flow, an ideal contextual model is expected to refer to the right amount of contextual utterances relevant in inferring the label of a target utterance. In contrast, bcLSTM shows performance improvement for emotion classification when utterances from the non-target speaker are dropped (refer to the "w/o inter" row in \cref{table:result-w/o-inter}). The performance does not change much for dialogue act and intent classification in the DailyDialog and MultiWOZ datasets, respectively, when we drop utterances of the target speaker. These contrasting results indicate a potential drawback of the bcLSTM model in efficiently utilizing contextual utterances of both interlocutors in unmodified dialogues for the above mentioned tasks.

\subsection{Context Flipping using Style Transfer}
\begin{figure}[ht]
    \centering
     \includegraphics[width=\linewidth]{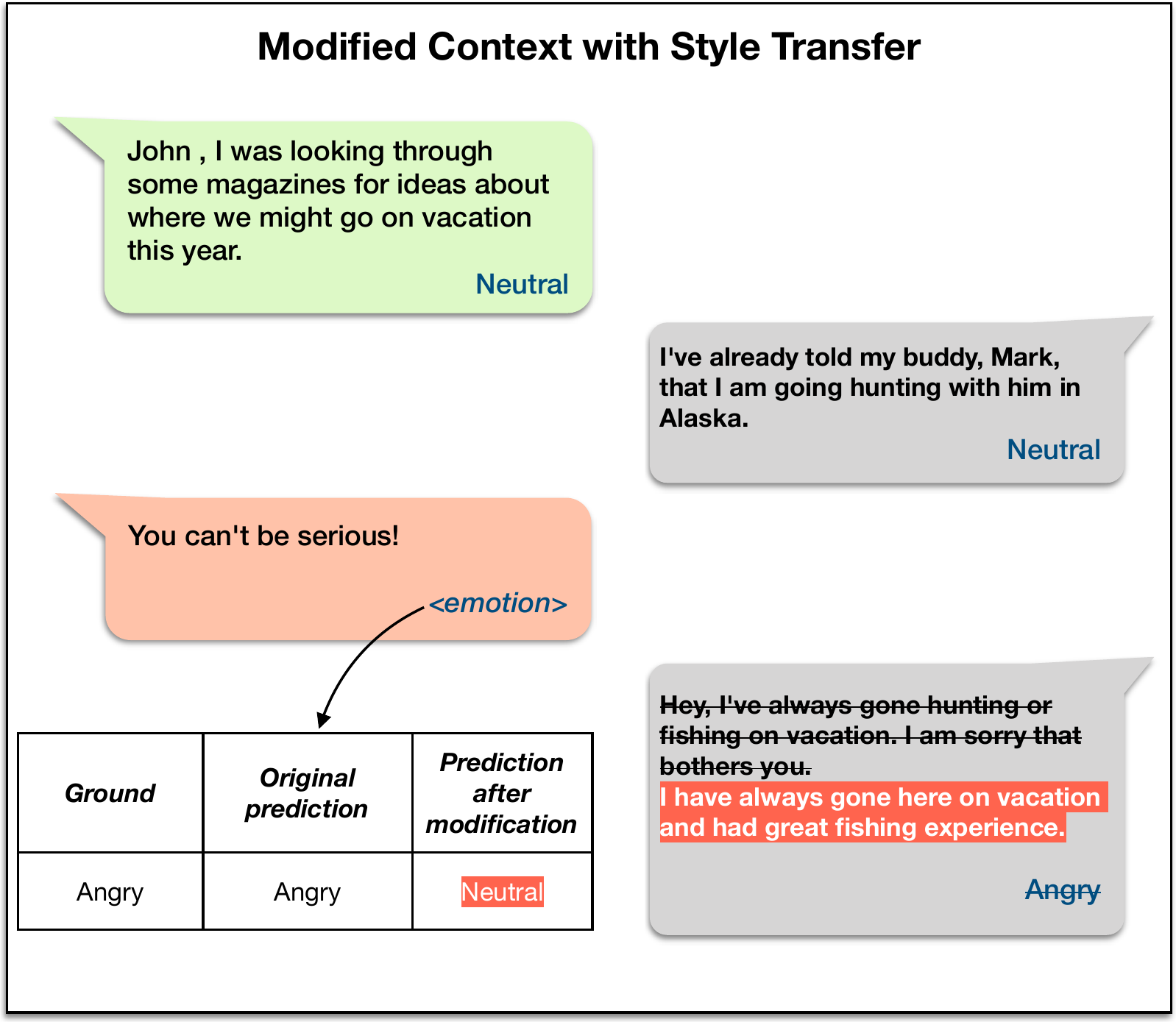}
     \label{fig:style}
     \caption{Predictions under context flipping using style transfer.}
\end{figure}
\begin{table*}[ht]
  \centering
   \begin{tabular}{c|ccc|c||cc}
    \toprule
     & \multicolumn{3}{c|}{Style Transfer Strategy} &  Window & 
    IEMOCAP & DailyDialog\\
    \# & Past & Future & Target & & Emotion & Emotion \\
    \midrule
    1 & \xmark & \xmark & \xmark & - & 61.9 & 41.16 \\
    \midrule
    2 & \checkmark & \xmark & \xmark & 3 & 59.47 & 36.69 \\
    3 & \checkmark & \xmark & \xmark & 5 & 58.28 & 36.13 \\
    4 & \checkmark & \xmark & \xmark & 10 & 55.58 & - \\
    5 & \xmark & \checkmark & \xmark & 3 & 60.76 & 38.52 \\
    6 & \xmark & \checkmark & \xmark & 5 & 59.92 & 38.12 \\
    7 & \xmark & \checkmark & \xmark & 10 & 57.16 & - \\
    8 & \checkmark & \checkmark & \xmark & 3 & 56.35 & 34.51 \\
    9 & \checkmark & \checkmark & \xmark & 5 & 51.96 & 33.20 \\
    10 & \checkmark & \checkmark & \xmark & 10 & 46.28 & - \\

    \bottomrule
   \end{tabular}
  \caption{Results for emotion classification in IEMOCAP (Weighted F1) and DailyDialog (Macro F1) with context flipping using style transfer. In DailyDialog, we constrain the window size to 3 and 5 as there are an average of 8 utterances per dialogue in the dataset. We don't change the style of the target utterance as that would imply an illogical evaluation with the gold emotion class.}
  \label{tab:style}
\end{table*}

In emotion classification tasks, the emotion of contextual utterances plays a vital role in determining the emotion of the target utterance. We can study this effect quantitatively using the method known as textual style transfer. Style transfer of text is defined as transferring a piece of text (generally a single sentence or a short paragraph) from one domain (style) to another while preserving the underlying content. In particular, we use the sentiment style transfer method that flips the sentiment of the input text while preserving the main semantic content. For example, given an input sentence: \textit{Hey, I've always gone hunting or fishing on vacation. I am sorry that bothers you.} with \textit{negative} sentiment would be transferred to \textit{I have always gone here on vacation and had great fishing experience.} with \textit{positive} sentiment. Sentiment style transfer is used because sentiment is closely related to emotion and ample parallel data is available between positive to negative and vice versa sentiment styles.

The main objective of this study is to analyze how affect or emotion label orientation of contextual utterances plays a vital role in utterance level emotion classification. Here, we devise a method based on the sentiment style transfer technique to study this effect systematically.

\paragraph{Method.} The YELP reviews dataset \citep{li2018delete} contains parallel sentences between different sentiment styles. We first fine-tune a pretrained T5 seq2seq model \cite{raffel2019exploring} on the parallel source and target sentences in the dataset using \textit{teacher forcing} via maximum likelihood estimation. After training, this model is capable of changing the sentiment of an input sentence from positive to negative or vice versa while preserving the main semantic content. We call this model the \textit{Style Transfer} model.

To apply this model in our datasets, we consider the positive group of emotions: \textit{happy, joy, excited} correspond to positive sentiment, and the negative group of emotions: \textit{sad, angry, frustrated, disgust, fear} correspond to negative sentiment. We train a \textit{utterance-based} bcLSTM model with unchanged train and validation data. During evaluation, we modify the test data using the following method:
\begin{itemize}
    \item For each target utterance $u_t$ in a dialogue, we take a window of $w$ neighbouring utterances (context) in the past, or future, or both.
    \item If the contextual utterance belong to positive (negative) emotion group we flip its style to negative (positive). This is achieved using the \textit{Style Transfer} model. The contextual utterance is kept unchanged if it belongs to \textit{neutral} emotion category.
    \item The target utterance is kept unchanged.
\end{itemize}

\paragraph{Observations.} We analyze the results with a window size of 3, 5, 10 in IEMOCAP and 3, 5 in DailyDialog in \cref{tab:style}. In IEMOCAP, style transfer in progressively larger window sizes results in bigger performance drops. With a window size of 10 in both past and future directions, the F1-score of 46.28 is around 15\% lesser than the original score of 61.9. In DailyDialog, the very high frequency of neutral labels would ensure that most of the contextual utterances will remain unchanged. However, we still observe a drop in performance in various settings in \cref{tab:style}. The drop when we modify the past utterances is relatively more compared to modifying the future utterances suggesting that past utterances are relatively more important for the emotion classification.

The context style transfer method keeps the content of the target and contextual utterances unaltered while reversing the affect or label orientation of the contextual utterances. This process results in significantly poor performance in both the emotion classification tasks suggesting that the label orientation of the contextual utterances plays a vital role in the overall classification performance. We strengthen this label dependency hypothesis with more extensive experiments in \cref{sec:lca}.

\subsection{Attacks with Context and Target Paraphrasing}
\label{sec:attack}
\begin{table*}[t]
  \centering
 \resizebox{\linewidth}{!}{
   \begin{tabular}{l|c|ccc|c||cccccc}
    \toprule
  & \multicolumn{1}{c|}{Method}  & \multicolumn{3}{c|}{Strategy} & Window & \textbf{IEMOCAP} & \textbf{Dailydialog}  & \textbf{Dailydialog} &\textbf{MultiWOZ}   & \textbf{Persuasion} &  \textbf{Persuasion}\\
    \# & PA/SA & Past & Future & Target & & Emotion & Emotion & Act & Intent & ER & EE\\
    \midrule
1 & - & - & - & - & - & 61.9 & 41.16 & 79.46 & 96.22 & 56.28 & 44.83 \\
\midrule
2 & PA & \checkmark & \xmark & \xmark  &  3  &  61.09 & 40.82 & 75.81 & 95.67 & 56.46 & 43.64 \\
3 & PA & \checkmark & \xmark & \xmark  &  5  &  60.93 & 38.79 & 77.23 & 95.53 & 56.41 & 41.93 \\
4 & PA & \checkmark & \xmark & \xmark  &  10  &  59.83 & - & - & 95.23 & 54.89 & 39.89 \\
5 & PA & \xmark & \checkmark & \xmark  &  3  &  61.58 & 39.6 & 79.11 & 95.94 & 55.83 & 43.21 \\
6 & PA & \xmark & \checkmark & \xmark  &  5  &  60.99 & 39.77 & 79.17 & 95.64 & 55.43 & 40.67 \\
7 & PA & \xmark & \checkmark & \xmark  &  10  &  60.72 & - & - & 95.77 & 57.12 & 43.36 \\
8 & PA  & \checkmark & \checkmark & \xmark  &  3  &  59.43 & 37.16 & 76.61 & 94.87 & 57.44 & 42.51 \\
9 & PA & \checkmark & \checkmark & \xmark  &  5  &  58.36 & 38.76 & 76.53 & 94.61 & 53.32 & 43.33 \\
10 & PA & \checkmark & \checkmark & \xmark  &  10  &  57.29 & - & - & 94.31 & 54.36 & 43.8 \\
11 & PA & \xmark & \xmark & \checkmark  & - &  58.08 & 37.16 & 75.3 & 93.78 & 50.24 & 38.78 \\
12 & PA & \checkmark & \checkmark & \checkmark  &  3  &  56.53 & 23.46 & 73.16 & 91.47 & 47.5 & 37.39 \\
13 & PA & \checkmark & \checkmark & \checkmark  &  5  &  53.64 & 28.59 & 73.18 & 90.98 & 45.31 & 35.16 \\
14 & PA & \checkmark & \checkmark & \checkmark  &  10  &  51.33 & - & - & 90.58 & 49.0 & 32.49 \\
\midrule
15  & SA &  \checkmark & \xmark & \xmark  &  3  &  59.59 & 36.5 & 76.07 & 95.63 & 56.81 & 43.14 \\
16 & SA &  \checkmark & \xmark & \xmark  &  5  &  59.67 & 36.86 & 76.14 & 95.49 & 57.28 & 42.63 \\
17 & SA &  \checkmark & \xmark & \xmark  &  10  &  59.06 & - & - & 95.44 & 54.87 & 41.77 \\
18 & SA &  \xmark & \checkmark & \xmark  &  3  &  61.11 & 39.3 & 79.42 & 95.94 & 56.15 & 45.46 \\
19 & SA &  \xmark & \checkmark & \xmark  &  5  &  61.05 & 37.53 & 79.31 & 95.87 & 56.96 & 41.2 \\
20 & SA &  \xmark & \checkmark & \xmark  &  10  &  59.31 & - & - & 95.93 & 56.36 & 42.49 \\
21 & SA &  \checkmark & \checkmark & \xmark  &  3  &  59.14 & 37.04 & 75.77 & 95.34 & 55.73 & 42.37 \\
22& SA &  \checkmark & \checkmark & \xmark  &  5  &  56.67 & 35.46 & 76.39 & 95.12 & 56.05 & 41.05 \\
23 & SA &  \checkmark & \checkmark & \xmark  &  10  &  54.2 & - & - & 94.98 & 55.67 & 40.51 \\
24 & SA &  \xmark & \xmark & \checkmark  & - &  53.91 & 30.42 & 75.63 & 94.8 & 50.44 & 38.74 \\
25 & SA &  \checkmark & \checkmark & \checkmark  &  3  &  48.11 & 22.55 & 75.77 & 93.1 & 46.55 & 35.21 \\
26 & SA &  \checkmark & \checkmark & \checkmark  &  5  &  44.32 & 20.58 & 76.39 & 92.81 & 49.08 & 32.65 \\
27 & SA &  \checkmark & \checkmark & \checkmark  &  10  &  40.22 & - & - & 92.72 & 49.04 & 35.64 \\
    \bottomrule
   \end{tabular}
   }
  \caption{Results for PA: \textit{Paraphrasing-based Attack}; SA: \textit{Spelling-based Attack} in \textit{utterance-based} GloVe bcLSTM model. In DailyDialog, we constrain the window size to 3 and 5 as there are an average of 8 utterances per dialogue in the dataset. Scores are W-Avg F1 in IEMOCAP Emotion and MultiWOZ Intent; Macro F1 in the rest.}
  \label{tab:pp}
\end{table*}

\begin{table*}[t]
  \centering
 \resizebox{\linewidth}{!}{
   \begin{tabular}{l|ccc|cccccc}
    \toprule
  & & \multicolumn{2}{c|}{Method} & \textbf{IEMOCAP} & \textbf{Dailydialog}  & \textbf{Dailydialog} &\textbf{MultiWOZ}   & \textbf{Persuasion} &  \textbf{Persuasion}\\
    \# & & PA & SA & Emotion & Emotion & Act & Intent & ER & EE\\
    \midrule
    1 &\parbox[t]{2mm}{\multirow{3}{*}{\rotatebox[origin=c]{90}{\small{GloveCNN}}}}& - & - & 51.08 & 38.72 & 71.2 & 84.64 & 54.44 & 39.95 \\
    2 && \checkmark & \xmark  &  39.19(${\downarrow}$23.27) & 23.82(${\downarrow}$39.64) & 62.93(${\downarrow}$13.01) & 70.34(${\downarrow}$16.89) & 42.8(${\downarrow}$21.38) & 33.59(${\downarrow}$15.91) \\
    3 && \xmark & \checkmark  &  44.68(${\downarrow}$12.52) & 22.7(${\downarrow}$41.37) & 61.86(${\downarrow}$13.11) & 74.58(${\downarrow}$11.88) & 42.95(${\downarrow}$21.10) & 28.99(${\downarrow}$27.43) \\
    \midrule
    4 &\parbox[t]{2mm}{\multirow{3}{*}{\rotatebox[origin=c]{90}{\small{bcLSTM}}}}& - & - & 61.9 & 41.16 & 79.46 & 96.22 & 56.28 & 44.83 \\
    5 && \checkmark & \xmark  &  58.08(${\downarrow}$6.17) & 37.16(${\downarrow}$9.71) & 75.3(${\downarrow}$5.23) & 93.78(${\downarrow}$2.53) & 50.24(${\downarrow}$10.73) & 38.78(${\downarrow}$13.49) \\
    6 && \xmark & \checkmark  &  53.91(${\downarrow}$12.90) & 30.42(${\downarrow}$26.09) & 75.63(${\downarrow}$4.82) & 94.8(${\downarrow}$4.82) & 50.44(${\downarrow}$10.37) & 38.74(${\downarrow}$13.58) \\

    \bottomrule
   \end{tabular}
   }
  \caption{Results for PA: \textit{Paraphrasing-based Attack}; SA: \textit{Spelling-based Attack} in GloVe CNN model and comparing it to bcLSTM results in \cref{tab:pp}. IEMOCAP Emotion, MultiWOZ Intent: W-Avg F1; Rest: Macro F1.  
  }
  \label{tab:pp2}
\end{table*}

Modern machine learning systems are often susceptible to attacks that slightly perturb the input without any drastic change in the semantics. Although prevalent in images, adversarial examples also exist in neural network-based NLP applications. In the context of NLP, crafting adversarial examples would require making character-, word-, or sentence-level modifications to the input text to trick the classifier into misclassification. Paraphrasing sentences is one such method to construct effective adversarial examples~\cite{iyyer2018adversarial}. We conduct several experiments to evaluate the sensitivity of utterance-level dialogue understanding systems to input paraphrasing. It should be noted that although task-specific adversarial strategies could be adopted, we chose to use a general set of attacking strategies in order to understand the behavior of the baseline across different tasks and datasets. This also facilitates a fair comparison among the tasks and whether there is a confounding factor that differentiates one task from another under the same attacking strategies.

\begin{figure}[ht!]
    \centering
     \includegraphics[width=\linewidth]{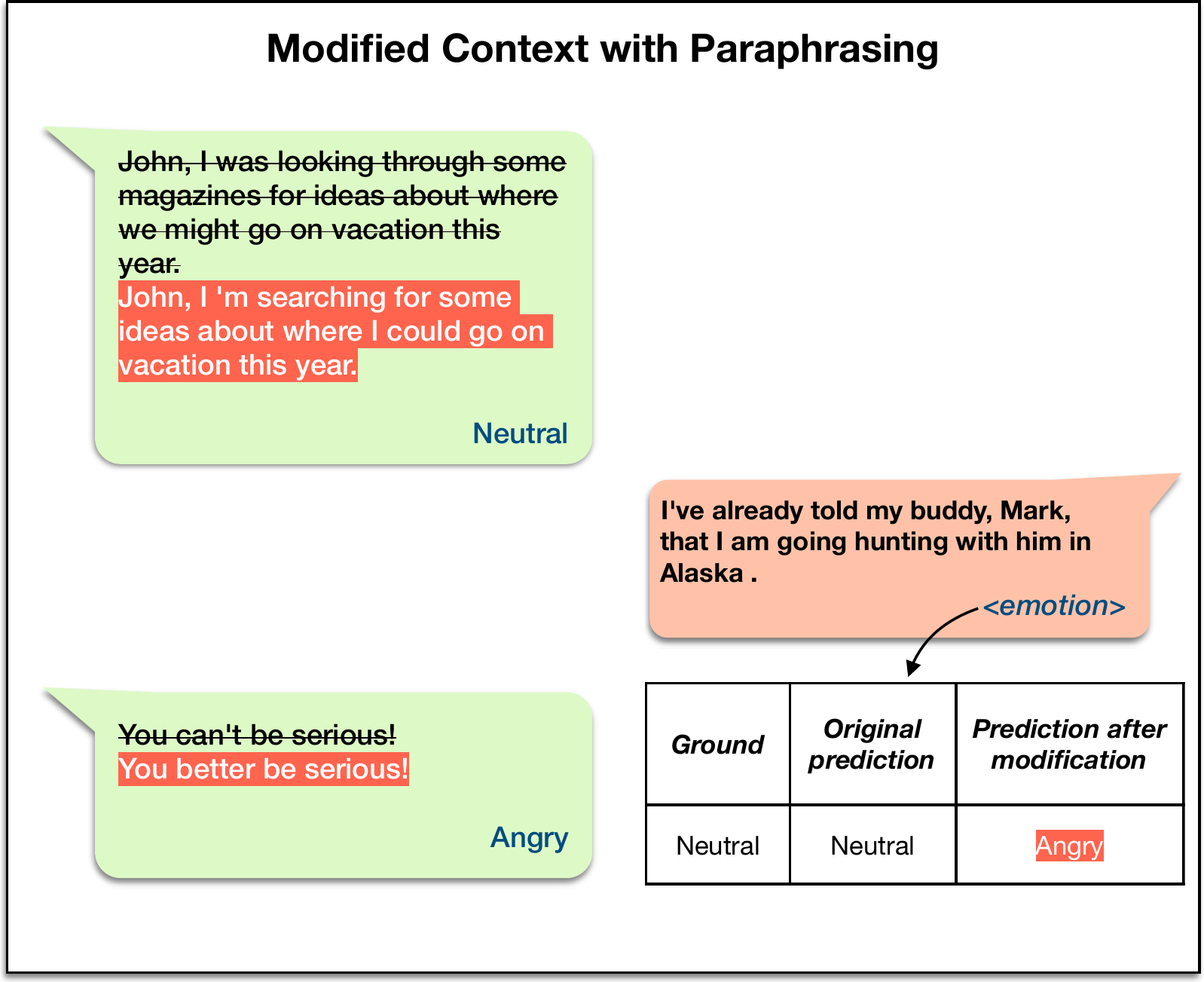}
     \caption{Predictions under modified context with \textit{Paraphrasing-based Attack}.}
     \label{fig:pp1}
\end{figure}

\begin{figure}[t]
    \centering
     \includegraphics[width=\linewidth]{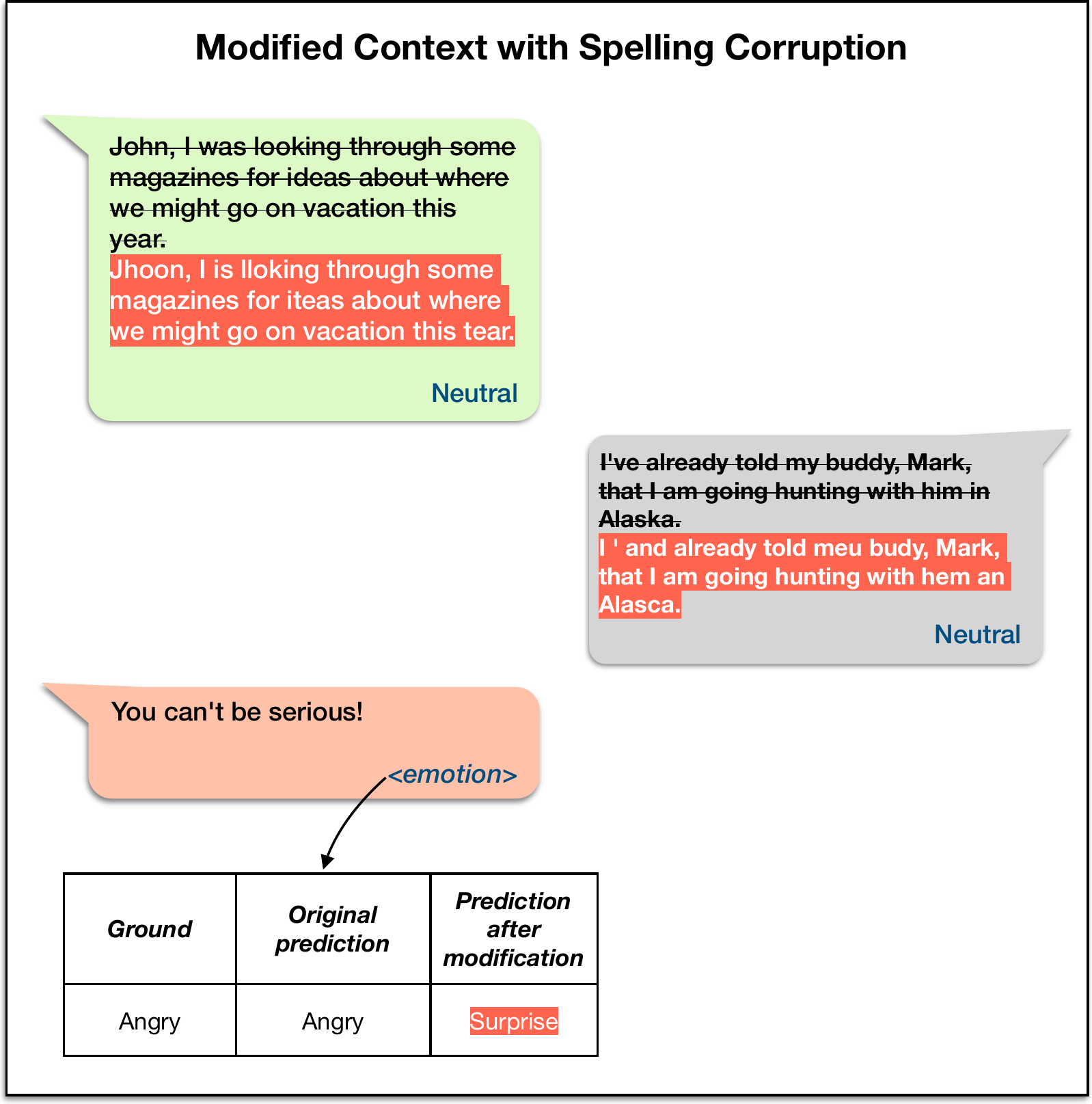}
    \caption{Predictions under modified context with \textit{Spelling-based Attack}.}
     \label{fig:pp2}
\end{figure}

\paragraph{Method.} We use the following scheme to analyze this effect:
\begin{itemize}
  \item The input utterances are modified at either word or character level.
  \begin{itemize}
  \item For word-level modification, an average of 3 to 4 words are selected per utterance and masked. The pre-trained RoBERTa model is then used to fill the masks with the most likely candidates. The utterance with substituted words form the new input. We call this method \textit{Paraphrasing-based Attack}.
  \item The character-level modification is achieved by changing the spelling of, on average, 3 to 4 words per input utterance. We call this method \textit{Spelling-based Attack}.
  \end{itemize}
  \item For each utterance ($u_t$) in a dialogue, we take a window of $w$ immediate neighbouring utterances (context) on which the above modifications are performed. The window is selected as follows:
  \begin{itemize}
  \item Only past $w$ utterances: $u_{t-w}, .., u_{t-1}$
  \item Only future $w$ utterances: $u_{t+1}, .., u_{t+w}$
  \item Past $w$ and future $w$ utterances: $u_{t-w}, .., u_{t-1}, u_{t+1}, .., u_{t+w}$
  \item Past $w$, future $w$, and the target utterance: $u_{t-w}, .., u_{t-1}, u_t, u_{t+1}, .., u_{t+w}$
  \item Only the target utterance: $u_t$
  \end{itemize}
  In the last case, the window is empty. In other cases, we experiment with window size $w=3,5,10$.
\end{itemize}

We train an \textit{utterance-based} GloVe bcLSTM and a GloVe CNN model with unadulterated train and validation data. During evaluation, however, the context and target are paraphrased as described before. The results of these experiments for bcLSTM and GloVe CNN are shown in \cref{tab:pp} and \cref{tab:pp2}, respectively. We show a few examples cases in \cref{fig:pp1} and \ref{fig:pp2}.

\paragraph{Observations.} We observe that the \textit{Spelling-based Attack} is significantly more effective than \textit{Paraphrasing-based Attack} in fooling the classifier in the emotion classification tasks. The classification performance progressively deteriorates with larger window sizes. This is expected, since spelling change would create out-of-vocabulary words that are missing from pre-trained GloVe. Models that operate on sub-word-based vocabulary would possibly be more resilient to this kind of attack.


In DailyDialog act classification, \textit{Paraphrasing-based Attack} or \textit{Spelling-based Attack} on only future utterances doesn't affect the results at all. The classification performance still remains very close to the original score of 79.46 \%. As evidenced in \cref{fig:heatmap1}, there is a strong reliance on the label and content of past utterance in this task. For example, a \textit{question} is likely to be followed by an \textit{inform} or another \textit{question} and much less likely to be followed by a \textit{commissive} utterance. Unchanged past context thus results in performance that is very close to the original setup. Attacking the past utterances combined with future and/or target utterances results in a relatively bigger performance drop. We also notice that the drop in performance is relatively much lesser than the other tasks except in MultiWOZ for intent classification. This is possibly because the act labels are mostly driven by the sentence type and hence unlikely to be affected from paraphrasing or spelling-based perturbations. For instance, around 30\% of the act labels are of type \textit{question}, and our attacks are almost guaranteed not to change an utterance with label \textit{question} to something which might be classified as \textit{inform}, \textit{commissive}, or \textit{directive}. Overall, we observe a consistent plunge in the performance when the target utterance is attacked by either paraphrase or spelling-based attacks.

For intent classification in MultiWOZ, utterances often have keywords which indicate the label (presence of \textit{train} might indicate class label of \textit{find train} or \textit{book train}). In these cases, if the target utterance is not paraphrased, the model is still likely to predict the correct label.
Finally, in Persuasion for Good, both the attack methods result in a performance drop in a similar range. We also observe that the attack methods are slightly more effective in fooling the classifier for persuadee strategy classification.

In terms of window direction, we observe that perturbations in the past or future utterances result in a similar range of reduction in performances. One notable exception is act prediction in DailyDialog, where the model continues to perform near the original score of 79.46\% irrespective of the attack in future utterances in the window.

\paragraph{Performance Comparison for Attacks in GloVe CNN and GloVe bcLSTM.} We summarize the performance of GloVe CNN and GloVe bcLSTM models against \textit{Paraphrasing-based Attack} and \textit{Spelling-based Attack} in \cref{tab:pp2}. For all the tasks, we observe a very significant drop in performance for GloVe CNN. For example, in emotion classification, the drop is around 23\% and 40\% for \textit{Paraphrasing-based Attack} in IEMOCAP and DailyDialog respectively. However, for the same setting, the relative decrease in performance is only around 6\% and 9\% for bcLSTM. We observe the same trend in other tasks where it can be seen that the bcLSTM model is much more robust against the attacks compared to the CNN model. This is because contextual models such as bcLSTM are harder to fool as the context carry key information regarding the semantics of the target and salient information can be inferred about the target using its' context. It is thus evident that even when the target utterance is corrupted, bcLSTM is capable of using contextual information to predict the label correctly, and subsequently the decline in performance is much lesser.

In principle, our findings in \cref{tab:pp2} can be related to how transformer-based pre-trained language models work. For example, in BERT \cite{devlin2018bert}, the masked language modeling (MLM) and the next sentence prediction (NSP) objective forces the model to infer or predict the target using contextual information. Such contextual models are more powerful and robust because context information plays a crucial role in almost every natural language processing task. \textbf{An objective similar to next sentence prediction in BERT or permutation language modeling in XLNET \cite{yang2019xlnet} can be used for conversation level pre-training to improve several downstream conversational tasks.} Such approaches have been found to be useful in the past \cite{hazarika2019emotion}.


\subsection{Label-Constrained Dialogue Level Utterance Augmentation}
\label{sec:lca}

\begin{figure*}[ht!]
    \centering
    \begin{subfigure}{0.49\textwidth}
     \includegraphics[width=0.9\linewidth]{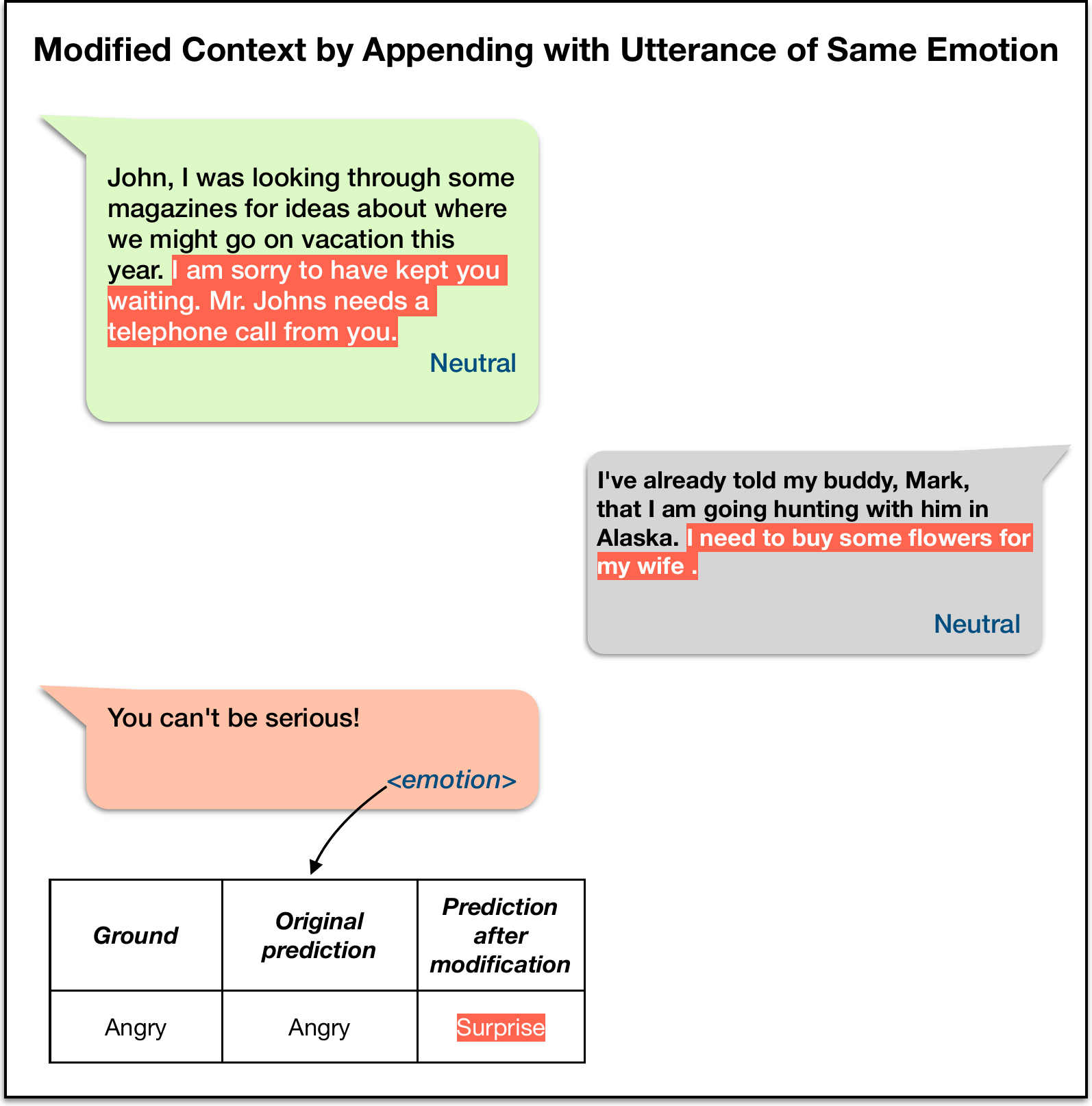}
      \label{fig:label-augment-concat-same}
     \end{subfigure}
     \begin{subfigure}{0.49\textwidth}
     \includegraphics[width=0.9\linewidth]{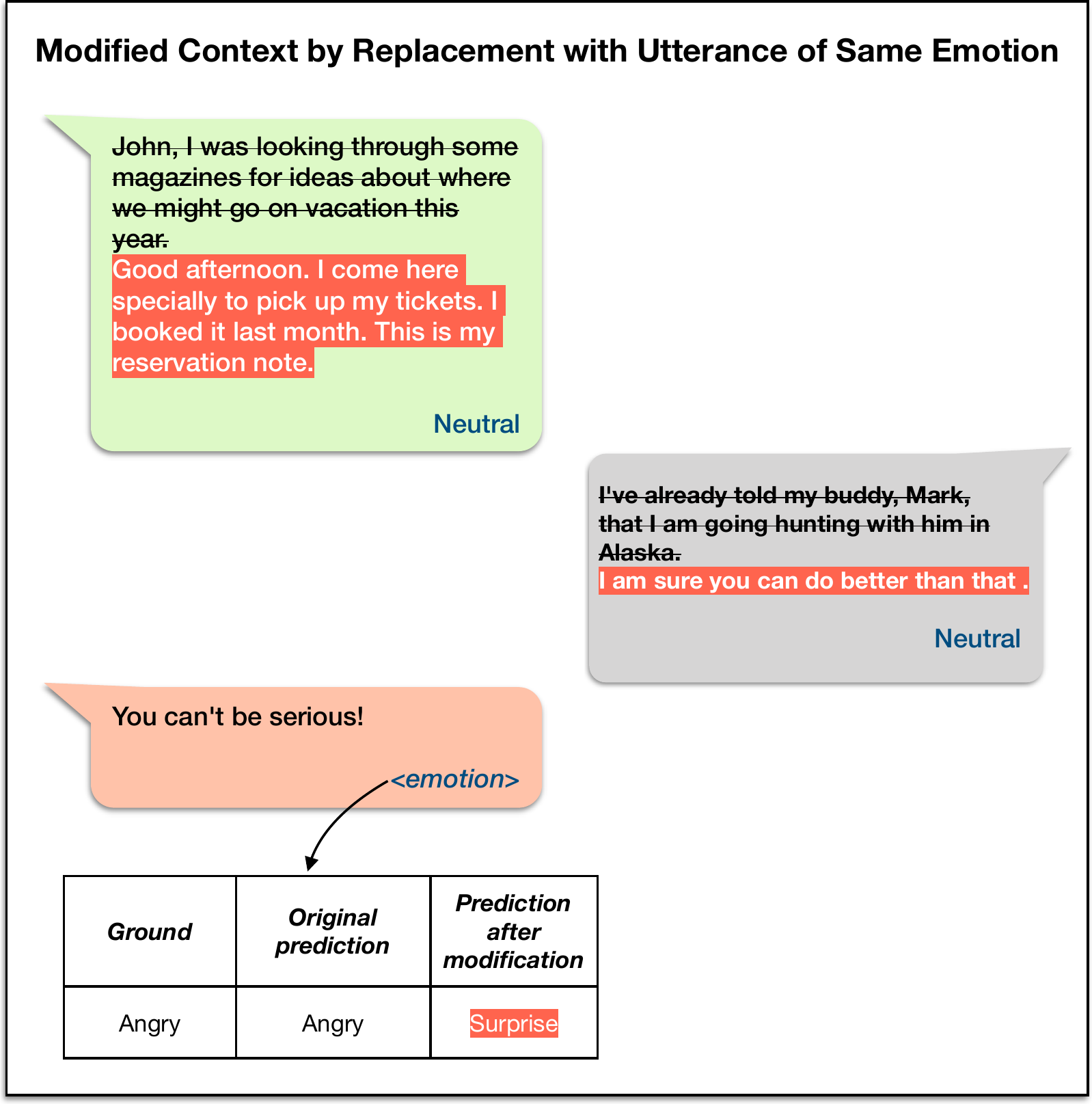}
      \label{fig:label-augment-replace-same}
     \end{subfigure}
     \begin{subfigure}{0.49\textwidth}
     \includegraphics[width=0.9\linewidth]{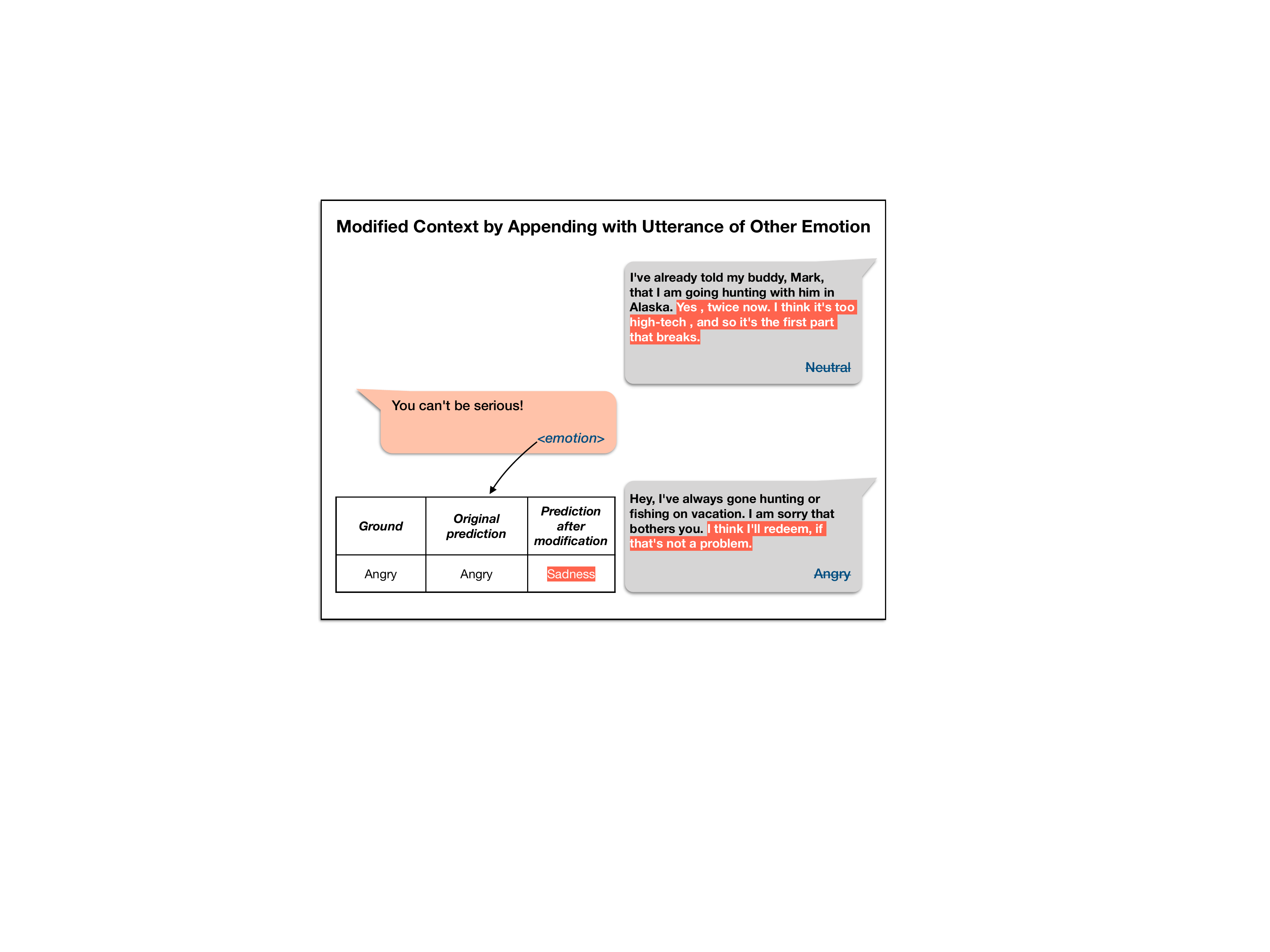}
      \label{fig:label-augment-concat-different}
     \end{subfigure}
     \begin{subfigure}{0.49\textwidth}
     \includegraphics[width=0.9\linewidth]{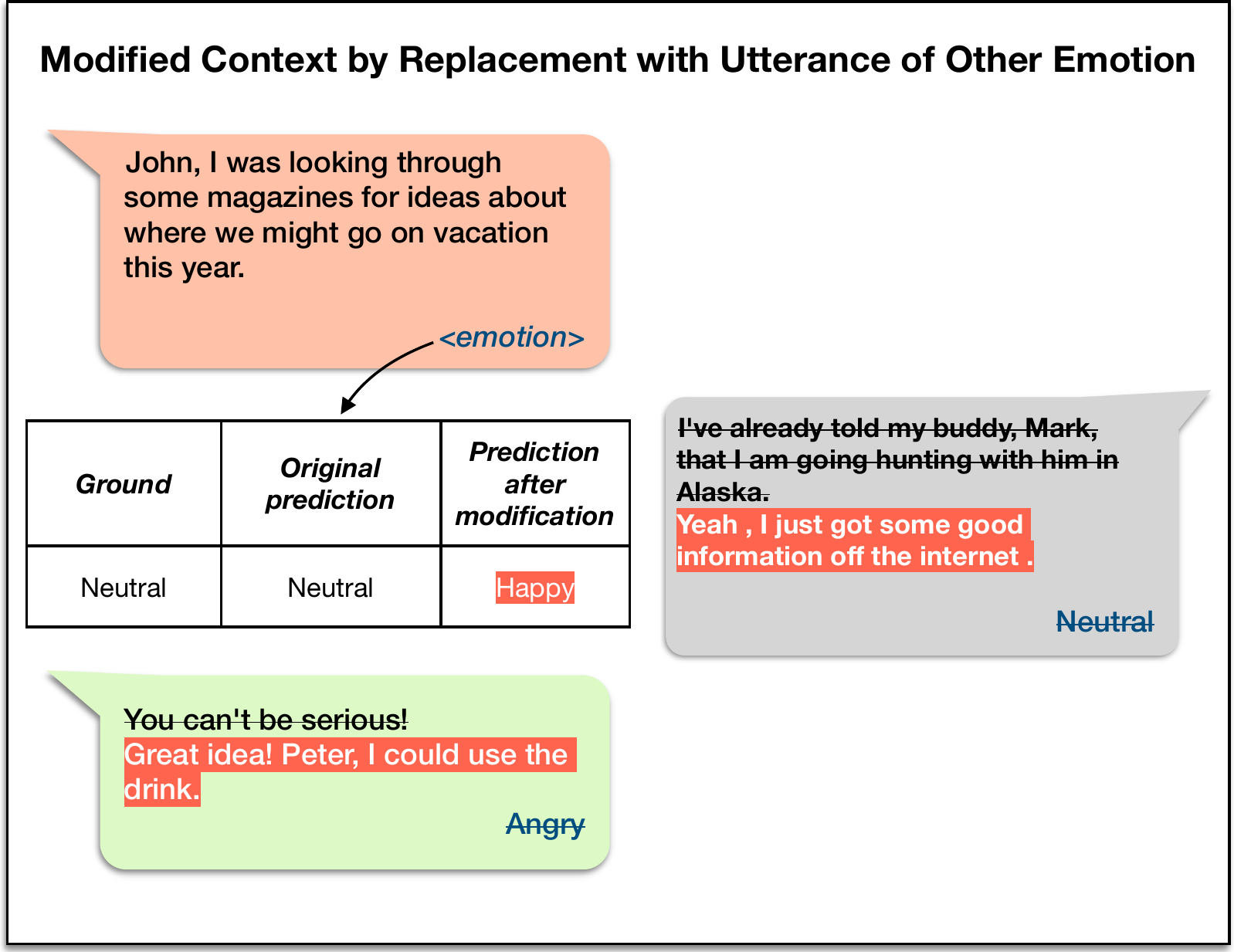}
      \label{fig:label-augment-replace-different}
     \end{subfigure}
     \caption{Predictions under \textit{Label-Constrained Dialogue Level Utterance Augmentation}. We concatenate or replace contextual utterances with other utterances belonging to same or different class category. Here, the examples are shown for DailyDialog emotion classification.}
     \label{fig:label-augment}
\end{figure*}

\begin{table*}[hbtp]
  \centering
 \resizebox{\linewidth}{!}{
   \begin{tabular}{l|c|c|c||cccccc}
    \toprule
  & \multicolumn{1}{c|}{Constraint} &
  \multicolumn{1}{c|}{Strategy} & Window & \textbf{IEMOCAP} & \textbf{Dailydialog}  & \textbf{Dailydialog} &\textbf{MultiWOZ}   & \textbf{Persuasion} &  \textbf{Persuasion}\\
  \# & SL/DL & Concat/Replace & & Emotion & Emotion & Act & Intent & ER & EE \\
\midrule
1 & \xmark & \xmark & - & 61.9 & 41.16 & 79.46 & 96.22 & 56.28 & 44.83 \\
\midrule
2 & SL & Concat & 5 & 61.5 & 34.3 & 81.09 & 88.71 & 56.71 & 43.44 \\
3 & SL & Replace & 5 & 61.31 & 30.06 & 77.72 & 77.92 & 55.56 & 43.2 \\
4 & SL & Concat & 10 & 61.03 & - & - & 89.31 & 56.3 & 44.02 \\
5 & SL & Replace & 10 & 59.92 & - & - & 77.23 & 53.27 & 41.9 \\
6 & SL & Concat & 20 & 60.44 & - & - & - & 54.39 & 42.28 \\
7 & SL & Replace & 20 & 59.74 & - & - & - & 52.28 & 41.19 \\
8 & SL & Concat & 30 & 60.99 & - & - & - & - & - \\
9 & SL & Replace & 30 & 59.11 & - & - & - & - & - \\
10 & SL & Concat & 40 & 61.39 & - & - & - & - & - \\
11 & SL & Replace & 40 & 59.53 & - & - & - & - & - \\
12 & SL & Concat & 50 & 61.36 & - & - & - & - & - \\
13 & SL & Replace & 50 & 59.97 & - & - & - & - & - \\
\midrule
14 & DL & Concat & 5 & 52.77 & 32.8 & 68.57 & 81.87 & 54.04 & 38.95 \\
15 & DL & Replace & 5 & 47.1 & 27.68 & 60.11 & 62.51 & 54.73 & 36.74 \\
16 & DL & Concat & 10 & 50.21 & - & - & 81.48 & 52.11 & 37.49 \\
17 & DL & Replace & 10 & 38.87 & - & - & 61.4 & 51.02 & 39.66 \\
18 & DL & Concat & 20 & 47.25 & - & - & - & 52.49 & 40.53 \\
19 & DL & Replace & 20 & 37.06 & - & - & - & 51.27 & 38.25 \\
20 & DL & Concat & 30 & 46.22 & - & - & - & - & - \\
21 & DL & Replace & 30 & 34.9 & - & - & - & - & - \\
22 & DL & Concat & 40 & 45.69 & - & - & - & - & - \\
23 & DL & Replace & 40 & 35.33 & - & - & - & - & - \\
24 & DL & Concat & 50 & 47.01 & - & - & - & - & - \\
25 & DL & Replace & 50 & 33.86 & - & - & - & - & - \\
    \bottomrule
   \end{tabular}
   }
  \caption{Results for \textit{Label-Constrained Dialogue Level Utterance Augmentation} in different datasets. We concatenate or replace contextual utterances with other utterances belonging to same (SL) or different (DL) class category. We constrain the window size according to average count of utterances per dialogue in each dataset. Scores are W-Avg F1 in IEMOCAP Emotion and MultiWOZ Intent; Macro F1 in the rest.}
  \label{tab:lca}
\end{table*}
In utterance level dialogue classification tasks, labels of consecutive utterances are often inter-related. Recurrent models will usually learn this pattern of label dependency during training. To analyze how large is the dependency with labels, we design the following experimental setup.

\paragraph{Method.} We train an \textit{utterance-based} bcLSTM model with unchanged train and validation data. During evaluation, we modify the test data in the following way:

\begin{itemize}
  \item For each utterance ($u_t$) in a dialogue, contextual utterances in the past and future with window size of $w$ are going to be modified. The target utterance $u_t$ will be kept unchanged.
  \item First, for each contextual utterance in the window $w$ ($u_{t-w}, .., u_{t-1}, u_{t+1}, .., u_{t+w}$), a new utterance will be selected using the following constraints:
  \begin{itemize}
      \item It will be drawn from some other dialogue in the test data.
      \item The label of the drawn utterance will either be same or different from the original label of the respective contextual utterance.
  \end{itemize}

  \item The modification will then be performed to the test data in one of the following two ways:
  \begin{itemize}
      \item The selected utterance will be concatenated at the end of the original utterance to form the new contextual utterance. We call this strategy $Concat$.
      \item The selected original utterance will be replaced with the selected utterance. The selected utterance will thus form the new contextual utterance. We call this strategy $Replace$.
  \end{itemize}
  \end{itemize}

\begin{figure*}[t]
    \centering
     \includegraphics[width=\linewidth]{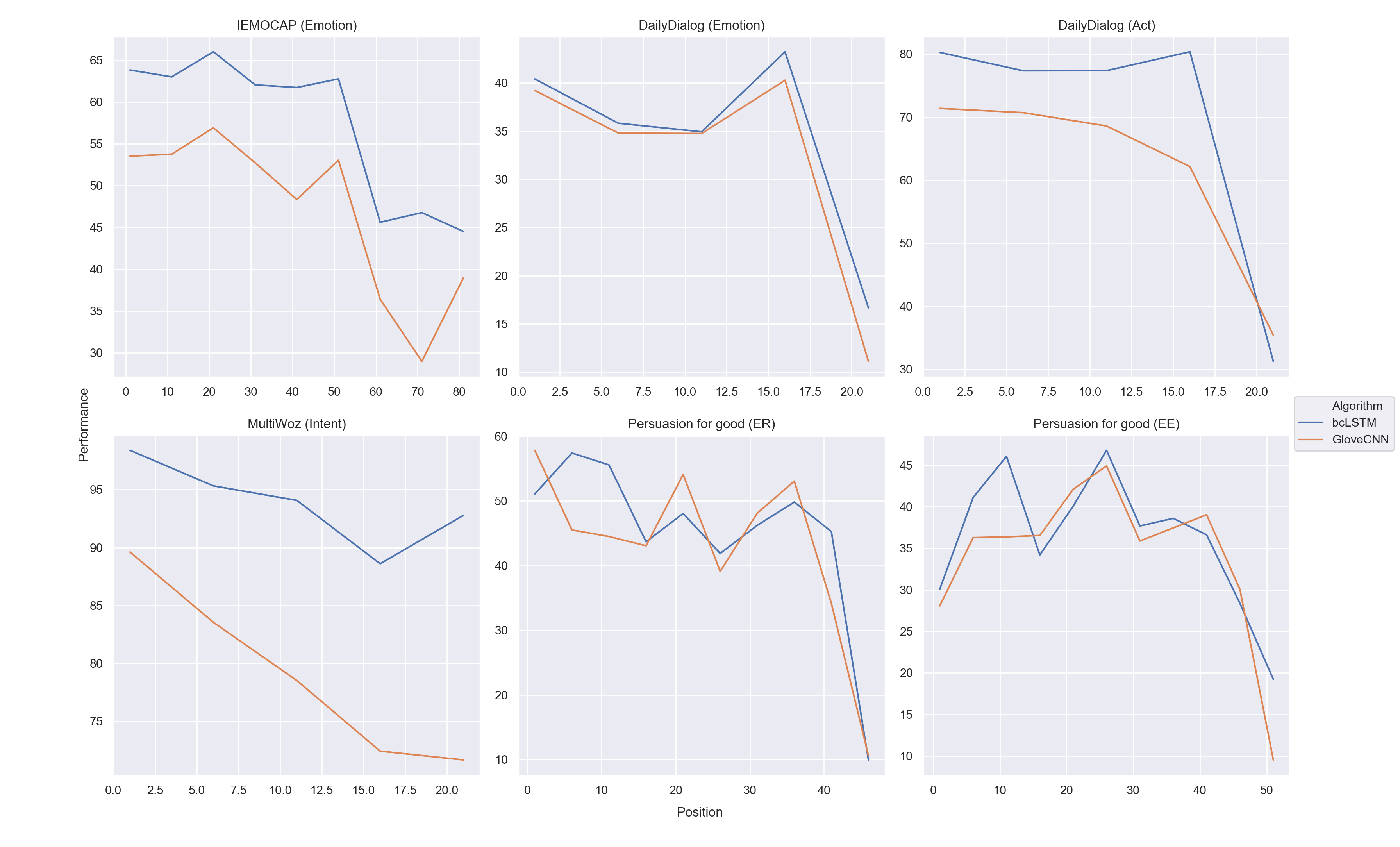}
     \caption{Classification performance with respect to the position of the utterances. Scores are Weighted-F1 in IEMOCAP, MultiWOZ and Macro F1 in the rest.}
     \label{fig:position}
\end{figure*}

We evaluate this experimental setup with window size $w=5,10,20,30,40,50$ in IEMOCAP, $w=5$ in DailyDialog, $w=5,10$ in MultiWOZ and $w=5,10,20$ in Persuasion for Good. The results of this experiments are shown in \cref{tab:lca}. We show a few example cases in \cref{fig:label-augment}.

\paragraph{Observations.} We first look at the results in the IEMOCAP dataset, which was originally curated from actors performing improvisations or scripted scenarios, specifically selected to elicit emotional expressions. The scripts were written by considering the affective aspect of the content. We think, to amplify and enrich the emotional content of the dialogues, the utterances in the IEMOCAP dataset were scripted by enforcing strong label dependency among the utterances e.g., the presence of strong negative emotion sequence is observed in the dialogues where one of the speakers elicits \emph{anger}. This phenomenon is unique to IEMOCAP and missing in the DailyDialog dataset. \textbf{From the experimental results in \cref{tab:lca}, we can argue that bcLSTM does not effectively learn to rely on the context for semantic understanding improvement of the target utterance. Rather, it basically learns to mimic the affective content of the contextual utterances due to the label copying feature in the training dataset of IEMOCAP as discussed in \cref{sec:experiment-setup}.}

Hence, we observe that in IEMOCAP, there is not much of a drop in performance if we conduct the modifications to the dialogues with the same label constraint. Surprisingly, even in the $Replace$ setup, the performance drop is only around 2\% from the original setup. This means that even if we replace 30, 40, or 50 utterances in the past and future context with totally unrelated utterances (but belonging to the same emotion category) from other test dialogues, the performance can still be kept near the original F1-score of 61.9\% (\cref{tab:lca} Row 9, 11, 13). For large window sizes in the $Replace$ setup, even though the flow of the dialogue is entirely absent, the F1-Score is still around 59\% demonstrating the importance of label dependency in this particular dataset. Furthermore, in the different label constraint, the performance drop is significant even if we use the \textit{Concat} strategy in a small window (\cref{tab:lca} Row 14, 16). Although \textit{Concat} strategy retains the original utterance, some text belonging to other emotion label is concatenated which would confuse the model about the overall label orientation. In the \textit{Replace} setup the performance is even poorer (Row 15, 17, 19). All this results indicate the evidence of label dependency especially for the task of emotion classification.

We observe a similar kind of trend in the Persuasion for Good dataset. In both persuader and persuadee strategy classification, the drop in performance is much lesser compared to DailyDialog emotion or MultiWOZ intent.
As observed in \cref{table:result2}, current contextual models do not provide substantial improvements over non-contextual models in this dataset. Hence, we conjecture that bcLSTM model is unable to model context effectively for classifying strategy labels in this dataset.

Observing how the result changes by varying the window size, we conclude that even if the local context is augmented with same labels (SL) the long distant context helps in the strategy classification tasks. In the different label (DL) experiments, augmenting with different labeled utterances results in an adverse effect and the performance is worse than the GloVe CNN (without context classifier) model in \cref{table:result2}.
We conclude that even though the true context is not greatly helpful in this task (compared to other datasets), corrupting the context is unfavourable for the bcLSTM model and will result in poor performance.

In DailyDialog act, concatenation with same label setup is more helpful because the act labels are primarily driven by sentence type and concatenation is likely to provide a stronger signal. For example, an utterance with act label \textit{question}, concatenated with another \textit{question} is unlikely to be classified into something else. Finally, in DailyDialog emotion and MultiWOZ intent it can be concluded that content of the contextual utterances are extremely vital and corrupting them results in very poor performances.

\subsection{Influence of Utterance Positions in the Prediction}
In this setting, we try to understand whether there is a general or dataset specific trend between the prediction f1-score and position of the utterances. We want to examine if utterances at the beginning of the dialogues are relatively easier to classify compared to the utterances at the middle or the utterances at the end. We show this trend of performance against position in \cref{fig:position}.

We see that there is a decreasing trend of performance w.r.t position in all the tasks across the datasets. At the beginning of a dialogue, the initial utterances setup the flow of the conversation. Naturally, they have more information, they are more self-dependent and less context dependent. However, as the dialogue progresses, utterances tend to have lesser self-sufficient information to be classified correctly. The classification of these utterances thus become largely context dependant. This is evident from the plot of the GloVe CNN model in \cref{fig:position}. GloVe CNN is a non-contextual model and there is an overall decreasing trend as we move towards the right in the \textit{Position} axis. This supports our initial hypothesis that utterances appearing later in the dialogue are generally harder to classify on their own without the use of any contextual information.

We see a similar kind of trend in the bcLSTM model as well. In three of the six tasks, it uses contextual information to provide a significant improvement over the GloVe CNN model. However, RNNs are not capable of maintaining entirety of the relevant context (e.g., coreference) across time, and hence the network also loses necessary contextual information for correct classification as the dialogue progresses. Nevertheless, we still observe that for those three tasks, barring the extreme right end of the graph, the gap between bcLSTM and GloVe CNN widens as we move towards higher position indices. This suggest that, w.r.t GloVe CNN, the bcLSTM model finds it less difficult to classify utterances at the end compared to utterances in the beginning as a result of its contextual nature.

We believe one other possible reason for the decreasing trend could be because that there are only a few dialogues with a very large number of utterances. The scores for the rightmost position indices are thus calculated over a very small number of samples and are probably not the most accurate estimate of the overall trend. It is also a fact that for these far distant indices with very less number of samples could potentially cause the models to learn any intended position-specific contextual information or positional encoding.

\subsection{Performance for Label Shift}
\label{sec:ldist}

\begin{figure*}[ht!]
    \centering
    \begin{subfigure}{0.49\textwidth}
     \includegraphics[width=\linewidth]{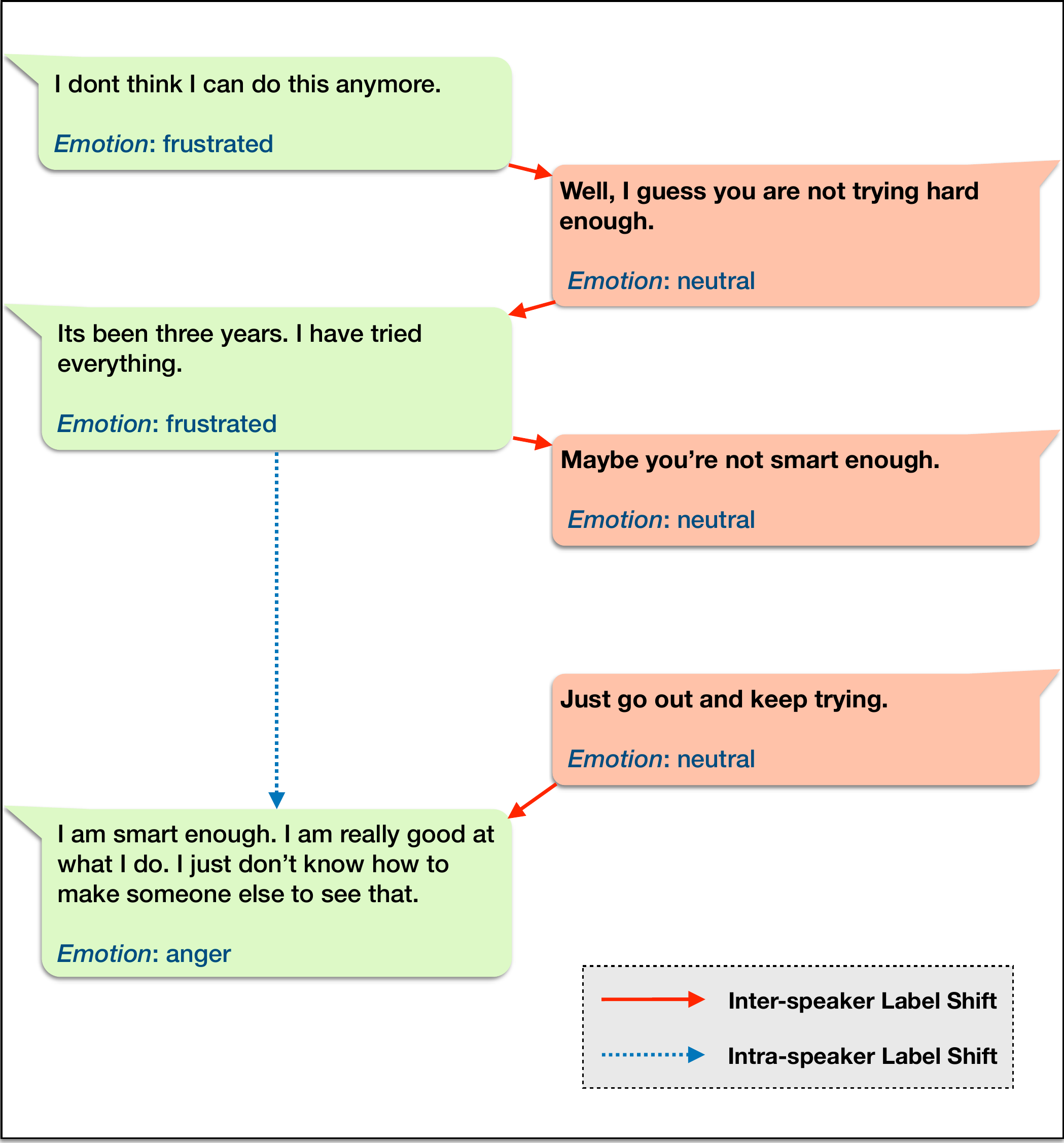}
      \label{fig:emo-shift}
     \end{subfigure}
     \begin{subfigure}{0.49\textwidth}
     \includegraphics[width=\linewidth]{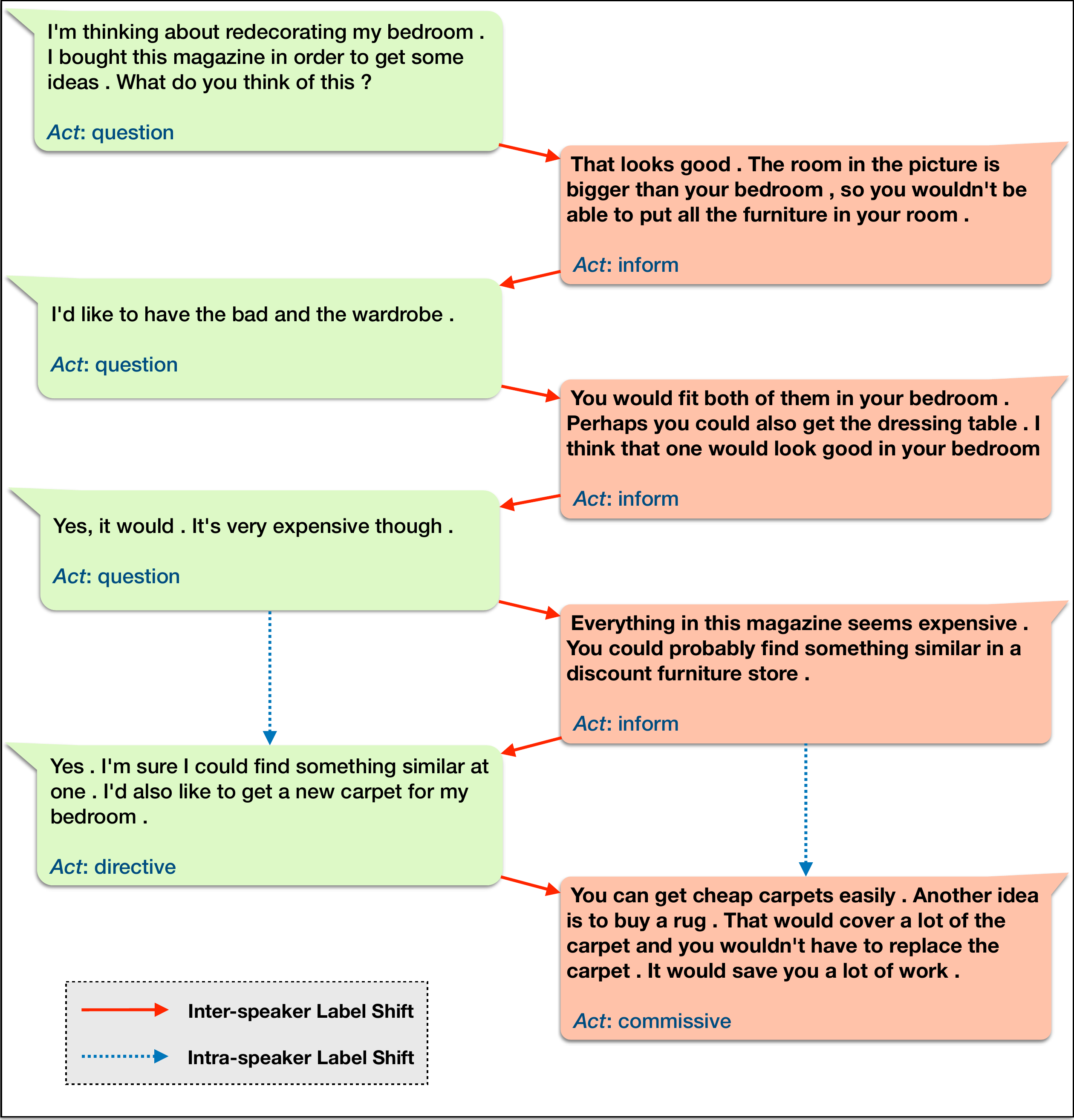}
      \label{fig:act-shift}
     \end{subfigure}
     \caption{Examples of label shift in IEMOCAP and DailyDialog datasets, respectively.}
     \label{fig:label-shift}
\end{figure*}

As illustrated in \cref{fig:heatmap1} and \cref{fig:heatmap2}, a few of our tasks of interest exhibit the label copying property which means consecutive utterances from the same speaker or different speakers often have the same emotion, act, or intent label. The inter-speaker and intra-speaker label copying is especially prevalent in the IEMOCAP emotion tasks, the DailyDialog act tasks, and the MultiWOZ intent tasks. Contextual models such as bcLSTM make correct predictions when utterances display such kind of continuation of the same label. But what happens when there is a change of label? Does bcLSTM continue to perform at the same level or is it affected from the change?
To understand this occurrence in more detail, we define this event as \emph{Label Shift} and look at the following two different kind of shifts that could happen in the course of a dialogue:
\begin{itemize}
    \item \emph{Intra-Speaker Shift}: The label of the utterance is different from the label of the previous utterance from the same speaker (refer to \cref{fig:label-shift}).
    \item \emph{Inter-Speaker Shift}: The label of the utterance is different from the label of the previous utterance from the non-target speaker (refer to \cref{fig:label-shift}).
\end{itemize}
In these two scenarios explained above, we are interested to see how LSTM performs at the utterances were the label shift takes place.

\begin{table*}[t]
  \centering
 \resizebox{\linewidth}{!}{
   \begin{tabular}{l|c|ccccccc}
    \toprule
  & \multirow{2}{*}{Setup} & \textbf{IEMOCAP} & \textbf{Dailydialog}  & \textbf{Dailydialog} &\textbf{MultiWOZ}   & \textbf{Persuasion} &  \textbf{Persuasion}\\
    \# & & Emotion & Emotion & Act & Intent & ER & EE\\
    \midrule
     1 & Original & 61.9 & 41.16 & 79.46 & 96.22 & 56.28 & 44.83 \\
     2 & Intra-Speaker Shift & 52.01 (13.2) & 44.23 (1.0) & 76.18 (2.9) & 94.91 (1.6) & 57.84 (6.9) & 49.4 (4.7) \\
     3 & Inter-Speaker Shift & 52.37 (22.0) & 47.77 (1.3) & 78.80 (4.9) & - & - & - \\
    \bottomrule
   \end{tabular}
   }
  \caption{Classification performance for utterances which exhibits \emph{Label Shift} in test data. Numbers in parenthesis indicate the average count of the corresponding shifts per dialogue. There is no \emph{Inter-Speaker Shift} in MultiWOZ or Persuasion for Good as we only classify user, persuader, or persuadee utterances. Scores are W-Avg F1 in IEMOCAP Emotion and MultiWOZ Intent; Macro F1 in the rest.
  }
  \label{tab:label-shift}
\end{table*}

\begin{table*}[t]
  \centering
  \resizebox{0.85\linewidth}{!}{
 \small
   \begin{tabular}{l|c|c|cc}
    \toprule
  & \multirow{2}{*}{Setup} & \multirow{2}{*}{Mode} & \textbf{IEMOCAP} & \textbf{Dailydialog}\\
    \# & & & Emotion & Emotion\\
    \midrule
     1 & Original & - & 61.9 & 41.16  \\
     2 & Intra-Speaker Shift & Emotion Shift & 52.01 (13.2) & 44.23 (1.0) \\
     3 & Inter-Speaker Shift & Emotion Shift & 52.37 (22.0) & 47.77 (1.3)  \\
    \midrule
     4 & Intra-Speaker Shift & Sentiment Shift & 53.21 (7.2) & 44.21 (1.0) \\
     5 & Inter-Speaker Shift & Sentiment Shift & 49.09 (13.6) & 50.61 (1.3) \\
    \midrule
     6 & Intra-Speaker Shift (w/o neutral) & Sentiment Shift & 51.98 (1.1) & 19.76 (0.02) \\
     7 & Inter-Speaker Shift (w/o neutral) & Sentiment Shift & 62.22 (1.0) & 45.63 (0.04) \\
    \bottomrule
   \end{tabular}
   }
  \caption{Classification performance for utterances which exhibits \emph{Emotion and Sentiment Shift} in test data. Numbers in parenthesis indicate the average count of the corresponding shifts per dialogue. Scores are W-Avg F1 in IEMOCAP Emotion and Macro F1 in DailyDialog.
  }
  \label{tab:emotion-sentiment-shift}
\end{table*}


We report results for utterances in the test data that show \emph{Intra-Speaker Shift} and \emph{Inter-Speaker Shift} in \cref{tab:label-shift}. The \emph{Inter-Speaker Shift} is not defined in MultiWOZ as we don't have intent labels for system utterances. We also don't report \emph{Inter-Speaker Shift} results in Persuasion for Good as the persuader and persuadee strategy set is different.

The emotion labels in IEMOCAP display the largest extent of label copying in \cref{fig:heatmap2} and label shift in \cref{fig:heatmap1}. We also observe in \cref{tab:label-shift} that label shifts occur with high frequency in IEMOCAP. These are the likely reasons why we observe significant number of errors for utterances with \emph{Label Shift} for this task in \cref{tab:label-shift}. The performance for both \emph{Intra-Speaker Shift} and \emph{Inter-Speaker Shift} stands at around 52.0\%, much lesser than the overall average of 61.9\% in test data. Although not as strong as IEMOCAP, the intra-speaker label copying feature can also be seen in MultiWOZ intent and DailyDialog act labels. For these two tasks, we again observe a drop of performance when either \emph{Intra-Speaker Shift} or \emph{Inter-Speaker Shift} occurs.

In contrast, the extent of transition is spread over a much larger combination of labels in DailyDialog emotion and Persuasion for Good. We observe that the results for utterances with \emph{Label Shift} in those tasks are in fact better than the overall score. In DailyDialog emotion, the scores are 44.23\% and 47.77\%, which is an improvement over the original 41.16\%. The scores of 57.84\% and 49.4\% in Persuasion for Good also stand over the scores of 56.28\% and 44.83\% in the original setup.

\paragraph{Sentiment Shift.} We further analyze the results for sentiment shift in \emph{intra-} and \emph{inter-speaker} level. For the emotion classification tasks, we group the different emotion labels into three broad categories: i) \textit{positive sentiment} group with emotions \textit{ excited, happy, surprise}, ii) \textit{negative sentiment} group with emotions \textit{angry, disgust, fear, frustrated, sad}, and iii) \textit{neutral sentiment} group with emotion \textit{neutral}. Sentiment shift is then defined as shifting from one of the three groups to any of the other two. We also define sentiment shift w/o neutral as switching from either \textit{positive} to \textit{negative} group, or \textit{negative} to \textit{positive} group.

The results for utterances showing the performance of bcLSTM at sentiment switching in \emph{intra-} and \emph{inter-speaker} scenarios are reported in \cref{tab:emotion-sentiment-shift}. In IEMOCAP, sentiment shift is naturally less frequent than emotion shift because of the emotion grouping. However, we found that sentiment and emotion shift results in almost similar kind of performance which can be attributed to the label dependency between the target speaker's utterances as explained in earlier sections. One noteworthy exception is \emph{inter-speaker} sentiment shift w/o neutral for which the W-Avg F1 is 62.22\%. One can infer the low prediction accuracy of bcLSTM when a sentiment shift between non-neutral  emotions (e.g., anger, happy, sad) to neutral emotion takes place between two speakers' consecutive utterances.

\begin{table*}[ht!]
  \centering
  \small
 \resizebox{\linewidth}{!}{
   \begin{tabular}{c|c|p{6.6cm}c|p{6.6cm}c}
    \toprule
    \multirow{2}{*}{Dataset} & \multirow{2}{*}{Size} & \multicolumn{2}{c|}{Train+Val} & \multicolumn{2}{c}{Test}\\
     &  & Pattern & Frequency (\%) & Pattern & Frequency (\%)\\
    \midrule

\multirow{10}{*}{\rot{IEMOCAP (Emotion)}} & \multirow{5}{*}{2} &  frustrated, frustrated  &  17.09  &  neutral, neutral  &  17.56  \\
& &  neutral, neutral  &  15.67  &  frustrated, frustrated  &  15.69  \\
& &  sad, sad  &  11.58  &  excited, excited  &  15.37  \\
& &  angry, angry  &  11.3  &  sad, sad  &  12.55  \\
& &  excited, excited  &  9.75  &  angry, angry  &  6.79  \\
\cline{2-6}
& \multirow{5}{*}{3} &  frustrated, frustrated, frustrated  &  12.54  &  neutral, neutral, neutral  &  14.44  \\
& &  neutral, neutral, neutral  &  12.18  &  excited, excited, excited  &  13.24  \\
& &  sad, sad, sad  &  10.16  &  frustrated, frustrated, frustrated  &  11.56  \\
& &  angry, angry, angry  &  8.74  &  sad, sad, sad  &  11.02  \\
& &  excited, excited, excited  &  8.04  &  angry, angry, angry  &  4.62  \\
\midrule
\multirow{10}{*}{\rot{DailyDialog (Emotion)}} & \multirow{5}{*}{2} &  neutral, neutral  &  76.33  &  neutral, neutral  &  74.17  \\
& &  neutral, happiness  &  7.72  &  neutral, happiness  &  8.03  \\
& &  happiness, happiness  &  5.49  &  happiness, happiness  &  5.76  \\
& &  happiness, neutral  &  3.6  &  happiness, neutral  &  3.84  \\
& &  neutral, surprise  &  1.54  &  neutral, surprise  &  1.22  \\
\cline{2-6}
& \multirow{5}{*}{3} &  neutral, neutral, neutral  &  70.99  &  neutral, neutral, neutral  &  67.71  \\
& &  neutral, neutral, happiness  &  7.89  &  neutral, neutral, happiness  &  8.07  \\
& &  happiness, happiness, happiness  &  3.56  &  happiness, neutral, neutral  &  3.65  \\
& &  happiness, neutral, neutral  &  2.8  &  happiness, happiness, happiness  &  3.39  \\
& &  neutral, happiness, neutral  &  2.54  &  neutral, happiness, neutral  &  2.86  \\
\midrule
\multirow{10}{*}{\rot{DailyDialog (Act)}} & \multirow{5}{*}{2} &  inform, inform  &  27.26  &  inform, inform  &  27.53  \\
& &  question, question  &  14.48  &  question, question  &  14.11  \\
& &  question, inform  &  11.24  &  question, inform  &  12.02  \\
& &  inform, question  &  6.64  &  inform, question  &  6.97  \\
& &  directive, inform  &  6.3  &  directive, inform  &  5.75  \\
\cline{2-6}
& \multirow{5}{*}{3} &  inform, inform, inform  &  18.36  &  inform, inform, inform  &  18.09  \\
& &  question, question, question  &  8.44  &  question, question, question  &  7.75  \\
& &  question, inform, inform  &  5.71  &  question, inform, inform  &  6.72  \\
& &  question, question, inform  &  5.46  &  question, question, inform  &  5.94  \\
& &  inform, question, inform  &  3.47  &  inform, question, inform  &  3.62  \\
\midrule
\multirow{10}{*}{\rot{MultiWOZ (Intent)}} & \multirow{5}{*}{2} &  find hospital, find hospital  &  16.71  &  find taxi, find taxi  &  17.58  \\
& &  find police, find police  &  15.75  &  find hospital, find hospital  &  14.62  \\
& &  find taxi, find taxi  &  15.27  &  book train, book train  &  13.14  \\
& &  book train, book train  &  11.69  &  find police, find police  &  12.71  \\
& &  find police, book hotel  &  5.73  &  find taxi, book restaurant  &  5.93  \\
\cline{2-6}
& \multirow{5}{*}{3} &  find hospital, find hospital, find hospital  &  11.82  &  find taxi, find taxi, find taxi  &  10.57  \\
& &  find police, find police, find police  &  9.9  &  find hospital, find hospital, find hospital  &  10.3  \\
& &  find taxi, find taxi, find taxi  &  9.9  &  find police, find police, find police  &  7.32  \\
& &  book train, book train, book train  &  6.39  &  book train, book train, book train  &  7.32  \\
& &  find taxi, find taxi, book restaurant  &  5.75  &  find taxi, find taxi, book restaurant  &  6.78  \\
\midrule
\multirow{12}{*}{\rot{Persuasion (ER)}} & \multirow{5}{*}{2} &  non strategy dialogue acts, non strategy dialogue acts  &  22.68  &  non strategy dialogue acts, non strategy dialogue acts  &  23.04  \\
& &  credibility appeal, credibility appeal  &  9.55  &  credibility appeal, credibility appeal  &  8.05  \\
& &  credibility appeal, non strategy dialogue acts  &  4.85  &  credibility appeal, non strategy dialogue acts  &  4.78  \\
& &  non strategy dialogue acts, credibility appeal  &  4.38  &  non strategy dialogue acts, credibility appeal  &  4.53  \\
& &  non strategy dialogue acts, donation information  &  3.43  &  donation information, non strategy dialogue acts  &  4.02  \\
\cline{2-6}
& \multirow{5}{*}{3} &  non strategy dialogue acts, non strategy dialogue acts, non strategy dialogue acts  &  13.55  &  non strategy dialogue acts, non strategy dialogue acts, non strategy dialogue acts  &  13.02  \\
& &  credibility appeal, credibility appeal, credibility appeal  &  5.34  &  credibility appeal, credibility appeal, credibility appeal  &  3.2  \\
& &  credibility appeal, credibility appeal, non strategy dialogue acts  &  2.58  &  non strategy dialogue acts, credibility appeal, credibility appeal  &  2.58  \\
& &  non strategy dialogue acts, credibility appeal, credibility appeal  &  2.12  &  credibility appeal, credibility appeal, non strategy dialogue acts  &  2.43  \\
& &  credibility appeal, non strategy dialogue acts, non strategy dialogue acts  &  2.01  &  donation information, non strategy dialogue acts, non strategy dialogue acts  &  2.32  \\
\midrule
\multirow{12}{*}{\rot{Persuasion (EE)}} & \multirow{5}{*}{2} &  other dialogue acts, other dialogue acts  &  20.49  &  other dialogue acts, other dialogue acts  &  20.34  \\
& &  positive reaction to donation, positive reaction to donation  &  10.35  &  positive reaction to donation, positive reaction to donation  &  9.64  \\
& &  other dialogue acts, positive reaction to donation  &  5.61  &  other dialogue acts, positive reaction to donation  &  5.91  \\
& &  positive reaction to donation, other dialogue acts  &  4.61  &  positive reaction to donation, other dialogue acts  &  4.04  \\
& &  other dialogue acts, ask org info  &  3.87  &  other dialogue acts, ask org info  &  3.3  \\
\cline{2-6}
& \multirow{5}{*}{3} &  other dialogue acts, other dialogue acts, other dialogue acts  &  12.6  &  other dialogue acts, other dialogue acts, other dialogue acts  &  11.51  \\
& &  positive reaction to donation, positive reaction to donation, positive reaction to donation  &  6.07  &  positive reaction to donation, positive reaction to donation, positive reaction to donation  &  4.8  \\
& &  other dialogue acts, positive reaction to donation, positive reaction to donation  &  2.89  &  other dialogue acts, positive reaction to donation, positive reaction to donation  &  2.56  \\
& &  positive reaction to donation, positive reaction to donation, other dialogue acts  &  2.51  &  positive reaction to donation, positive reaction to donation, other dialogue acts  &  2.43  \\
& &  other dialogue acts, other dialogue acts, positive reaction to donation  &  2.28  &  other dialogue acts, other dialogue acts, positive reaction to donation  &  2.11  \\

    \bottomrule
   \end{tabular}
   }
  \caption{Five most frequent \emph{intra-speaker} $n$-gram label patterns in various datasets. The numbers reported in the \textit{Frequency} column is the percentage of occurrence. For example, in IEMOCAP, 17.09\% of 2-grams in train+validation set is \{\textit{frustrated}, \textit{frustrated}\}.We report the top frequencies for train$+$validation set and test set in window sizes of 2, 3 
  labels.
  }
  \label{tab:top-intra}
\end{table*}


On the other hand, due to the high frequent occurrence of \emph{neutral} emotion in the DailyDialog dataset, the number of both emotion and sentiment shifts are very low in this dataset. The higher frequency of the neutral emotion in this dataset also corroborates bcLSTM to learn \emph{non-neutral} to \emph{neutral} emotion and sentiment shifts effectively. Hence, when we remove the \emph{neutral} emotion, the performance dipped significantly which implies that classifying utterances at sentiment shifts between \emph{non-neutral} emotions are relatively poor in both intra- and inter-speaker scenarios.


\subsection{Performance for $n$-gram Label Patterns}
\begin{table*}[ht!]
  \centering
 \resizebox{\linewidth}{!} {
   \begin{tabular}{c|c|p{5cm}c|p{5cm}c}
    \toprule
    \multirow{2}{*}{Dataset} & \multirow{2}{*}{Size} & \multicolumn{2}{c|}{Train+Val} & \multicolumn{2}{c}{Test}\\
     &  & Pattern & Frequency(\%) & Pattern & Frequency(\%) \\
    \midrule

\multirow{5}{*}{\begin{tabular}[c]{@{}c@{}}IEMOCAP\\ Emotion\end{tabular}} & \multirow{5}{*}{2} &  frustrated, frustrated  &  13.49  &  excited, excited  &  13.06  \\
& &  sad, sad  &  10.43  &  sad, sad  &  11.57  \\
& &  neutral, neutral  &  9.55  &  frustrated, frustrated  &  11.49  \\
& &  excited, excited  &  8.6  &  neutral, neutral  &  10.05  \\
& &  angry, angry  &  7.84  &  neutral, frustrated  &  7.42  \\
\midrule
\multirow{5}{*}{\begin{tabular}[c]{@{}c@{}}DailyDialog\\ Emotion\end{tabular}} & \multirow{5}{*}{2} &  neutral, neutral  &  75.81  &  neutral, neutral  &  73.0  \\
& &  happiness, happiness  &  6.6  &  happiness, happiness  &  6.82  \\
& &  neutral, happiness  &  6.3  &  neutral, happiness  &  6.68  \\
& &  happiness, neutral  &  3.52  &  happiness, neutral  &  3.86  \\
& &  neutral, surprise  &  1.61  &  surprise, neutral  &  1.34  \\
\midrule
\multirow{5}{*}{\begin{tabular}[c]{@{}c@{}}DailyDialog\\ Act\end{tabular}} & \multirow{5}{*}{2} &  question, inform  &  24.31  &  question, inform  &  24.15  \\
& &  inform, inform  &  19.8  &  inform, inform  &  20.44  \\
& &  inform, question  &  15.87  &  inform, question  &  16.0  \\
& &  directive, commissive  &  10.04  &  directive, commissive  &  9.93  \\
& &  inform, directive  &  6.26  &  inform, directive  &  6.37  \\
\midrule
\multirow{5}{*}{\begin{tabular}[c]{@{}c@{}}Persuasion\\ ER to EE\end{tabular}} & \multirow{5}{*}{2} &  non strategy dialogue acts, other dialogue acts  &  20.34  &  non strategy dialogue acts, other dialogue acts  &  17.39  \\
& &  credibility appeal, positive reaction to donation  &  5.78  &  credibility appeal, positive reaction to donation  &  6.68  \\
& &  credibility appeal, ask org info  &  4.79  &  credibility appeal, other dialogue acts  &  6.68  \\
& &  credibility appeal, other dialogue acts  &  4.49  &  non strategy dialogue acts, positive reaction to donation  &  5.21  \\
& &  personal related inquiry, other dialogue acts  &  4.29  &  task related inquiry, other dialogue acts  &  4.42  \\
\midrule
\multirow{5}{*}{\begin{tabular}[c]{@{}c@{}}Persuasion\\ EE to ER\end{tabular}} & \multirow{5}{*}{2} &  other dialogue acts, non strategy dialogue acts  &  20.04  &  other dialogue acts, non strategy dialogue acts  &  17.26  \\
& &  ask org info, credibility appeal  &  12.22  &  ask org info, credibility appeal  &  10.98  \\
& &  positive reaction to donation, non strategy dialogue acts  &  9.86  &  positive reaction to donation, non strategy dialogue acts  &  8.58  \\
& &  provide donation amount, non strategy dialogue acts  &  5.04  &  provide donation amount, non strategy dialogue acts  &  6.28  \\
& &  task related inquiry, non strategy dialogue acts  &  3.54  &  other dialogue acts, credibility appeal  &  4.39  \\
  \bottomrule
   \end{tabular}
  }
  \caption{Five most frequent \emph{inter-speaker} $2$-gram label patterns in various datasets. The numbers reported in the \textit{Frequency} column represent the percentage of occurrence. We restrict the window size to only 2 as we are reporting \emph{inter-speaker} transitions here. We report the top frequencies for train$+$validation set and test set. As MultiWOZ Intent does not have label for system utterances, \emph{inter-speaker} patterns cannot be defined.
  }
  \label{tab:top-inter}
\end{table*}
\begin{figure*}[ht!]
    \centering
     \includegraphics[width=\linewidth]{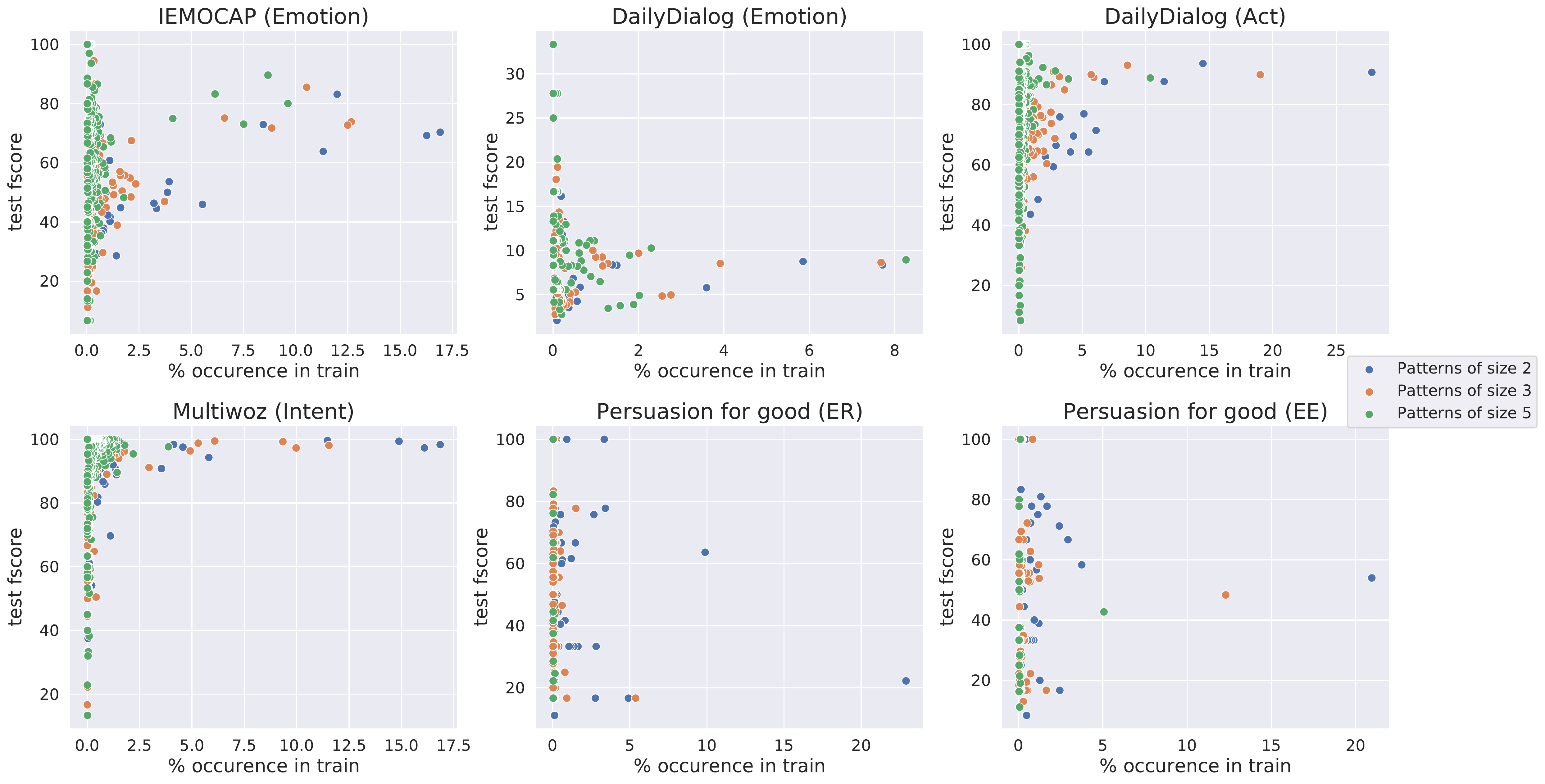}
     \caption{Performance of different $n$-gram label patterns in test data against their percentage of occurrence in the training data. We plot $n=2$-, $3$-, $5$-gram \emph{intra-speaker} label patterns here. Patterns which are more frequent in the training set are also likely to be predicted more correctly during evaluation.}
     \label{fig:freq-plot}
\end{figure*}

We perform a detailed study to understand whether $n$-gram label patterns that are frequently encountered by the learning algorithm during training, are more likely to be predicted correctly during evaluation. In this context, $n$-gram label patterns are simply the ordered list of labels from $n$ consecutive utterances in a dialogue. Contextual models such as bcLSTM will often learn the dependency of labels for patterns which it comes across more often in the training set. We then intend to see how the model performs in the evaluation set for all distinct patterns observed during the training. This distinct patterns will appear in a range of frequency in the training set, with some being much more common than others. By observing the test performance against the training frequency, we can understand how our models perform for label patterns occurring with different frequencies. This study is a continuation of the label dependent analysis we illustrate in \cref{fig:heatmap1} and \ref{fig:heatmap2}, and evaluation we perform in \cref{sec:ldist}.

We first tabulate the most frequent patterns in \emph{intra-speaker} level in \cref{tab:top-intra}. We report the top five frequent patterns with $n=2$-, $3$-grams. The numbers reported in the \textit{Frequency} column represent the percentage of occurrence. For example, in IEMOCAP, 17.09\% of 2-grams in train+validation set is \{\textit{frustrated}, \textit{frustrated}\}. For some of the tasks, the percentage occurrence reported in \cref{tab:top-intra} varies a lot even within the top five bracket, implying that consecutive utterances having some of those label patterns are much more frequent compared to the others. We also report the most frequent patterns in \emph{inter-speaker} level in \cref{tab:top-inter}. The numbers in the \emph{inter-speaker} level also indicate the presence of imbalanced $n$-gram label patterns.

Now, we collect all such \emph{intra-speaker} $n$-gram patterns (with $n=2,3,5$) appearing in the training data and plot their corresponding averaged performance if and when they appear in the test data in \cref{fig:freq-plot}. The performance score is plotted against the percentage of occurrence in the training data in \cref{fig:freq-plot}. Note that the percentage is reported for the top five patterns in \cref{tab:top-intra}, but is plotted for all possible patterns in \cref{fig:freq-plot}. The scores shown in \cref{fig:freq-plot} are computed from the predictions of the bcLSTM model. 

The scores in DailyDialog emotion are quite poor as most of the $n$-gram patterns contain one or more neutral emotion label. However, as neutral emotions are not considered in our evaluation setup, the scores are mostly in the lower range of 0-15\%. Apart from that, we see that there is a strong correlation between the higher frequency of occurrence in the training set and the performance in the test set. Patterns that are encountered more during the training phase are also predicted with higher F-scores during evaluation. All $n=2$-, $3$-, $5$-gram patterns with high frequencies show higher or at-least more than average F-scores reported for the whole dataset. A notable exception is the score of \{\textit{non strategy dialogue acts}, \textit{non strategy dialogue acts}\} pattern in Persuasion ER classification. Even though it constitutes around 24\% of 2-gram patterns in the training data, the test score is quite poor.

However, the scores for patterns that are less frequent in the training set varies considerably between the whole range of 0\%-100\%. Naturally, most of the lesser frequent patterns in the training data also appear less frequently in the test data. Hence, the scores plotted for these patterns are drawn from a very small number of samples and thus the variance is a lot higher in the left-most part of the plots in \cref{fig:freq-plot}.

\subsection{Sequence Tagging using Conditional Random Field (CRF)}
On the surface, the task of utterance level dialogue understanding looks similar to sequence tagging. Are there any distinct label dependency and patterns across the tasks that are dataset agnostic and likely to be captured by CRF~\cite{lafferty2001conditional}? In the quest to answer this, we plug in three different CRF layers on top of the bcLSTM network.

\begin{figure*}[ht!]
    \centering
    \begin{subfigure}{\textwidth}
     \includegraphics[width=\linewidth]{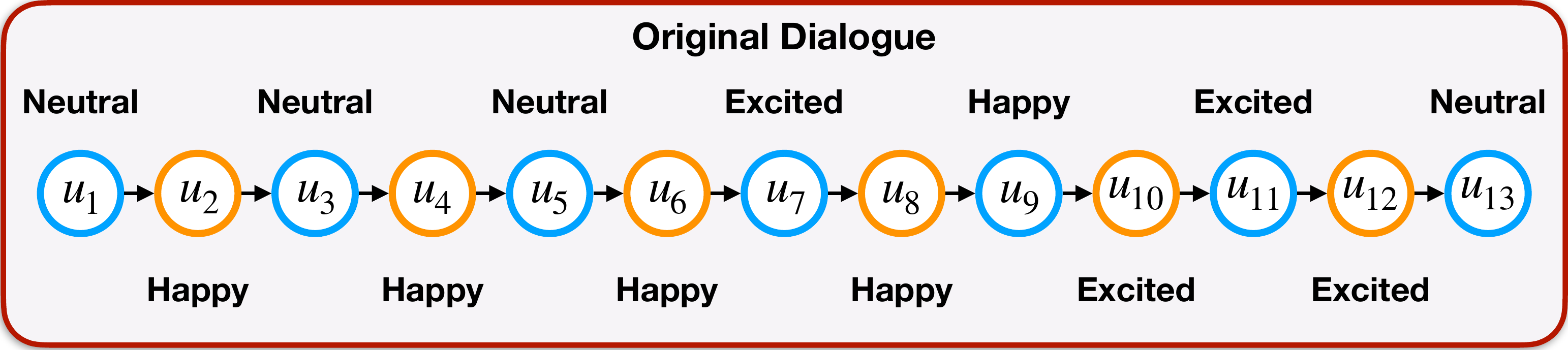}
      \label{fig:original-dialogue}
     \end{subfigure}
    \begin{subfigure}{\textwidth}
     \includegraphics[width=\linewidth]{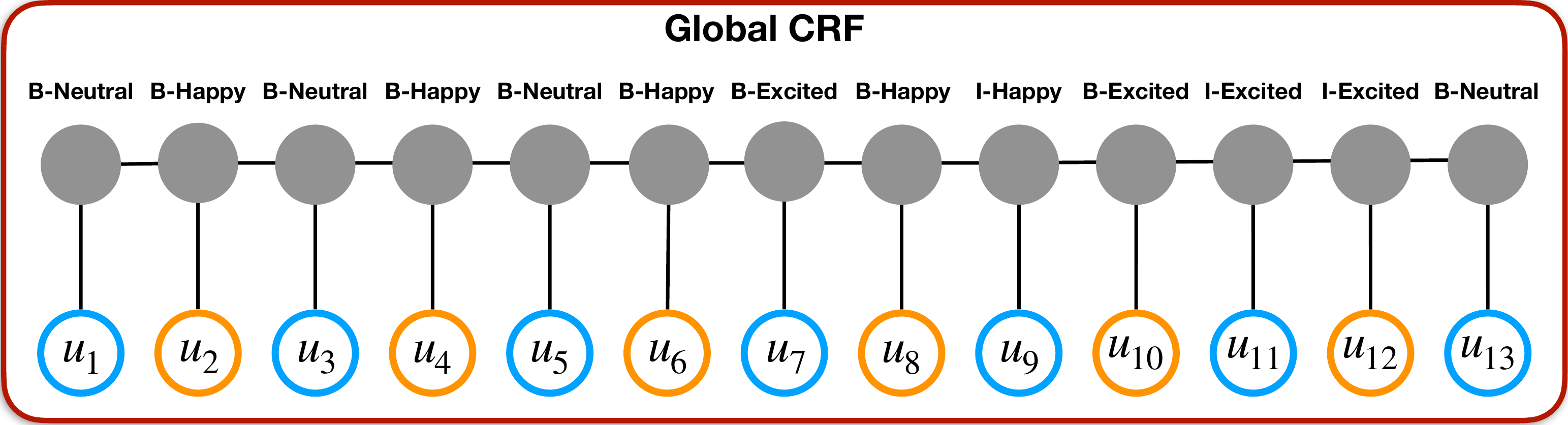}
      \label{fig:gcrf}
     \end{subfigure}
    \begin{subfigure}{\textwidth}
     \includegraphics[width=\linewidth]{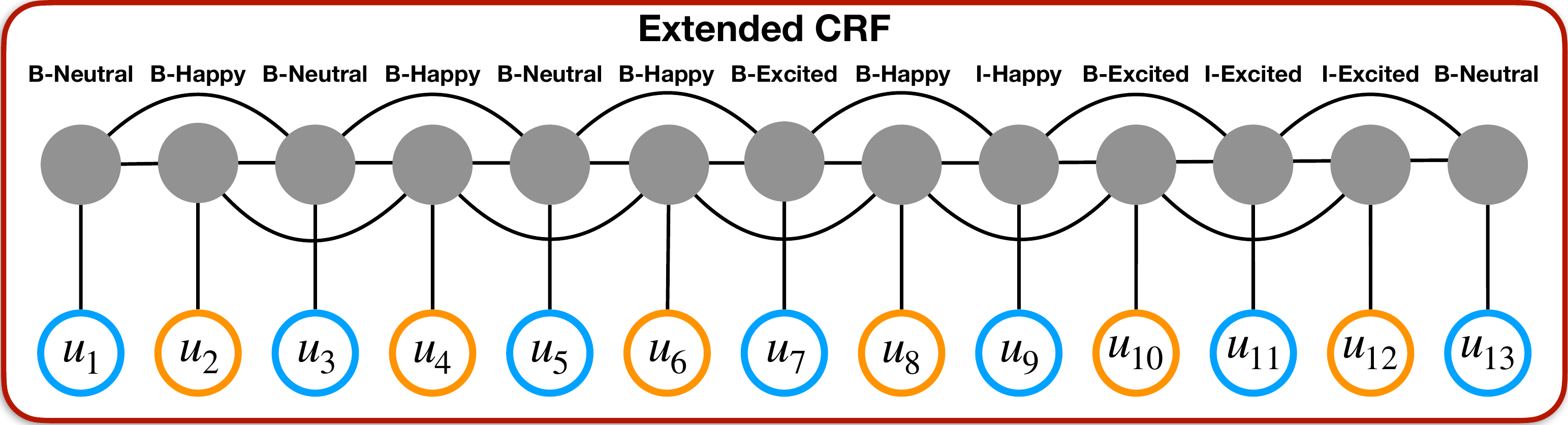}
      \label{fig:e}
     \end{subfigure}
     \begin{subfigure}{0.49\textwidth}
     \includegraphics[width=\linewidth]{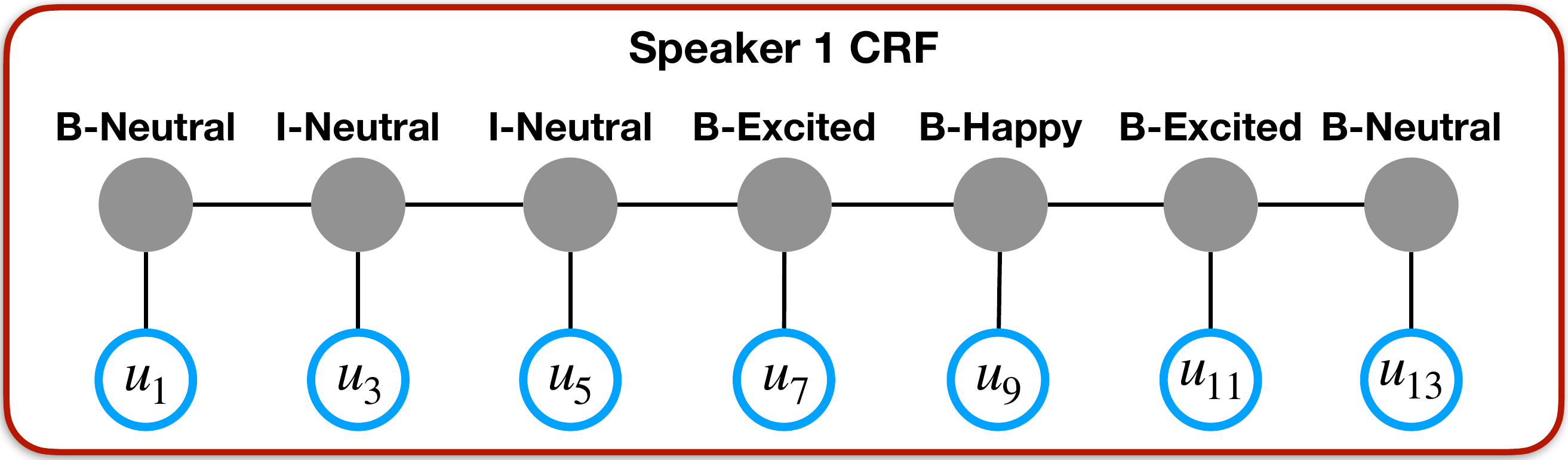}
      \label{fig:s1crf}
     \end{subfigure}
     \begin{subfigure}{0.49\textwidth}
     \includegraphics[width=\linewidth]{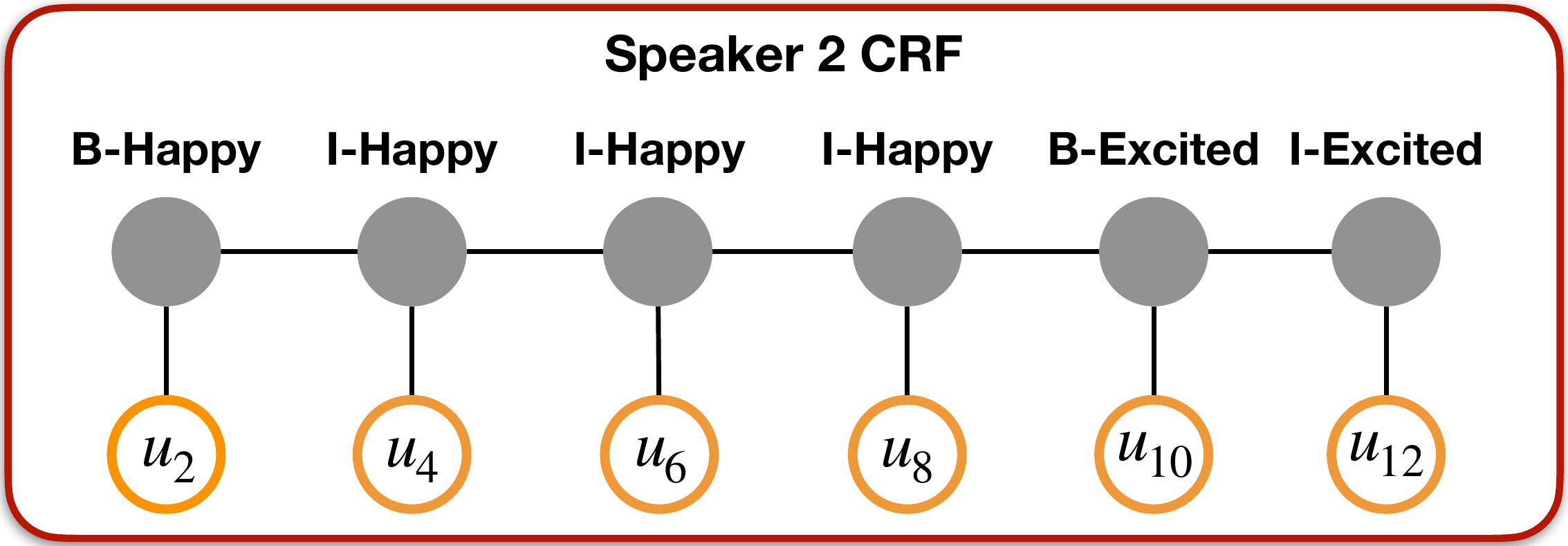}
      \label{fig:s2crf}
    \end{subfigure}
     \caption{Different CRF configurations.}
     \label{fig:crf}
\end{figure*}

\begin{table*}[ht]
  \centering
 \resizebox{\linewidth}{!}{
   \begin{tabular}{l||c||ccc|cc}
    \toprule
    \multirow{4}{*}{Methods} & \textbf{IEMOCAP} & & \multicolumn{4}{c}{\textbf{DailyDialog}} \\
     & Emotion & \multicolumn{3}{c|}{Emotion} &  \multicolumn{2}{c}{Act}\\
     
     \cline{2-7} & W-Avg F1 & W-Avg F1& Micro F1 & Macro F1 & W-Avg F1 & Macro F1\\
\hline
    GloVe CNN & 52.04 & 49.36 & 50.32 & 36.87 & 80.71 & 72.07 \\
    GloVe bcLSTM & 61.74 & 52.77 & 53.85 & 39.27 & \textbf{84.62} & 79.12 \\
    \quad  \footnotesize{w/o inter} & \textbf{63.73}  & 52.39 & 52.86 & \textbf{39.99} & 81.32 &  74.50 \\
    \quad  \footnotesize{w/o inter w/ speaker-CRF} & 62.94 & 52.47  & 54.04  & 39.77  & 81.19 & 74.12 \\
    \quad  \footnotesize{w/ global-CRF} & 61.62  & 53.05 & 53.86 & 39.27 & 83.91 & 79.10 \\
     \quad  \footnotesize{w/ global-CRF$_{\text{ext}}$} & 61.64 & 53.06 & 54.40 & 39.64 & 84.27 & \textbf{79.25} \\
    \quad  \footnotesize{w/ speaker-CRF} & 62.21 & \textbf{53.16} & \textbf{54.68} & \textbf{39.74} & 84.15 & 79.20 \\
    \bottomrule
   \end{tabular}
   }
  \caption{Classification performance in test data for emotion prediction in IEMOCAP, emotion prediction in DailyDialog, and act prediction in DailyDialog using different CRF configurations. All scores are average of at least 10 different runs. Test F1 scores are calculated at best validation F1 scores.}
  \label{table:crf1}
\end{table*}

\begin{table*}[ht]
  \centering
   \begin{tabular}{l||c||cc|cc}
    \toprule
    \multirow{4}{*}{Methods} & \textbf{MultiWOZ} & \multicolumn{4}{c}{\textbf{Persuasion For Good}} \\
     & Intent & \multicolumn{2}{c|}{Persuader} &  \multicolumn{2}{c}{Persuadee}\\
     
     \cline{2-6} & W-Avg F1 & W-Avg F1 & Macro F1 & W-Avg F1 & Macro F1\\
     \hline
    GloVe CNN & 84.30 & 67.15 & 54.45 & 58.00 & 41.03 \\
    GloVe bcLSTM & \textbf{96.14} & \textbf{69.26}  & 55.27 & 61.18 & 42.19\\
     \quad  \footnotesize{w/o inter} & 95.05 &67.81 & 53.24 & 59.44 & 40.63 \\
    \quad  \footnotesize{w/o inter w/ speaker-CRF} & 94.11 & 68.13 & 54.45 & 58.93 & 40.16 \\
    \quad  \footnotesize{w/ global-CRF/speaker-CRF} & 95.48 & 68.59 & 55.60 & 61.24 & 42.62 \\
    \quad  \footnotesize{w/ global-CRF$_{\text{ext}}$} & 95.51 & 69.23 & \textbf{56.80} & \textbf{61.89} & \textbf{43.68} \\
    \bottomrule
   \end{tabular}
  \caption{Classification performance in test data for intent prediction in MultiWOZ, persuader and persuadee strategy prediction in Persuasion for Good using different CRF configurations. All scores are average of atleast 10 different runs. Test F1 scores are calculated at best validation F1 scores. In MultiWOZ and Persuasion for Good, the global-CRF and speaker-CRF setting are identical as we only classify utterances coming from one of the speakers (user in MultiWOZ, persuader or persuadee in Persuasion for Good).}
  \label{table:crf2}
\end{table*}

\paragraph{Global-CRF.}
It is a linear chain CRF used on top of bcLSTM. In this setting, we do not consider speaker information. It can be defined using the equations below:
\begin{flalign}
P(Y|D) &= \frac{1}{Z(D)}\prod_{i=1}^{n}\phi_T(y_{i-1}, y_i)\phi_E(y_i, u_i),\\
Z(D) &= \sum_{y^\prime \in \mathcal{Y}} \prod_{i=1}^{n}\phi_T(y^\prime_{i-1}, y^\prime_i)\phi_E(y^\prime_i, u_i).
\end{flalign}

\paragraph{Global-CRF$_{\text{ext}}$.}

The linear-chain CRF is extended to include not only the transition potential from the previous label to the current label, but also from the prior-to-previous label. Concisely, the current label is predicated on the previous two labels. Therefore, the transition potential function $\phi_T$ takes one extra argument $y_{i-2}$. The advantage here is it also considers the previous label from the target speaker should utterance $i-2$ have come from the target speaker. This becomes useful in the tasks where the speakers tend to retain label from its last utterance. It can be defined using the equations below:
\begin{flalign}
P(Y|D) &= \frac{1}{Z(D)}\prod_{i=1}^{n}\phi_T(y_{i-2}, y_{i-1}, y_i)\phi_E(y_i, u_i),\\
Z(D) &= \sum_{y^\prime \in \mathcal{Y}} \prod_{i=1}^{n}\phi_T(y^\prime_{i-2}, y^\prime_{i-1}, y^\prime_i)\phi_E(y^\prime_i, u_i).
\end{flalign}

\paragraph{Speaker-CRF.}

In this setting, we use two distinct CRFs for the two speakers in a dialogue. Inter-speaker label dependency and transitions are not likely to be captured in this setting by the CRFs.

\paragraph{Negative Results.}

Aside from well-known sequence tagging tasks, such as, Named Entity Recognition (NER) and Part of Speech Tagging, CRF does not improve the performance of utterance-level dialogue understanding tasks. There could be multiple reasons as below:
\begin{itemize}
    \item As shown in \cref{fig:controlling_vars}, a dialogue is governed by multiple variables or pragmatics, e.g., topic, personal goal, past experience, expressing opinions or presenting facts based on personal knowledge, and the role of the interlocutors. Hence, the response pattern can vary depending on these variables. The personality of the speakers add an extra layer of complexity to this which causes speakers to respond differently under the same circumstances. An identical utterance can be uttered with different emotions by two different speakers. CRF relies on surface label patterns which can vary with datasets. Due to this dynamic nature of dialogues and the presence of latent controlling variables, the label transition matrix of CRF does not learn any distinct pattern that is complementary to what is learned by the feature extractor.
    \item Some of the datasets --- IEMOCAP and MultiWOZ --- contain distinct label-transition patterns (see \cref{fig:heatmap1} and \cref{fig:heatmap2}) between the same and distinct speakers. We expected bcLSTM w/ global-CRF to outperform vanilla bcLSTM on these two datasets. However, we do not observe any statistically significant improvement using bcLSTM w/ global-CRF over bcLSTM. We posit that the evident label-transition patterns that exist in these two datasets are straightforward to capture without a CRF. In fact, we also tried GloveCNN with a CRF layer on it, and surprisingly the result was not significantly higher than that of GloveCNN. This can be attributed to the absence of explicit contextual and label transition-based features in the CRF.
\end{itemize}

\paragraph{Results in IEMOCAP, DailyDialog and Persuasion for Good Datasets.}
We observe a \emph{minor} performance improvement in the IEMOCAP and DialyDialog datasets using speaker-CRF for emotion recognition. This observation directly correlates to the experiment under ``w/o inter" setting in the \cref{table:result2-w/o-inter}. In ``w/o inter" setting, contextual utterances of the speaker B are not utilized to classify utterances of speaker A vice versa. The results do not improve when we use speaker-level CRF on bcLSTM under the ``w/o inter" setting. From these observations, we can conclude that CRF is not learning any distinct label dependency and transition patterns that are not learned by the feature extractor or bcLSTM alone.

Global-CRF$_{\text{ext}}$ shows significant performance improvement on the Persuasion for Good dataset. Some of the key controllable factors of the dialogues such as topics in this dataset are fixed and can be learned intrinsically by the classifier. The scope of the dialogues in this dataset is very limited as there are only two possible outcomes of the dialogues  -- \emph{agree to donate}, and \emph{disagree to donate}. Hence, there can be some label transition patterns learned by the Global-CRF$_{\text{ext}}$ using a larger label-context window in the transition potential.

\subsection{Case Studies}
\label{sec:case-study}

\begin{figure}[ht!]
    \centering
    \includegraphics[width=\linewidth]{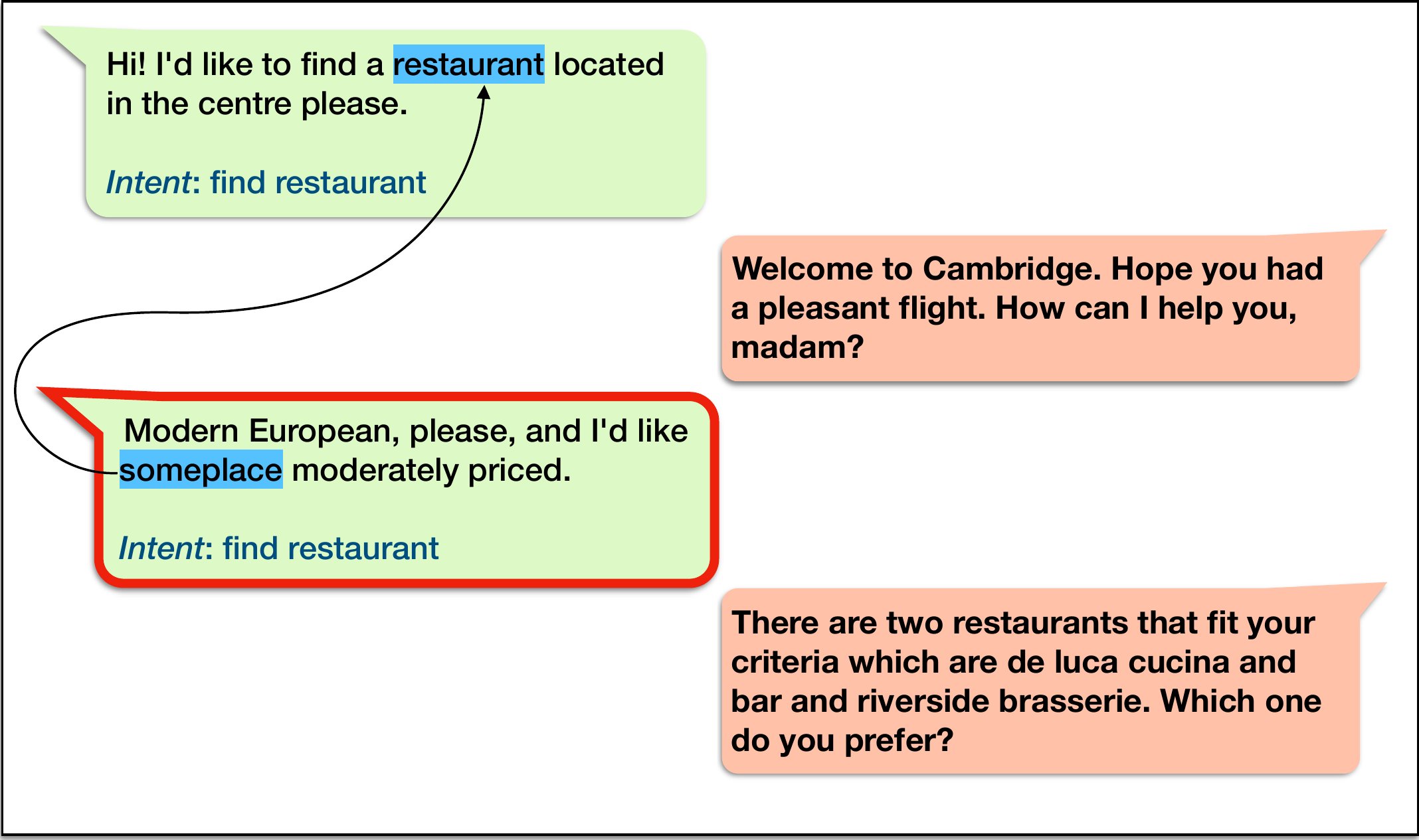}
    \caption{Context dependency by implicit mention in MultiWOZ dataset; red border indicates target utterance.}
    \label{fig:intent-ex1}
\end{figure}

\begin{figure}[ht!]
    \centering
    \includegraphics[width=\linewidth]{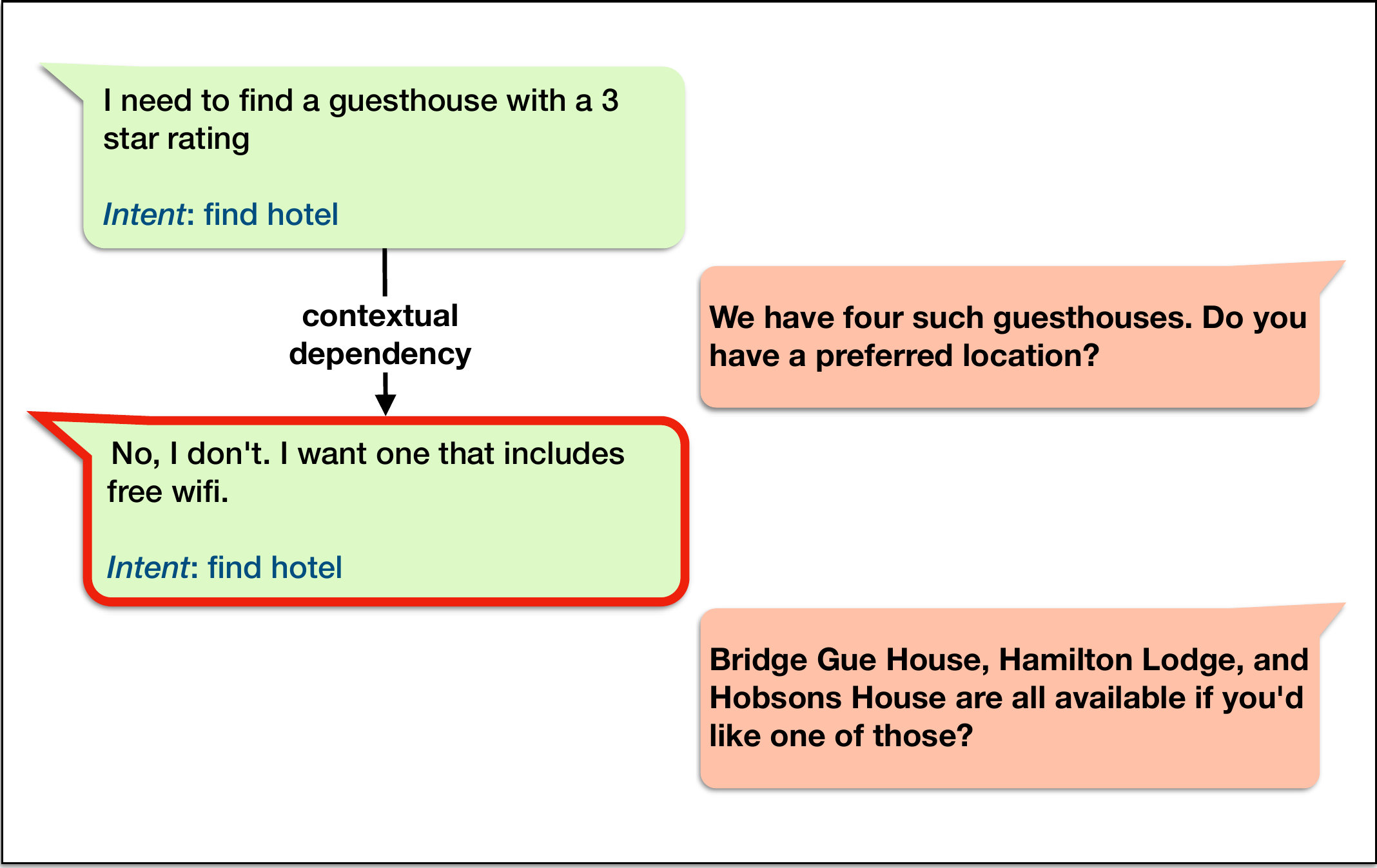}
    \caption{Context dependency in MultiWOZ dataset; red border indicates target utterance.}
    \label{fig:intent-ex2}
\end{figure}

The impetus behind context modeling in utterance-level dialogue understanding tasks is to capture the missing pieces necessary to understand a given utterance. Context often contain these missing pieces. \cref{fig:intent-ex1} illustrates this point where the intent of the target utterance could be resolvable by decoding the coreference of \emph{someplace} as \emph{restaurant}. Similarly, in \cref{fig:intent-ex2}, the correct intent could be determined through the context given by the previous utterance from the same speaker.

\begin{figure}[ht!]
    \centering
    \includegraphics[width=\linewidth]{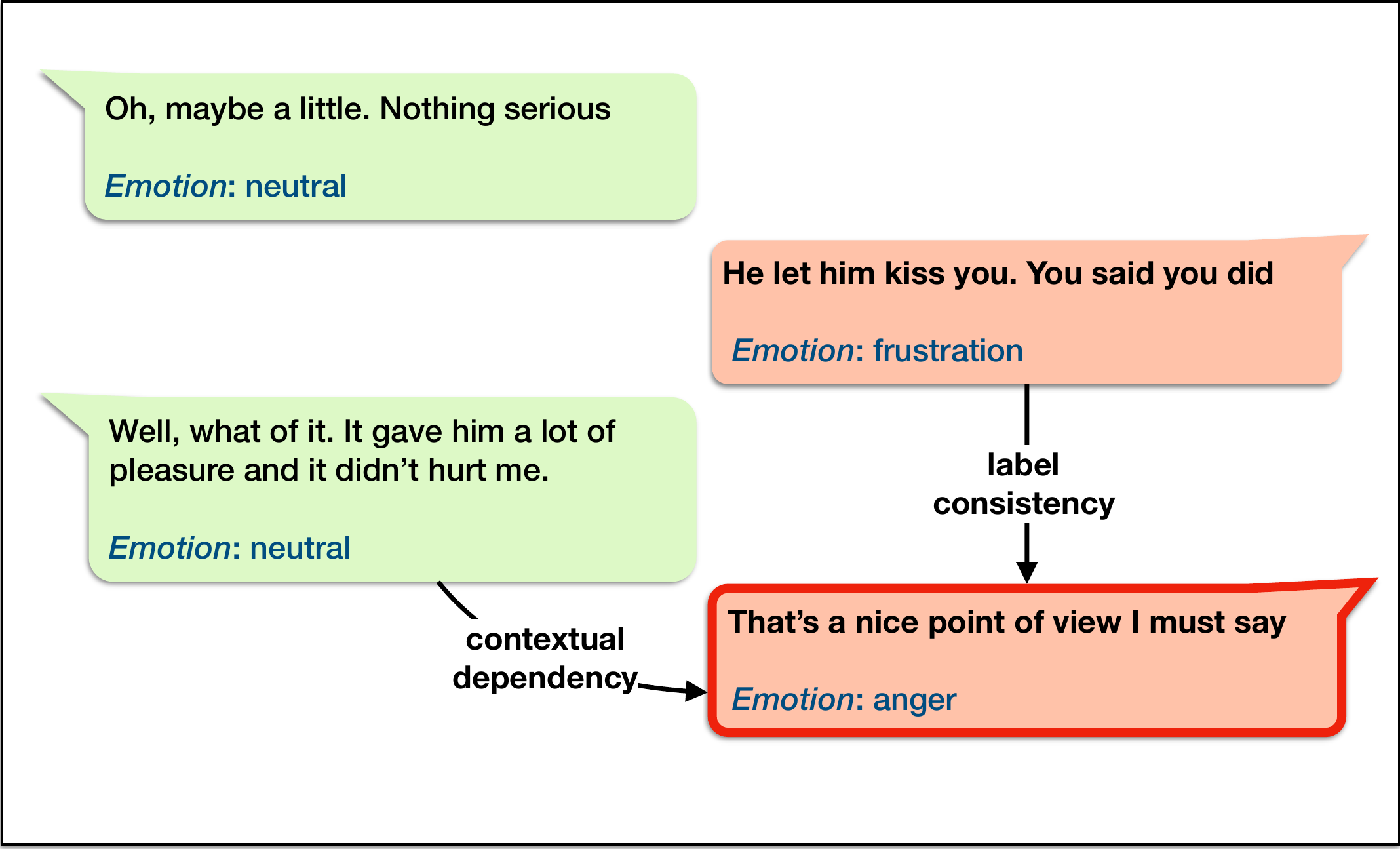}
    \caption{Context dependency in IEMOCAP dataset; red border indicates target utterance.}
    \label{fig:emo-ex1}
\end{figure}

The role of context is pivotal in the case of sarcasm. The target utterance in \cref{fig:emo-ex1} can only be correctly construed as sarcastic through the consideration of contextual utterances. The context is two way in this particular case. Firstly, the non-target speaker's nonchalant attitude infuriates the target speaker. On the other hand, the previous utterance of the target speaker suggests his foul mood that exacerbates in the target utterance. Contextual model bcLSTM could invoke these kind of contextual reasoning to arrive at the correct output label.

\textbf{It must be noted that the possible explanations shown in these examples are contrived. Whereas, the labels are produced by bcLSTM.}

We repeatedly observed across various utterance-level dialogue understanding tasks that GloVe CNN fail and bcLSTM succeed in producing correct labels in such context-reliant cases. To verify the role of context, we removed the contextual utterances around the target utterances and observed similar misclassification produced by GloVe CNN and bcLSTM alike. This indicates the likely ability of bcLSTM to capture the right context from the neighbouring utterances in a dialogue. \textbf{However, the inner workings of such networks still remain veiled to this day.} Thus, in the future, we should design approaches that are more explicit about the reasoning behind the output. Such explainable AI systems could pave the way to even richer systems.

\section{Future Directions}
Future work on utterance-level dialogue understanding could focus on a number of different directions:
\begin{itemize}
    \item As evidenced by results in \cref{table:result-w/o-inter} and \ref{table:result2-w/o-inter} current contextual models such as bcLSTM often lacks the ability to make effective use of context considering inter- and intra-speaker dependencies. It is particularly true for the emotion classification tasks in IEMOCAP and DailyDialog, where we found that dropping inter- and intra-speaker specific context leads to an improvement over the results of bcLSTM which uses full context. bcLSTM is thus not efficient in making use of the contextual utterances of both the interlocutors and lacks the ability to use the right amount of context. Future work in this direction could focus on development of better contextual models which are efficient in making use of their context considering speaker specific information. One promising direction is to use both context and speaker sensitive dependence which has been shown to be effective for emotion recognition \cite{zhang2019modeling,ghosal2019dialoguegcn}.


    \item \textbf{Task-specific Context Modeling: } Both interlocutors' utterances play a vital role in several tasks (refer to \cref{table:result2-w/o-inter}) addressed in this paper---intent classification in the MultiWOZ dataset, Persuader's and Persuadee's act classification in the \emph{Persuasion for Good} dataset. However, in some other tasks, i.e., emotion classification on the IEMOCAP and DailyDialog, act classification on the DailyDialog dataset, dropping target speaker's, and non-target speaker's utterances from context exhibit contrasting results (refer to \cref{table:result-w/o-inter}). In particular, for the act classification task in the DailyDialog dataset, non-target speaker's utterances are more informative as compared to the target speaker's utterances. We observe a stark opposite outcome for the emotion classification task in IEMOCAP and DailyDialog datasets where removing non-target speaker's utterances improves the overall performance. Interestingly, in the case of the DailyDialog dataset, the same contextual input yields two divergent trends in the results for two different tasks. Due to these contrasting yet interesting phenomena, we believe \textbf{it may not be optimal to adopt a task-agnostic unified context modeling approach for all the tasks} which call for task-specific context modeling approaches.

    \item Given an utterance, present contextual approaches to these problems can't explain where and how contextual information are used. In particular, they fail to explain their decisions which is a general problem of the AI models. In this work, we only investigate the general trend in how contextual information can aid dialogue understanding tasks. Our probing techniques do not attempt to reason \emph{how} contextual information help to infer the labels of target utterances e.g., whether the model relies on coreferential, affective contextual information, and as shown in \cref{fig:context-role-all}, how it jointly fuses these different types of contextual information for reasoning.

    \item Speaker-specific modeling shows its efficacy in most of the tasks that we consider in this paper. However, in this work, we chose to exclude it from our detailed analysis. In the future, we should strive to address this issue and analyze why and how speaker specific context modeling is effective for utterance-level dialogue understanding.

    \item The perturbation and adversarial attacking strategies used in this work are task agnostic which may not be ideal as the utterance-level dialogue understanding tasks differ from each other. In the future, we plan to design task-specific probing strategies to gain further insights from these contextual models.

    \item \textbf{Adapting the Proposed Probing Methods to Other Tasks:} The proposed probing methods engineered to understand the role of context can easily be adapted or extended to other context-dependent tasks --- summarization, dialogue generation, document-level sentiment analysis to name a few. With ample computational support, the role of RoBERTa like models in context modeling can be analyzed. This aligns well with the quest of explaining RoBERTa and other transformer-based models in contextual tasks by the means of attention visualization, measuring cosine similarity between the [CLS] token and the tokens in the contextual utterances to understand the role of these tokens in inferring the target utterances. The latter can be very useful in explaining the case studies in \cref{sec:case-study}.

    \item \textbf{What about Multi-party Dialogues?} Readers at this point may ponder the absence of multi-party dialogues in our study. The primary rationale for this is the additional complexity associated with multi-party dialogues. Multi-party dialogues involve many speakers and hence introduce complex coreferences that make inferences and context modeling harder than dyadic dialogues. The level of convolution that multi-party dialogues bring can be considered as a separate topic of research. MELD~\cite{poria2019meld} is one of the publicly available datasets for emotion and sentiment classification in multi-party datasets. Our preliminary experiments on this dataset with bcLSTM and DialogueRNN, reported in \cite{poria2019meld}, shows only a slight improvement over the non-contextual models like GloVe CNN. MELD contains very short utterances, like \emph{yeah}, \emph{oh}, which although appear \emph{neutral}, contain \emph{non-neutral} emotions when perceived in their associated context. This solidifies the need for further research on context representation modeling to understand multi-party dialogues and this is one of our future research goals.
\end{itemize}

\section{Conclusion}

This paper establishes a unified baseline for all the utterance-level dialogue understanding subtasks. Furthermore, we probed the contextual baseline bcLSTM with different strategies engineered to understand the role of context. This consequently lends us insight into the behaviour of bcLSTM at the presence of various context perturbations. Such probes have bolstered many interesting intuitions about utterance-level dialogue understanding---the role of label dependency and future utterances; the robustness of contextual models as opposed to their non-contextual counterparts against adversarial probes; the impact of position of an utterance on its correct classification. We also compared two different mini-batch-creation schemes for training---dialogue-based and utterance-based mini-batch---and compared their performance under varied settings. We believe that these probing strategies can be straightforwardly adapted to other context-reliant tasks.

\subsubsection*{Acknowledgments}
This research is supported by ($i$) A*STAR under its RIE 2020 Advanced Manufacturing and Engineering (AME) programmatic grant, Award No. --  A19E2b0098, and ($ii$) DSO, National Laboratories, Singapore under grant number RTDST190702 (Complex Question Answering).
We are thankful to Rishabh Bhardwaj for his time and insightful comments toward this work.

\bibliographystyle{acl_natbib}
\bibliography{refs}

\end{document}